\documentclass[12pt,a4paper]{article}
\usepackage[margin=1in]{geometry}
\usepackage{mathptmx,amsmath,amssymb,xfrac,mathtools,booktabs,longtable, multirow}
\usepackage{bbm}
\usepackage{lscape}
\usepackage[stable]{footmisc}
\usepackage{subcaption}
\usepackage{changepage}
\usepackage{fixltx2e}
\usepackage{titlesec}

\titleformat{\paragraph}
{\normalfont\normalsize\bfseries}{\theparagraph}{1em}{}
\titlespacing*{\paragraph}
{0pt}{3.25ex plus 1ex minus .2ex}{1.5ex plus .2ex}

\usepackage{cite}
\usepackage{natbib}
\usepackage[table]{xcolor}
\usepackage{array}
\usepackage[onehalfspacing]{setspace}
\usepackage{amsmath}
\DeclareMathOperator*{\argmin}{arg\,min}
\usepackage{graphicx}
\usepackage{caption}
\usepackage{subcaption}
\usepackage{float}
\usepackage{relsize}
\usepackage{url}
\usepackage[toc,page]{appendix}
\usepackage{pdflscape}

\newlength\savedwidth

\usepackage[aboveskip=1pt,labelfont=bf,labelsep=period,justification=raggedright,singlelinecheck=off]{caption}

\usepackage{authblk}
\title{\Large Demand Forecasting for Platelet Usage: from Univariate Time Series to Multivariate Models}
\author[a]{Maryam Motamedi}
\author[a]{Jessica Dawson}
\author[b,c,a]{Na Li}
\author[a,*]{Douglas G. Down}
\author[c,d]{Nancy M. Heddle}

\affil[a]{Department of Computing and Software, McMaster University, Hamilton, Ontario L8S 4L8, Canada}
\affil[b]{Community Health Sciences, University of Calgary, Calgary, Alberta T2N 1N4, Canada}
\affil[c]{McMaster Centre for Transfusion Research, Department of Medicine, McMaster University, Hamilton, Ontario L8S 4L8, Canada}
\affil[d]{Centre for Innovation, Canadian Blood Services, Ottawa, Ontario K1G 4J5, Canada}
\affil[ ]{Email: motamedm@mcmaster.ca [Motamedi]; dawsor1@mcmaster.ca [Dawson]; na.li@ucalgary.ca [Li];\newline downd@mcmaster.ca [Down]; heddlen@mcmaster.ca [Heddle]}
\date{}                     
\begin{document}
\maketitle
\noindent\rule{\textwidth}{0.5pt}
\begin{abstract}
Platelet products are both expensive and have very short shelf lives. As usage rates for platelets are highly variable, the effective management of platelet demand and supply is very important yet challenging. The primary goal of this paper is to present an efficient forecasting model for platelet demand at Canadian Blood Services (CBS). To accomplish this goal, five different demand forecasting methods, ARIMA (Auto Regressive Integrated Moving Average), Prophet, lasso regression (least absolute shrinkage and selection operator), random forest and LSTM (Long Short-Term Memory) networks are utilized and evaluated via a rolling window method. We use a large clinical dataset  for a centralized blood distribution centre for four hospitals in Hamilton, Ontario, spanning from 2010 to 2018 and consisting of daily platelet transfusions along with information such as the product specifications, the recipients' characteristics, and the recipients' laboratory test results. This study is the first to utilize different methods from statistical time series models to data-driven regression and machine learning techniques for platelet transfusion using clinical predictors and with different amounts of data. We find that the multivariate approaches have the highest accuracy in general, however, if sufficient data are available, a simpler time series approach appears to be sufficient. We also comment on the approach to choose predictors for the multivariate models.
\end{abstract}

\noindent \textbf{Keywords}: demand forecasting; time series forecasting; platelet products; blood demand and supply chain; long short-term memory networks.

\section{Introduction}
\label{sec:Intro}
Platelet products are a vital component of patient treatment for bleeding problems, cancer, AIDS, hepatitis, kidney or liver diseases, traumatology and in surgeries such as cardiovascular surgery and organ transplants \citep{kumar2015platelet}. In addition, miscellaneous platelet usage and supply are associated with several factors such as patients with severe bleeding, trauma patients, aging population and emergence of a pandemic like COVID-19 \citep{stanworth2020effects}. The first two factors affect the uncertain demand pattern, while the latter two factors result in donor reduction. Platelet products have five to seven days shelf life before considering test and screening processes that typically last two days \citep{fontaine2009improving}, so the remaining shelf life of the platelets that arrive at the hospitals is typically between three to five days. The extremely short shelf life along with the highly variable daily platelet usage makes platelet demand and supply management a highly challenging task, invoking a robust blood product demand and supply system.

Canadian Blood Services (CBS) is the national blood supplier for Canadian patients. The current blood supply chain for CBS is an integrated network consisting of a regional CBS distribution centre and several hospitals, as illustrated in Figure 1. Hospitals request blood products from the regional blood centres for the next day, yet, the regional blood centres are not aware of the actual demands as each hospital has its own blood bank. Furthermore, recipients’ demographics and hospitals’ inventory management systems are not disclosed to CBS or the regional blood centres. Hospitals hold excess inventory to manage the highly variable platelet demand. However, holding surplus inventory makes platelet demand forecasting even more challenging for blood distribution centres. In particular, it results in wastage and does not allow the blood suppliers to recognize the real demand, which in turn yields an inefficient demand forecasting system. Accordingly, accurately forecasting the demand for blood products is a core requirement of a robust blood demand and supply management system. 

This research is motivated by the platelet management problem confronted by CBS. Currently, there is a yearly wastage rate of about 9\% for the hospitals in Hamilton, Ontario (with an approximate cost of $\$$400,000 per year) and about 15\% for CBS with seasonal variation \citep{Auditor:2020}. The current frequent same-day urgent orders, considered as shortages, are about 14\% of the total orders in Hamilton, Ontario. Given the wastage rates and shortage rates, forecasting short-term demand for platelets is of particular value. In this research, we forecast platelet demand to overcome the mentioned challenges. The forecasting models used in this research can help both suppliers and consumers of platelets to make operational decisions, including inventory decisions, by providing information about future demand.

\begin{figure}[ht]
  \centerline{\includegraphics[width=0.8\hsize, height=0.4\hsize]{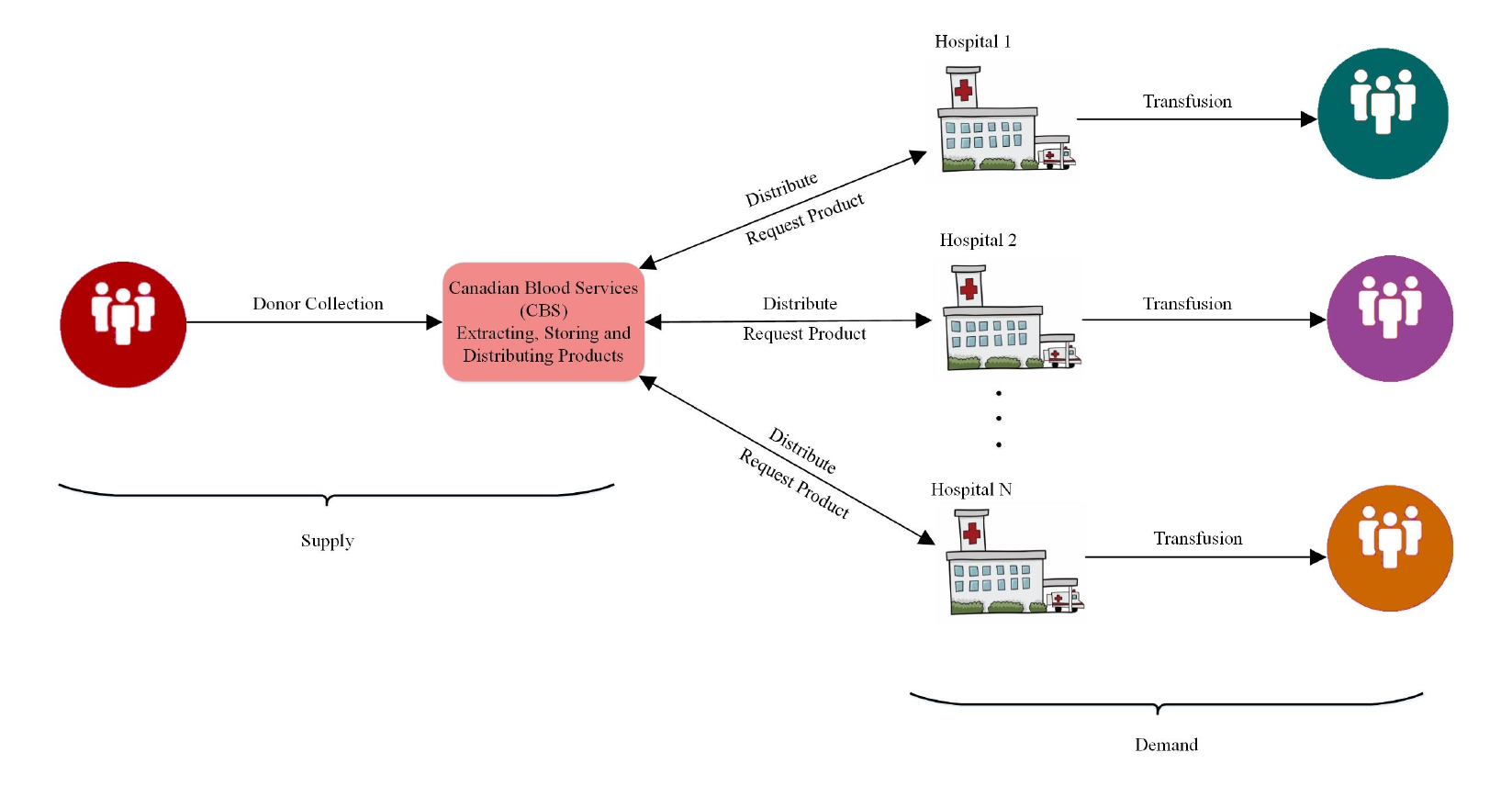}}
  \caption{CBS blood supply chain with one regional blood centre and multiple hospitals}
  \label{fig:CBS}
\end{figure}

We study a large clinical database with 61377 platelet transfusions for 47496 patients in hospitals in Hamilton, Ontario from 2010 to 2018. We analyse the database to extract trends and patterns, and find relations between the demand and clinical predictors. We find that there are three key issues that should be considered in the demand forecasting process: seasonality, the effect of clinical predictors on demand, and nonlinear dependencies among these clinical predictors. Consequently, we progressively build five demand forecasting models (of increasing complexity) that address these issues. The proposed methods are applied on the data to determine the influence of demand history as well as clinical predictors on demand forecasting. The first two models are univariate time series that only consider the demand history, while the remaining three methods, multivariate regression, random forest, and artificial neural networks, consider clinical predictors. These five methods are utilized to pursue the following goals: i) more precise platelet demand forecasting for the benefit of both CBS and hospitals, ii) reducing the bullwhip effect, as a consequence of effective demand forecasting; and iii) investigating the impact of clinical predictors on the platelet demand. The main contributions of this study are as follows:
\begin{enumerate}
\item We analyze the time series of platelet transfusion data by decomposing it into trend, seasonality and residuals, and detect meaningful patterns such as weekday/weekend and holiday effects that should be considered in any platelet demand predictor.
\item We utilize five different demand forecasting methods from univariate time series methods to multivariate methods including regression and machine learning. Since CBS has no access to recipients' demographic data, our first method, ARIMA, only considers demand history for forecasting, while the second model, Prophet, includes seasonalities, trend changes and holiday effects. We found that these models have issues with respect to accuracy, in particular when a limited amount of data are available, accordingly we apply a lasso regression method to include clinical predictors for demand forecasting. Finally, random forests and LSTM networks are used for demand forecasting to explore the nonlinear dependencies among the clinical predictors and the demand.
\item We utilize clinical predictors in the demand forecasting process, and select those that are most impactful by using lasso regression for structural variable selection and regularization. Results show that incorporating the clinical predictors in demand forecasting enhances the forecasting accuracy.
\item We investigate the effect of different amounts of data on the forecasting accuracy and model performance and provide a holistic evaluation and comparison for different forecasting methods to evaluate the effectiveness of these models for different data types, providing suggestions on using these robust demand forecasting strategies in different circumstances. Results show that when having a limited amount of data (two years in our case), multivariate models outperform the univariate models, whereas having a large amount of data (eight years in our case) results in the ARIMA model performing nearly as well as the multivariate methods.
\end{enumerate}

The rest of this paper is organized as follows. In Section \ref{sec:LiteratureReview} we provide a literature review of demand forecasting methods for blood products, with a focus on platelets. Section \ref{sec:Methods} provides the data description and an overview of the five models used for forecasting platelet demand. The main results of the study are presented in Section \ref{sec:Res}. In Section \ref{sec:Comparison}, a comparison of the models is provided, and finally, in Section \ref{sec:Conclusion}, concluding remarks are provided, including a discussion of ongoing work for this problem.
\section{Literature Review}
\label{sec:LiteratureReview}
There is a limited literature on platelet demand forecasting; most investigates univariate time series methods. In these studies, forecasts are based solely on previous demand values, without considering other features that may affect the demand. \cite{critchfield1985automatic} develop models for forecasting platelet usage in a blood centre using several time series methods including Moving Average (MA), Winter’s method and Exponential Smoothing (ES). \cite{silva2012decision} develop a Box-Jenkins Seasonal Autoregressive Integrated Moving Average (BJ-SARIMA) model to forecast weekly demand for blood components, including platelets, in hospitals. They later extend this work in \citep{silva2013demand}. \cite{kumari2016efficient} propose a blood inventory management model for the daily supply of platelets focusing on reducing platelet shortages by applying three time series methods, MA, Weighted Moving Average (WMA) and ES. \cite{volken2018red} use generalized additive regression and time-series models with ES to predict future whole blood donation, including platelets, and RBC transfusion trends. \cite{fanoodi2019reducing} use artificial neural networks and ARIMA models to forecast platelet demand by considering daily demands for eight types of blood platelets. They consider different demand data lags, 1 day, 2 days, 3 days, 4 days, 5 days, 6 days, 1 week, 15 days, 30 days, 90 days, 120 days, and 365 days, as the input data for the artificial neural networks. 

On the other hand, many studies focus on univariate whole blood demand forecasting rather than a specific blood product, using time series or machine learning models. \cite{frankfurter1974management} develop transfusion forecasting models using ES methods for a blood collection and distribution centre. \cite{fortsch2016reducing} apply various approaches to predict blood demand such as Naïve, MA, ES, and multiplicative Time Series Decomposition (TSD), amongst which a Box-Jenkins (ARMA) approach results in the highest prediction accuracy. \cite{lestari2017forecasting} apply four time series models, MA, WMA, ES and ES with trend, to forecast demand for blood components. \cite{TWUMASI20221258} apply K-Nearest Neighbour regression (KNN), Generalised Regression Neural Network (GRNN), Neural Network Auto-regressive (NNAR), Multi-Layer Perceptron (MLP), Extreme Learning Machine (ELM), and an LSTM network for forecasting and backcasting blood demand to predict future and lost past demand data respectively, by using a rolling-origin procedure.

Several recent studies include additional features other than demand history for demand forecasting. \cite{drackley2012forecasting} estimate long-term blood demand for Ontario, Canada based on previous transfusions' age and sex-specific patterns. They forecast blood supply and demand for Ontario by considering demand and supply patterns, and demographic forecasts, with the assumption of fixed patterns and rates over time. \cite{khaldi2017artificial} apply Artificial Neural Networks (ANNs) to forecast the monthly demand for three blood components, red blood cells (RBCs), platelets and plasma for a case study in Morocco. \cite{guan2017big} propose an optimization ordering strategy in which they forecast the platelet demand for several days into the future and build an optimal ordering policy based on the predicted demand, concentrating on minimizing the wastage. Their main focus is on an optimal ordering policy and they integrate their demand model in the inventory management problem, meaning that they do not try to precisely forecast the platelet demand. \cite{li2021decision} develop a hybrid model consisting of seasonal and trend decomposition using Loess (STL) time series and eXtreme Gradient Boosting (XGBoost) for RBC demand forecasting and incorporate it in an inventory management problem.

In this study, we utilize multiple demand forecasting methods, including univariate analysis (time series methods) and multivariate analysis (regression and machine learning methods), and evaluate the performance of these models for platelet demand forecasting. We explore the value gained from including a range of clinical predictors for platelet demand forecast models. More specifically, we consider clinical predictors, consisting of laboratory test results, patient characteristics and hospital census data as well as operational related indicators, including the previous week’s platelet usage and previous day's received units with the aim of accurate demand forecasting. In addition to the linear effects of the clinical predictors, we study the nonlinear effect of these clinical predictors in our choice of machine learning models. Moreover, we explore the effect of having different amounts of data on the accuracy of the forecasting methods. To the best of our knowledge, this study is the first that utilizes and evaluates different demand forecasting methodologies from univariate time series to multivariate models for platelet products and explores the effect of the amount of available data on these approaches.

\section{Methods}
\label{sec:Methods}
In this section, we present the general problem setting, a comprehensive data description and the methods used for data exploration. We also provide an overview of the five models used for forecasting platelet demand, an overview of the rolling window analysis used for retraining the models, and the error measures used for evaluation.
\subsection{Problem Setting}
In this study, we consider a blood supply system consisting of one regional CBS distribution centre and four major hospitals operating in the city of Hamilton, Ontario. As a result of internal inventory management practices, these four hospitals are considered as one entity. At the beginning of the day, hospitals receive platelet products that were ordered on the previous day, from CBS. In the case of shortages, hospitals can place expedited (same-day) orders at a higher cost. Prior to September 2017, platelets had five days of shelf life, while after this date, the shelf life of platelets was increased to seven days. After exceeding the shelf life, platelet products are expired and discarded.
\subsection{Data Description}
\label{sec:DataDesc}
The data in this study are constructed by processing CBS shipping data and the TRUST (Transfusion Research for Utilization, Surveillance and Tracking) database at the McMaster Centre for Transfusion Research (MCTR) for platelet transfusion in Hamilton hospitals. The study is approved by the Canadian Blood Services Research Ethics Board and the Hamilton Integrated Research Ethics Board (HiREB number 7293). The data are high dimensional, with more than 100 variables that can be divided into four main groups: 1.\ the blood inventory data such as product name and type, received date, expiry date, 2.\ patient characteristics such as age, gender, patient ABO Rh blood type, 3.\ the transfusion location such as intensive care, cardiovascular surgery, hematology, and 4.\ available laboratory test results for each patient such as platelet count, hemoglobin level, creatinine level, and red cell distribution width. The laboratory tests are prescribed by physicians based on clinical needs and can help to decide whether a patient needs platelet transfusion. In this research, the laboratory test results are processed and used along with other information to forecast future platelet demand.

Additionally, we add new calculated predictors such as the number of platelet transfusions in the previous day and previous week, the number of received units in the previous day, and day of the week. Table \ref{tab:VarDef} gives the set of predictors that are used in this study along with their descriptions. These predictors are selected by a lasso regression model \citep{tibshirani1996regression} which is explained in detail in Section \ref{sec:Lasso}. As we can see from Table \ref{tab:VarDef}, predictors have different ranges, and hence are standardized by $z$-score normalization. All data processing and analysis and model implementations are carried out using the Python 3.7 programming language.
\begin{table}[H]
    \centering
    \caption{Data variable definition and description}
    \scalebox{0.7}{
    \begin{tabular}{| l | l |}
    \hline
    \textbf{Name} & \textbf{~~~~~~Description~~~~~~} \\ \hline
    abnormal\_ALP & Number of patients with abnormal alkaline phosphatase	\\ \hline
    abnormal\_MPV & Number of patients with abnormal mean platelet volume	\\ \hline
    abnormal\_hematocrit & Number of patients with abnormal hematocrit \\ \hline
    abnormal\_PO2 & Number of patients with abnormal partial pressure of oxygen \\ \hline
    abnormal\_creatinine & Number of patients with abnormal creatinine \\ \hline
    abnormal\_INR	& Number of patients with abnormal international normalized ratio \\ \hline
    abnormal\_MCHb & Number of patients with abnormal mean corpuscular hemoglobin	\\ \hline
    abnormal\_MCHb\_conc & Number of patients with abnormal mean corpuscular hemoglobin concentration \\ \hline
    abnormal\_hb & Number of patients with abnormal hemoglobin \\ \hline
    abnormal\_mcv & Number of patients with abnormal mean corpuscular volume	\\ \hline
    abnormal\_plt & Number  of patients with abnormal platelet count	\\ \hline
    abnormal\_redcellwidth & Number of patients with abnormal red cell distribution width	\\ \hline
    abnormal\_wbc & Number of patients with abnormal white cell count	\\ \hline
    abnormal\_ALC & Number of patients with abnormal absolute lymphocyte count \\ \hline
    location\_GeneralMedicine & Number of patients in general medicine \\ \hline
    location\_Hematology & Number of patients in hematology \\ \hline
    location\_IntensiveCare & Number of patients in intensive care \\ \hline
    location\_CardiovascularSurgery & Number of patients in cardiovascular surgery \\ \hline
    location\_Pediatric	& Number of patients in pediatrics \\ \hline
    Monday & Indicating the day of the week	\\ \hline
    Tuesday & Indicating the day of the week \\ \hline
    Wednesday & Indicating the day of the week \\ \hline
    Thursday & Indicating the day of the week \\ \hline
    Friday & Indicating the day of the week	\\ \hline
    Saturday & Indicating the day of the week \\ \hline
    Sunday & Indicating the day of the week	\\ \hline
    lastWeek\_Usage & Number of units transfused in the previous week \\ \hline
    yesterday\_Usage & Number of platelet units transfused in the previous day \\ \hline
    yesterday\_ReceivedUnits & Number of units received by the hospital in the previous day \\ \hline
    \end{tabular}
    }
    \captionsetup{justification=centering}
    \label{tab:VarDef}
\end{table}

\subsection{Exploratory Analysis for Trends, Seasonality and Holiday Patterns}
\label{sec:ExplorAnalysis}
In order to propose a short-term demand forecasting model, we first explore the data for identifying temporal (daily/monthly) patterns that can inform our demand forecasting techniques. In particular, we investigate correlations among the predictors, seasonality, day of the week, and non-stationarity effects.

textbf{Identify non-stationarity:} The Augmented Dickey-Fuller (ADF) test \citep{cheung1995lag} is applied on the time series data to examine the stationarity.

\textbf{Identify seasonality:} We apply the Seasonal and Trend decomposition using Loess (STL) model to decompose the time series data into trend, seasonality, and residuals. We also apply the one-way ANOVA test to compare the means of the transfused units in different months, and the means of the transfused units during weekdays and weekends. Moreover, we explore the trend, holidays, weekly seasonality, and yearly seasonality using the Prophet model, explained in Section \ref{sec:Univariate}.

\textbf{Identify day of the week effect:} We also compare the mean daily units transfused based on day of the week by plotting the mean against day of the week, and also by applying the t-test to compare the mean daily units transfused during weekdays and weekends.


\subsection{Demand Forecasting Models}
\label{sec:DemandForecasting}
This section explains the five forecasting models used for forecasting the platelet demand in Hamilton hospitals. The ARIMA and Prophet models are univariate models that forecast the demand based on demand history. Lasso regression, random forest, and LSTM networks are multivariate models that consider various predictors in addition to demand history for forecasting the demand.
\subsubsection{Univariate Models}
\label{sec:Univariate}
The univariate models, ARIMA and Prophet, forecast the demand solely based on the previous demand values. The ARIMA model does not consider seasonality in data and is considered as a baseline model. The Prophet model, on the other hand, considers trend, seasonality, and holidays for forecasting the demand.

\noindent \textbf{ARIMA Model}

An autoregressive integrated moving average model consists of three components, an autoregressive (AR) component that considers a linear combination of lagged values as the predictors, a moving average (MA) component of past forecast errors (white noise), and an integrated component where differencing is applied on the data to make it stationary. Let $y_1, y_2, \dots, y_t$ be the demand values over time period $t$; the time series data can be written as:
\begin{equation}\
    y_t = f(y_{t-1},y_{t-2},y_{t-3},… ,y_{t-n}) + \epsilon_t
\end{equation}
An ARIMA model assumes that the value of demand is a linear function of a number of previous past demand values and previous error values. Thus, the ARIMA model can be written as:
\begin{multline}
      \hat{y}_t = \mu + \vartheta_1 y_{t-1} + \vartheta_2 y_{t-2} + \vartheta_3  y_{t-3} + \cdots + \vartheta_p  y_{t-p}  + \epsilon_t \\
      - \phi_1 \epsilon_{t-1} - \phi_2 \epsilon_{t-2} - \phi_3 \epsilon_{t-3} - \cdots - \phi_q \epsilon_{t-q}
\end{multline}
where $\hat{y}_t$ is the response variable (the predicted demand), $\mu$ is a constant, $\vartheta_i$ and $\phi_j$ are model parameters in which $i = {1,2,\dots,p}$ and $j = {0,1,2,\dots,q}$, $p$ and $q$ are the model orders and define the number of autoregressive terms and moving average terms, respectively.

In order to fit an ARIMA model, first the ADF test is applied on the time series data to examine the stationarity, and the standard auto\_arima() function in Python is used for hyperparameter tuning and determining the optimal model order. A function is developed in Python to implement the ARIMA model via a rolling-origin strategy.

\noindent \textbf{Prophet Model}

Prophet is a time series model introduced by \cite{taylor2018forecasting} that considers common features of business time series: trends, seasonality, holiday effects and outliers. The Prophet model was developed for forecasting events created on Facebook and is implemented as an open source software package in both Python and R. Let $g_t$ be the time series trend function which shows the long-term pattern of data, $s_t$ be the seasonality which captures the periodic fluctuations in data such as weekly, monthly or yearly patterns, and finally $h_t$ be the non-periodic holiday effect. These features are combined through a generalized additive model (GAM) \citep{hastie1987generalized}, and the Prophet time series model can be written as:
\begin{equation}\
    \hat{y}_t = g_t + s_t + h_t + \epsilon_t
\end{equation}
The normally distributed error $\epsilon_t$ is added to model the residuals. We use the Prophet library in Python for implementing the Prophet model and develop a function for implementation via a rolling-origin strategy.
\subsubsection{Multivariate Models}
In order to explore the effect of including clinical predictors in the forecasting process, in the next step we introduce three multivariate models that incorporate clinical predictors: lasso regression, random forest, and LSTM networks. These machine learning models are implemented to forecast the demand based on demand history and multiple predictors. Lasso regression is used as a forecasting model and a variable selection method to select the most relevant predictors for the multivariate models.

\noindent \textbf{Lasso Regression}
\label{sec:Lasso}

We use lasso regression \citep{tibshirani1996regression} since it allows predictors to be included in the demand forecasting process. The lasso regression model performs variable selection to reduce the complexity of the model, as well as improving the prediction accuracy. By considering the actual demand on day $t$ ($t = 1,2,...,N$) as $y_t$ and the predicted demand on day $t$ as the product of the clinical predictors ($z_{tj}$) and their corresponding coefficients $\beta_j$, where $j = 1,2,...,M$ specifies the clinical predictor, the lasso regression is the solution to the following optimization problem:
\begin{equation}\label{eq:1}
    \argmin{\sum_{t=1}^{N} (y_t - \sum_{j}^{} \beta_j z_{tj})^2 + \lambda \sum_{j=1}^{M} |   \beta_j|}
\end{equation}
\begin{equation}\label{eq:2}
    \text{subject to }~ \sum_{j = 1}^{M} |\beta_j| \le \tau.
\end{equation}
The optimization problem defined in (\ref{eq:1})-(\ref{eq:2}) chooses the coefficients, $\beta$, that minimize the sum of squares of the errors between the actual values ($y$) and the response variable values, with a sparsity penalty ($\lambda$) on the sum of the absolute values of the model coefficients. Constraint (\ref{eq:2}) forces some of the coefficients (that have a minor contribution to the estimate) to be zero. predictors that have non-zero coefficients are selected in the model, and the response variable is calculated as follows:
\begin{equation}
    \hat{y}_t = \beta z_t
\end{equation}
In this study, lasso regression is used as a variable selection method to find important predictors for platelet demand. Subsequently, this information is used for demand forecasting. We use the LassoCV function from the sklearn package in Python to implement the lasso regression. The penalty coefficient $\lambda$ is chosen through five-fold cross-validation. A function is developed to implement the lasso regression via a rolling-origin strategy.

\noindent \textbf{Random Forest}

Random forests, first proposed by \cite{ho1995forests}, are ensemble methods that use decision trees. We chose to explore random forests as they can capture nonlinear relationships between predictors while also being interpretable, as what a decision tree does can be understood by simply looking at it. Decisions trees in a forest are trained using bootstrapped samples and are only allowed to consider a subset of the predictors when choosing splits. Considering the actual demand on day $t$ as $y_t$, and the set of days in the bootstrap samples as $D$, a tree starts with a root node that has an attached value $\mu$: 

\begin{equation}\label{eq:treeBegin}
    \mu=\frac{1}{|D|} \sum_{t \in D} y_t
\end{equation}

\noindent This node creates two child nodes that separate data points based on a clinical predictor, $u$, where one node gets data with the value of $u$ on day $t$ ($z_{tu}$) less than a value $v$ and the other node gets data with $z_{tu}$ greater than or equal to $v$. These child nodes have attached values calculated in the same way as the root, $\mu_1 = \frac{1}{|\{t | z_{tu} < v\}|} \sum_{t:z_{tu}<v} y_t$ and $\mu_2 = \frac{1}{|\{t | z_{tu} \geq v\}|} \sum_{t:z_{tu} \geq v} y_t$.


\noindent The split measures, $u$ and $v$, are chosen by minimizing the variance of the model.
\noindent A random forest grows a number of these trees, $K$, and produces a prediction for a set of clinical predictors, $z_t$, by averaging together the predictions of each of the trees:

\begin{equation}
    \hat{y}_t = \sum_{i=1}^K T_i(z_t)
\end{equation}

\noindent where each tree $T_i$ takes a set of clinical predictors and traverses the nodes of tree $i$ using the splits found with the above equations. Forecasting problems can have linear or nonlinear relationships among the model predictors. Random forests can work on both linear and nonlinear data, and are able to capture nonlinear dependencies among the predictors. We use the RandomForestRegressor function from the scikit-learn package in Python to implement the random forest. Hyperparameter tuning is achieved by using grid search on the number of trees, maximum tree depth, and the number of features to consider when looking for the best split. The best split in a tree is chosen by minimizing MSE (Mean Square Error) and five-fold cross-validation is used to reduce overfitting. We developed a function in Python to implement the random forest model via a rolling-origin strategy.

\noindent \textbf{LSTM Network}

LSTM networks are a class of recurrent neural networks (RNN) that were introduced by \citep{hochreiter1997long} and are capable of learning long-term dependencies in sequential data. In theory, RNNs should be capable of learning long-term dependencies, however they suffer from the so-called vanishing gradient problem. Consequently, LSTM networks are designed to resolve this issue. An LSTM network maps a set of input neurons (also called units) to a set of output neurons through a hidden layer. A neuron or unit in an LSTM network consists of an input gate ($i_t$), a forget gate ($f_t$), a cell state ($c_t$), and an output gate ($o_t$).

The hidden layer output can be written as a function of the gates, the model input (here the clinical predictors ($z_t$)), and the previous output of the hidden layer:
\begin{equation}
   h_t = \sigma_h(i_t, f_t, c_t, o_t, z_t, h_{t-1})
\end{equation}
The output of the LSTM network, here the demand forecasts, is calculated as a weighted value of the hidden layer output plus a bias, $b$:
\begin{equation}
    \hat{y}_t = wh_t + b
\end{equation}
Like random forests, LSTM networks are able to capture nonlinear dependencies among the predictors. We implement the LSTM network using the TensorFlow package \citep{abadi2016tensorflow}. The LSTM network is trained by using the ADAM optimizer \citep{kingma2014adam}, and MSE is used as the loss function for this optimizer. For hyperparameter tuning, grid search is performed to find the best model parameters (including the number of epochs, batch size, and number of hidden layers) toward the minimum MSE. Moreover, 10-fold cross-validation is used to reduce overfitting. A function is developed in Python to implement the LSTM network model via a rolling-origin strategy.
\subsection{Rolling Window Analysis}
\label{sec:Rolling}
\begin{figure}[H]
  \centering
  \includegraphics[width=0.9\textwidth, height=0.7\hsize]{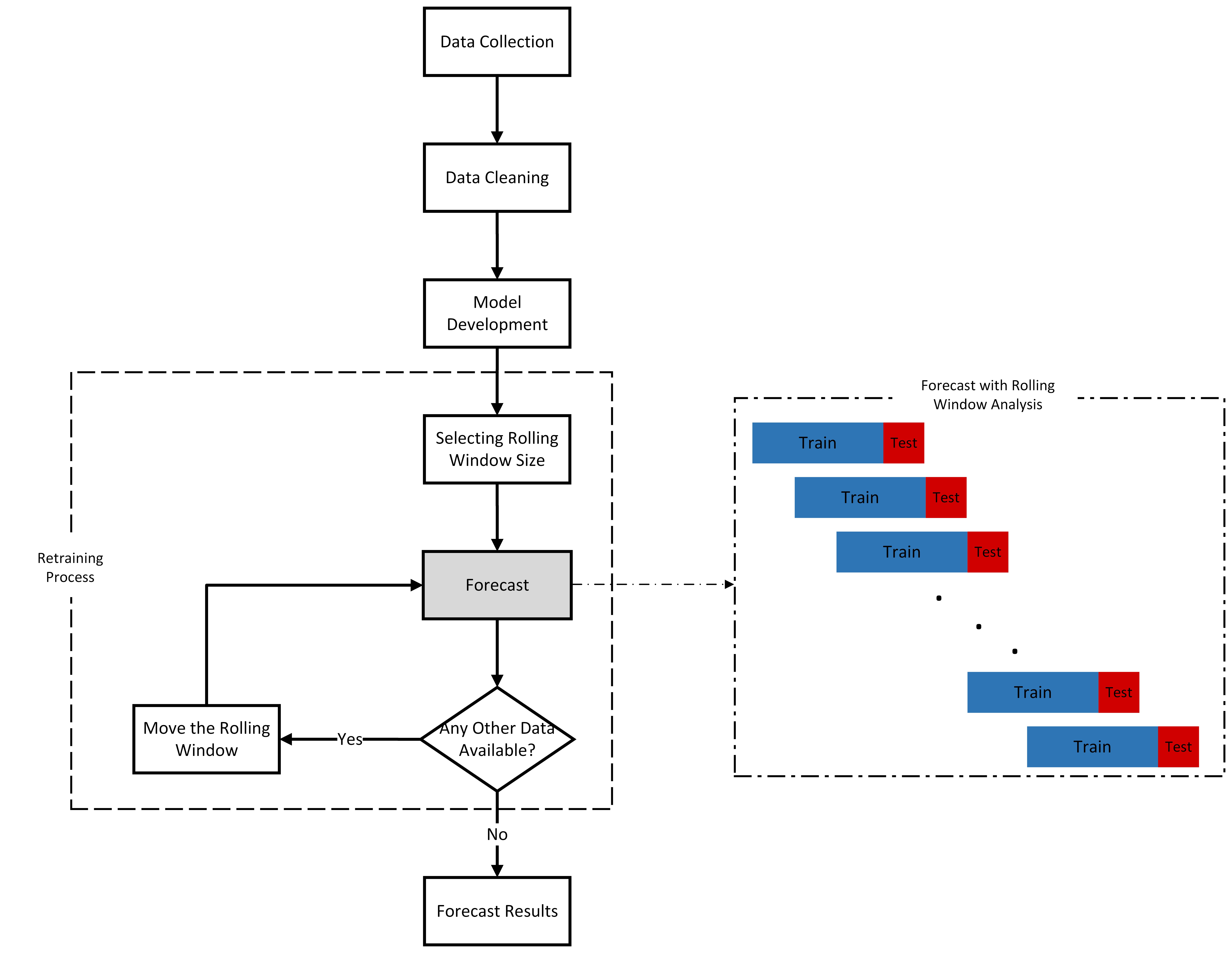}
  \captionsetup{justification=centering}
  \caption{Flowchart of the proposed system}
  \label{fig:Flow}
\end{figure}

We fit the forecasting models multiple times in order to collect multiple out-of-sample one-step ahead forecast errors by using a rolling window. The rolling window is used as part of the demand forecasting process to periodically retrain the models and use more recent data. The flowchart of the proposed demand forecasting system is given in Figure \ref{fig:Flow}. We retrain each model periodically, according to two parameters, the training window and the retraining period. When we retrain a model, we use a training window of the most recent data. For evaluation, we consider a rolling-origin evaluation, similar to the one presented in \citep{tashman2000out}. Many studies consider a fixed-origin evaluation, but we consider a rolling-origin evaluation to improve the efficiency and reliability of out-of-sample tests \citep{tashman2000out}. In a rolling-origin evaluation, the forecasting origin is successively updated and new forecasts are produced from each new origin. We set the forecasting window and rolling steps to be the same as the retraining period.

Here we consider two training windows, two years (starting from 2016) and eight years (starting from 2010), to explore the impact of data volume. The forecasting horizon is one year (2018) in which one-step ahead forecasts are generated for each of the retraining periods. We consider retraining periods of 1, 7, 30, and 90 days, to examine the trade-off between the accuracy and the overhead of retraining. The forecasting accuracy is computed by averaging the Root Mean Square Error (RMSE),  Mean Absolute Error (MAE), Mean Absolute Percentage Error (MAPE), and Symmetric Mean Absolute Percentage Error (SMAPE) over the forecasting window for each rolling origin.

\section{Results}
\label{sec:Res}
This section presents the results of the exploratory analysis for trends, seasonality and holiday patterns. We also present demand forecasting comparisons for univariate and multivariate models, and the forecasting performance of the models trained with training window sizes of two and eight years and retraining periods of 1, 7, 30, and 90 days. We implement the models to forecast the daily demand aggregated over four hospitals for one day ahead via a rolling-origin strategy. We periodically retrain our models based on the rolling window analysis explained in Section \ref{sec:Rolling}.

\subsection{Trends, Seasonality and Holiday Patterns}
\label{sec:Trends}
The data analysis ranges from 2010/01/01 to 2018/12/31. An initial observation is that the demand is highly variable, with a transfused daily average of 17.90 units and a standard deviation of 7.05 units.

\textbf{Observations for non-stationarity:} The results of the ADF test show that the data is not stationary ($P$ value = 0.085) before 2016, but it becomes stationary from 2016 onwards ($P$ value \textless 0.001).

\begin{figure}[H]
  \centerline{\includegraphics[width=0.8\textwidth, height=0.35\hsize]{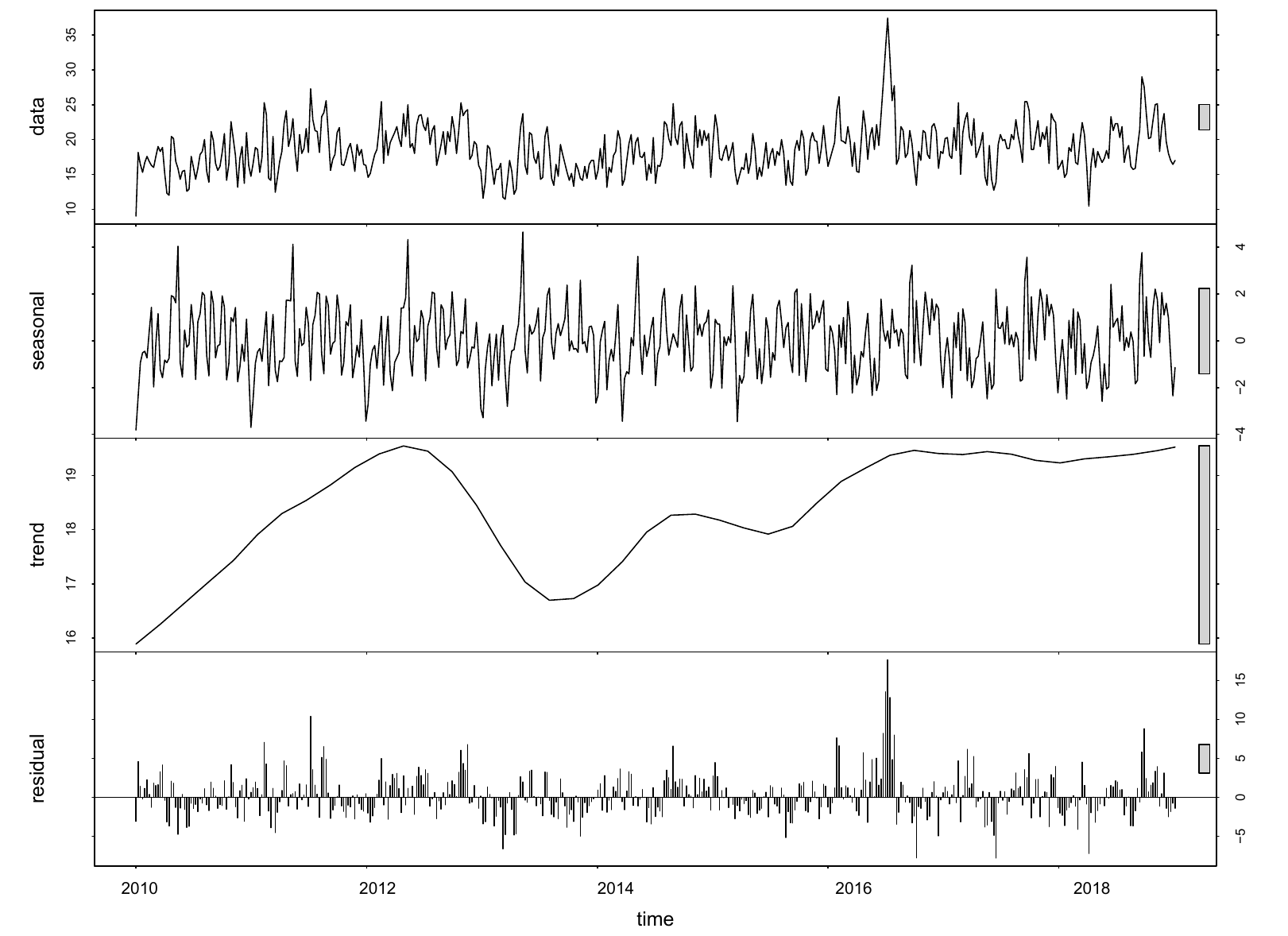}}
  \captionsetup{justification=centering}
  \caption{Time series decomposition using STL method}
  \label{fig:TSD}
\end{figure}

\begin{figure}[H]
  \begin{subfigure}{0.5\linewidth}
  \includegraphics[width=\textwidth]{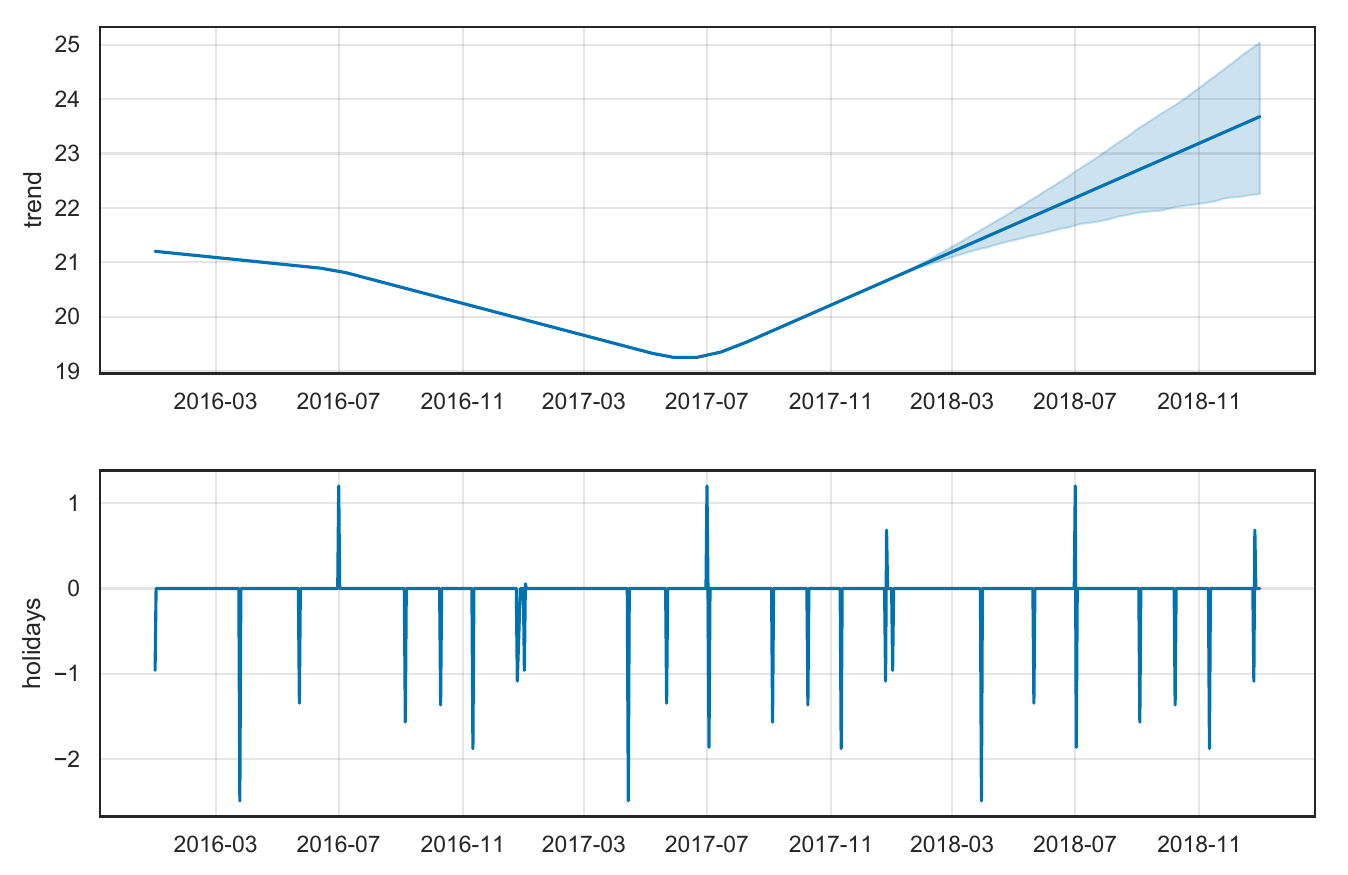}
  \end{subfigure}
  \begin{subfigure}{0.5\linewidth}
  \includegraphics[width=\textwidth]{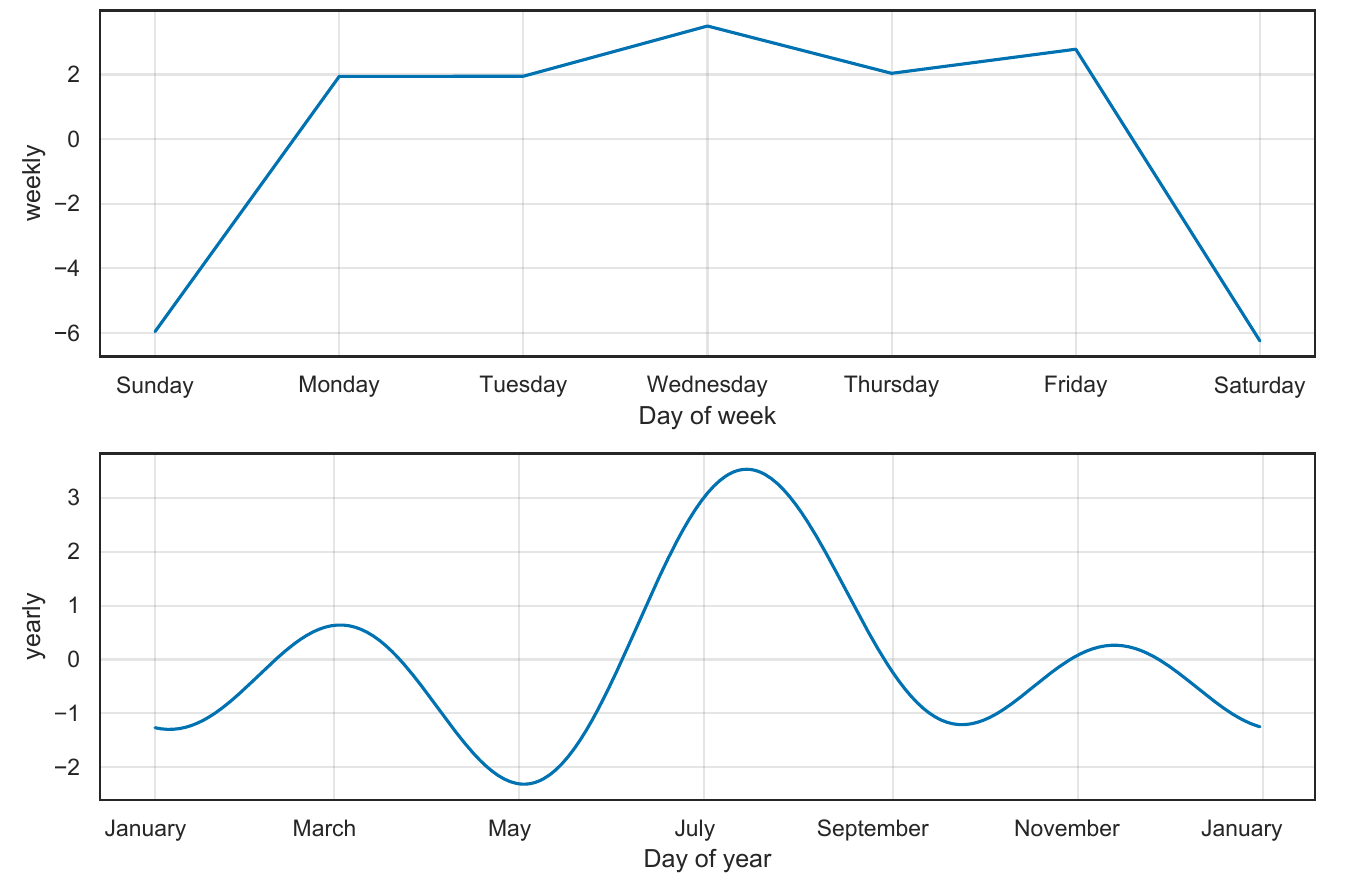}
  \end{subfigure}
  \captionsetup{justification=centering}
  \caption{Prophet model for exploring trends, holiday effects, weekly and yearly seasonality - Since these components are combined through a generalized additive model, the values of y-axes in the plots represent the quantity to be added to or substracted on each specific day}
  \label{fig:Prophet}
\end{figure}

\textbf{Observations for seasonality:} Figure \ref{fig:TSD} shows the time series data decomposition using the STL model. As we can see in the seasonal part, there are recurring temporal patterns in the data. The results of the one-way ANOVA test also show that there is a significant difference between the means of the daily transfusions during weekdays and weekends ($F$ = 5.13, $P$ value \textless 0.001) and the means of daily transfusions in different months ($F$ = 3.94, $P$ value \textless 0.001), which provide strong evidence in favour of the presence of weekly and monthly seasonalities.
Since the data becomes stationary from 2016 onwards, we also explore the trend, holidays, weekly seasonality, and yearly seasonality (seasonality within a year) starting from 2016 using the Prophet model. As we can see from Figure \ref{fig:Prophet}, there is a downward trend from the beginning of 2016 to July 2017 and an upward trend from July 2017 to the end of 2018. Almost all holidays have a negative effect on the model, except for July 1st. This means that the demand is lower than regular weekdays for almost all of the holidays, except for July 1st.
We can also see that there is weekly seasonality in which Wednesdays have the highest platelet usage while the weekends have the lowest usage. Moreover, the yearly seasonality, captured by Fourier series in the Prophet model, depicts three cycles: 1.\ January to May in which March has the highest demand while May has the lowest demand; 2.\ May to September in which the demand is highly variable. July has the highest demand in this cycle and the highest demand of all months while May has the lowest demand in the cycle and also the lowest demand of all months; 3.\ September to January with a slight variation in demand - November with the highest and January with the lowest demands.

\begin{figure}[H]
  \begin{subfigure}{\linewidth}
  \centerline{\includegraphics[width=0.70\hsize, height=0.30\hsize]{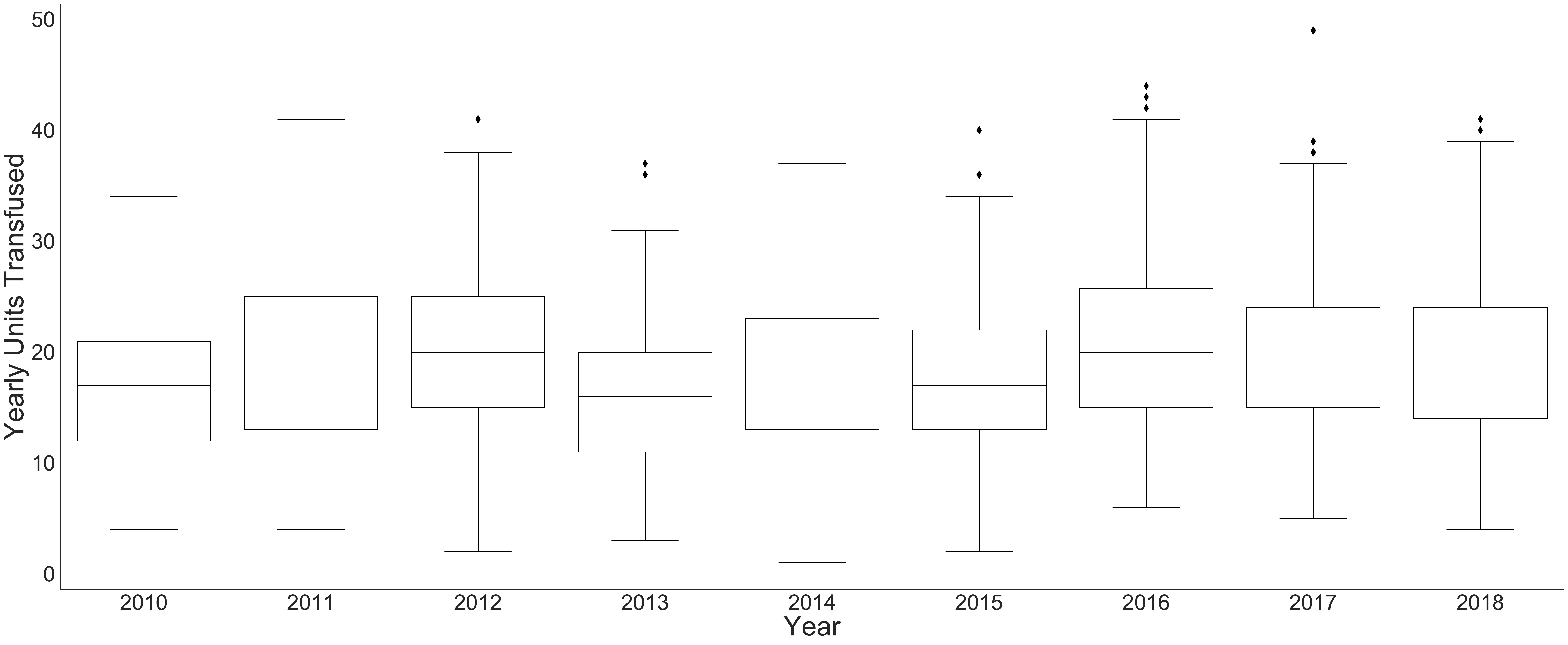}}
  \captionsetup{justification=centering}
  \caption{Mean daily units transfused (year)}
  \label{fig:MeanDailyTR(a)}
  \end{subfigure}
  \begin{subfigure}{0.5\linewidth}
  \centerline{\includegraphics[width=\textwidth, height=0.55\hsize]{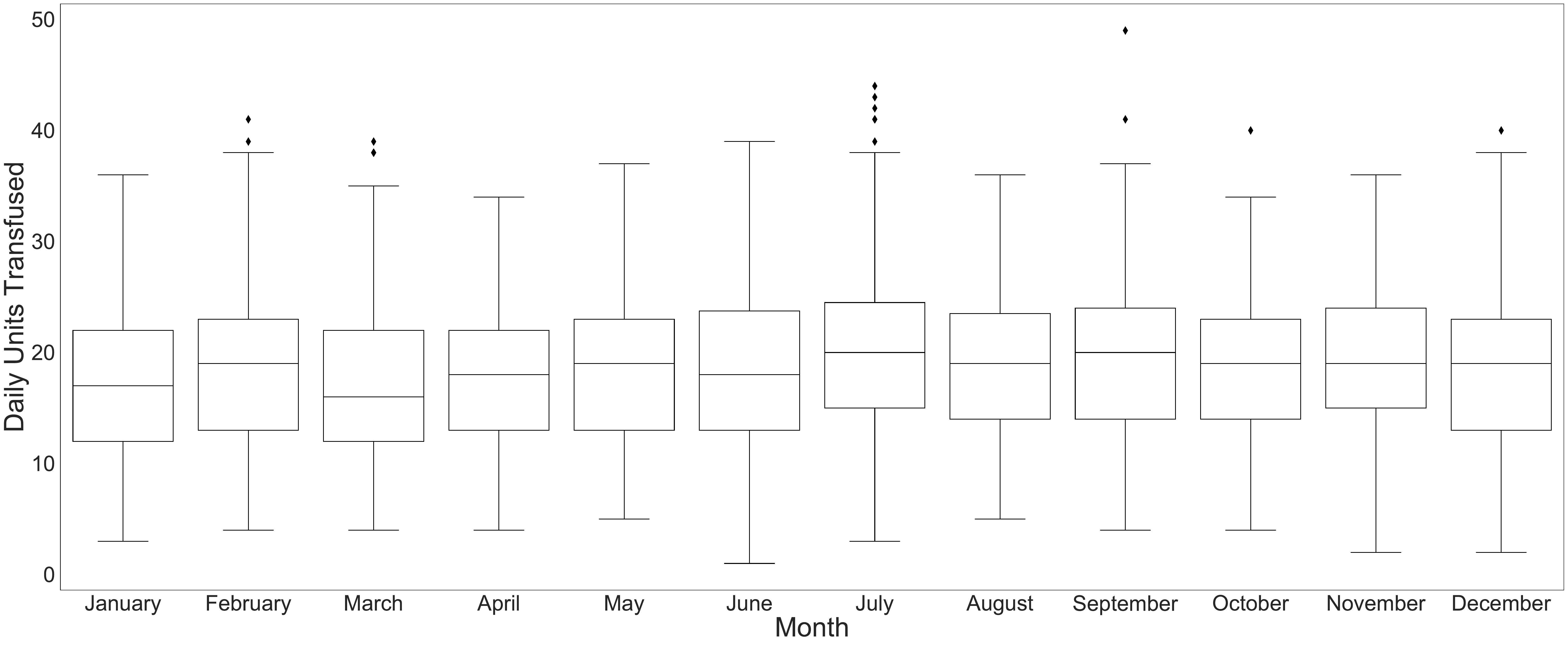}}
  \captionsetup{justification=centering}
  \caption{Mean daily units transfused (month)}
  \label{fig:MeanDailyTR(b)}
  \end{subfigure}%
  \begin{subfigure}{0.5\linewidth}
  \centerline{\includegraphics[width=\textwidth, height=0.55\hsize]{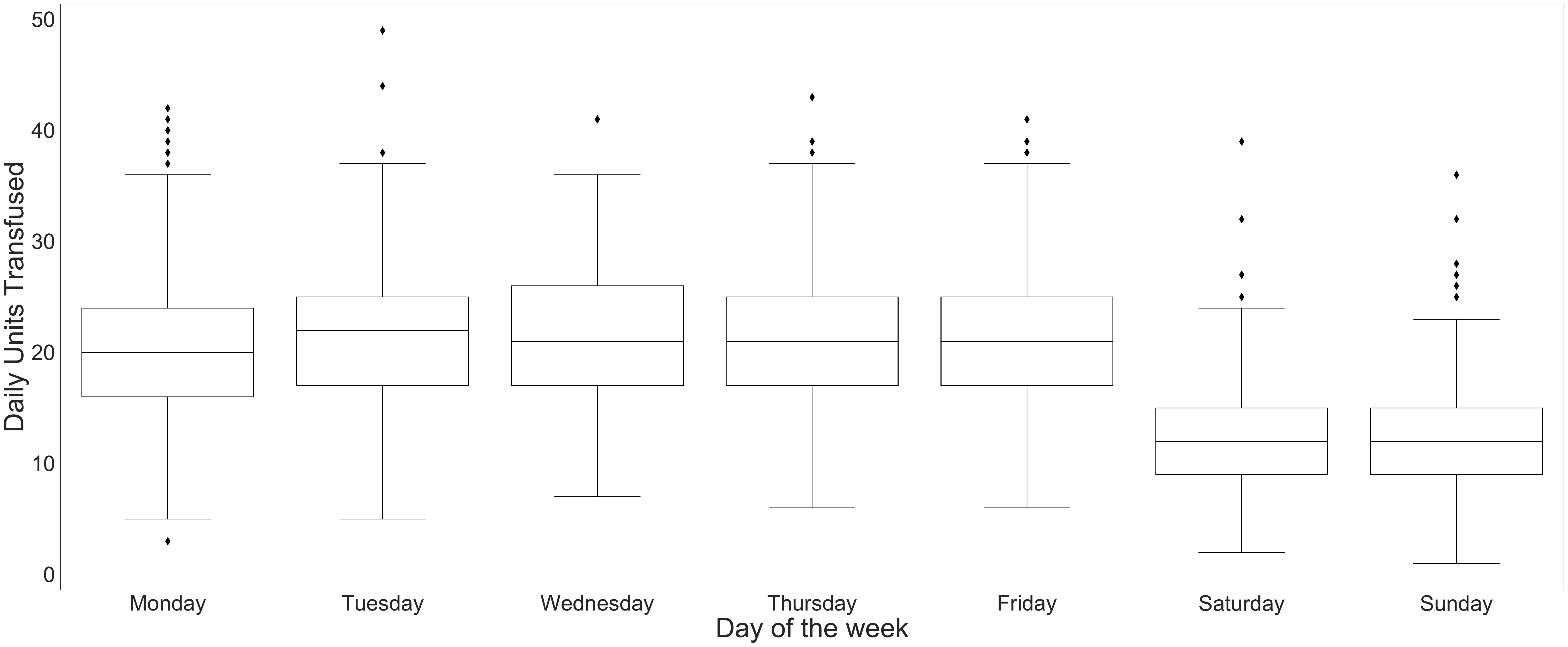}}
  \captionsetup{justification=centering}
  \caption{Mean daily units transfused (day-of-the-week)}
  \label{fig:MeanDailyTR(c)}
  \end{subfigure}
  \captionsetup{justification=centering}
  \caption{Mean daily units transfused}
  \label{fig:MeanDailyTR}
\end{figure}

\textbf{Observations for day of the week effect:} Figure \ref{fig:MeanDailyTR} compares the mean daily units transfused based on day of the week, month, and year. As we can see from Figure \ref{fig:MeanDailyTR}(\subref{fig:MeanDailyTR(c)}), there is a significant difference in the mean daily platelet usage when comparing weekdays to weekends (weekday = mean [sd]: 21.20 [6.22], weekend = mean [sd]: 12.37 [4.60], t-test: 95\% confidence interval for the difference in means: (7.97, 10.34), $P$ value \textless 0.001). Consequently, there is a clear weekday/weekend effect, in agreement with Figure \ref{fig:Prophet}, which appears to be caused by various reasons including lower staffing levels and operating hours over the weekends and prophylactic platelet transfusions to cancer patients on Fridays to ensure that their platelet counts remain sufficiently high over the weekend.

\subsection{Demand Forecasting Comparisons for Univariate Models}
\label{sec:ForecastsUni}
\begin{figure}[H]
  \centering
  \includegraphics[width=0.9\textwidth, height=0.4\hsize]{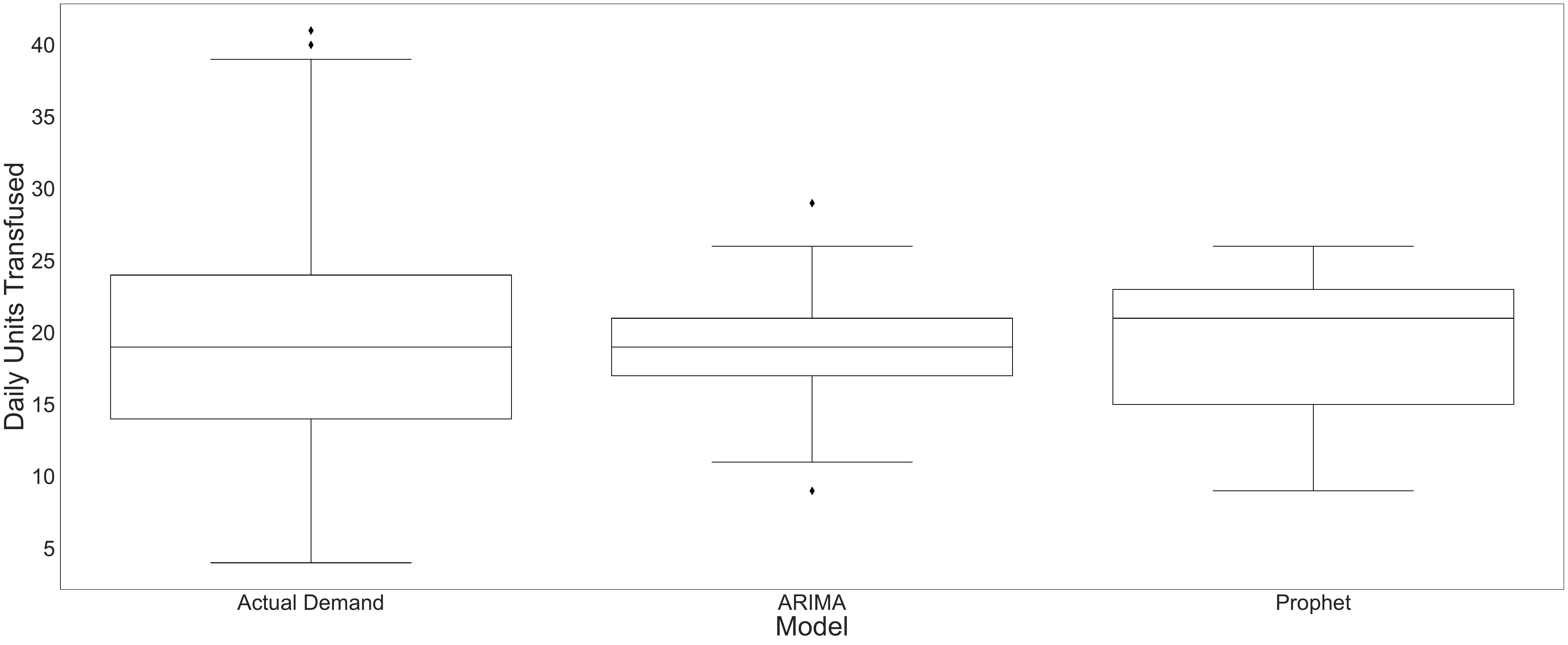}
  \captionsetup{justification=centering}
  \caption{Comparison of the actual demand and the predicted demand from univariate models}
  \label{fig:UniModels}
\end{figure}

Figure \ref{fig:UniModels} compares the forecasts generated by the univariate models, with a training window of two years and by retraining every day, and the actual demand. The actual demand has a large variance (mean [sd]: 19.28 [7.36]). The ARIMA model's forecasts have significantly lower variance (mean [sd]: 18.89 [3.09]) in comparison to the actual demand, meaning that the forecasts cannot capture the wide range of the actual demand. Despite having a larger variance than the ARIMA model, Prophet shows a similar behavior (mean [sd]: 19.35 [4.40]).

\begin{figure}[H]
  \begin{subfigure}{0.5\linewidth}
  \centerline{\includegraphics[width=\textwidth, height=0.60\hsize]{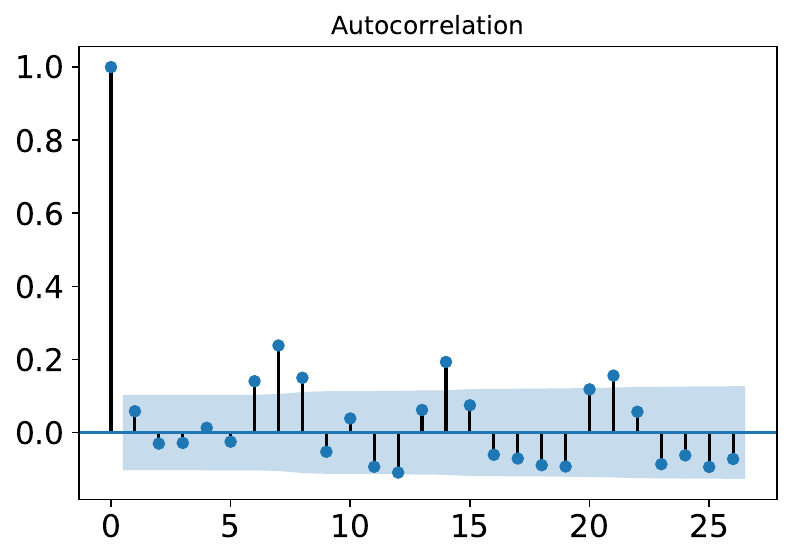}}
  \captionsetup{justification=centering}
  \caption{ACF for ARIMA residuals}
  \label{fig:ACF(a)}
  \end{subfigure}
  \begin{subfigure}{0.5\linewidth}
  \centerline{\includegraphics[width=\textwidth, height=0.60\hsize]{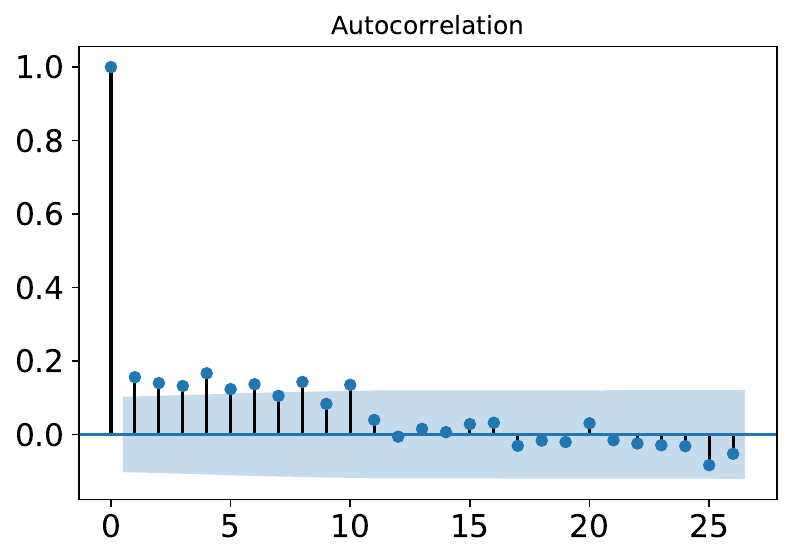}}
  \captionsetup{justification=centering}
  \caption{ACF for Prophet residuals}
  \label{fig:ACF(b)}
  \end{subfigure}%
  \captionsetup{justification=centering}
  \caption{ACF plots for ARIMA and Prophet residuals with a training window of two years and by retraining every day}
  \label{fig:ACF}
\end{figure}

Next, we examine the univariate models' residuals via the ACF (Autocorrelation Function). Figure \ref{fig:ACF} gives the coefficients of correlation between a value and its lag for ARIMA and Prophet. As we can see in Figure \ref{fig:ACF}(\subref{fig:ACF(a)}), there is an autocorrelation at time seven (and multiples of seven) due to weekly seasonality that is not incorporated in the model. Since seasonality is one of the primary features of our time series data, we include seasonality directly in the forecasting process by using the Prophet model. As we can see in Figure \ref{fig:ACF}(\subref{fig:ACF(b)}), there is no repeated autocorrelation pattern for Prophet residuals.

We also perform a pairwise t-test to compare the the univariate models' residuals with each other. The results show a statistically significant difference between the ARIMA residuals (mean [sd]: 0.39 [6.80]) and Prophet residuals (mean [sd]: -0.07 [5.90], t-test: 95\% confidence interval for the difference in means: (0.08, 0.85), $P$ value = 0.018), indicating higher residuals in the ARIMA model.


\subsection{Demand Forecasting Comparisons for Multivariate Models}
\label{sec:ForecastsMulti}
We begin this section with an examination of selecting the clinical predictors for the multivariate models. Next, we compare the forecasts generated by the multivariate models and the actual demand.

\subsubsection{Selecting the predictors using Lasso Regression}
\label{sec:CI}
As discussed in Section \ref{sec:DataDesc}, the data has more than 100 features, and we select predictors via lasso regression. The 29 clinical predictors that are introduced in Section \ref{sec:DataDesc} are selected by lasso regression and used for training the multivariate models. One of the data characteristics is that clinical predictors are highly correlated. These high correlations can affect the performance of a regression model, mainly because of the violation of model assumptions.

\begin{figure}[H]
  \centering
  \includegraphics[width=1\textwidth, height=0.85\hsize]{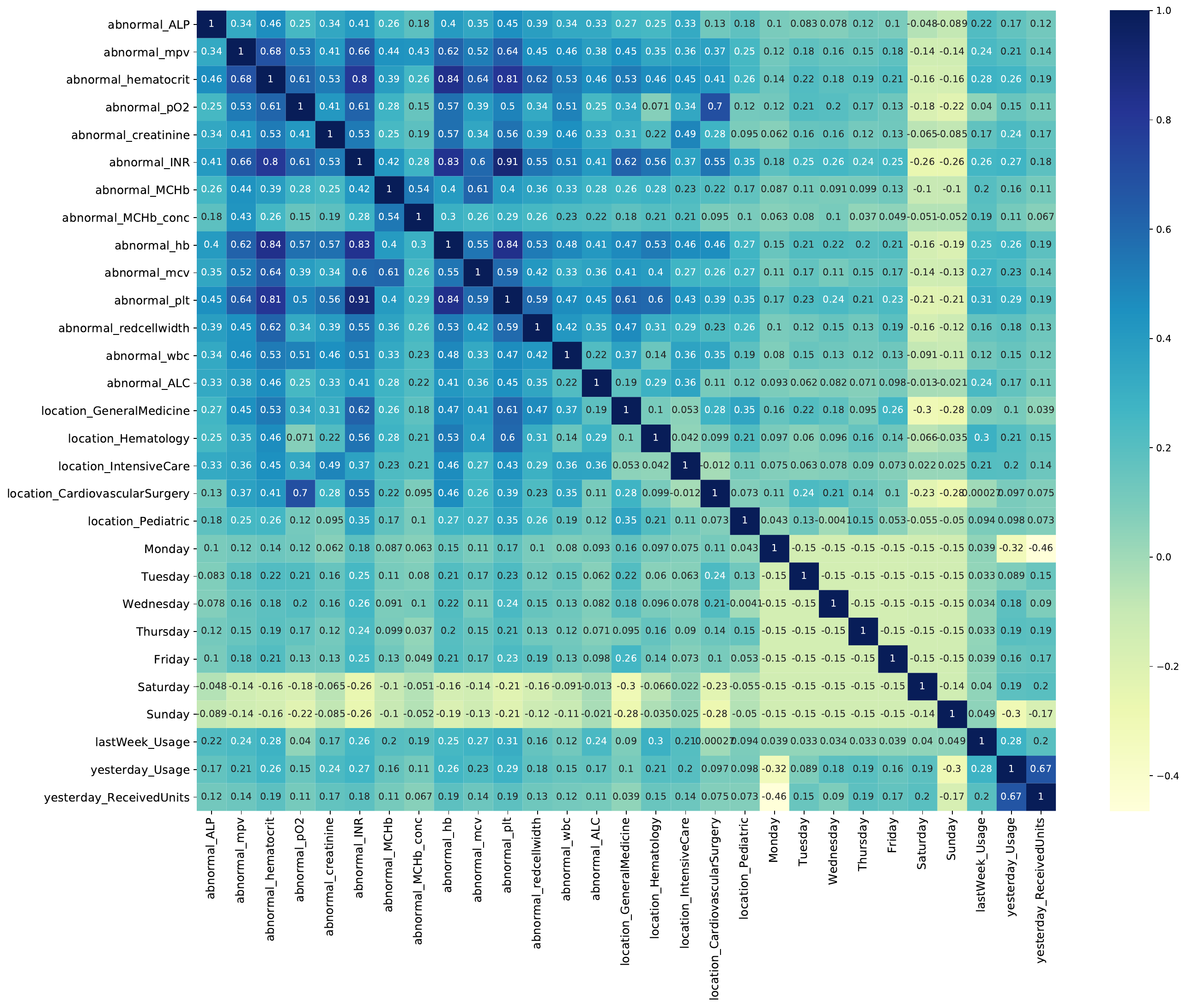}
  \captionsetup{justification=centering}
   \caption{Pearson correlation among variables}
  \label{fig:Corr}
\end{figure}

We calculate the Pearson correlation between the selected predictors. The Pearson correlation measures the linear relationship between two variables, ranging from -1 to 1, where -1 corresponds to a perfect negative correlation and 1 corresponds to a perfect positive correlation. As shown in Figure \ref{fig:Corr}, the predictors, in particular the daily numbers of patients with abnormal laboratory test results, are highly correlated. These high correlations give rise to some challenges when the predictors are considered in the demand forecasting process, as discussed in Table \ref{tab:CovLasso} of Appendix A.

\begin{figure}[H]
  \centering
  \includegraphics[width=0.9\textwidth, height=0.4\hsize]{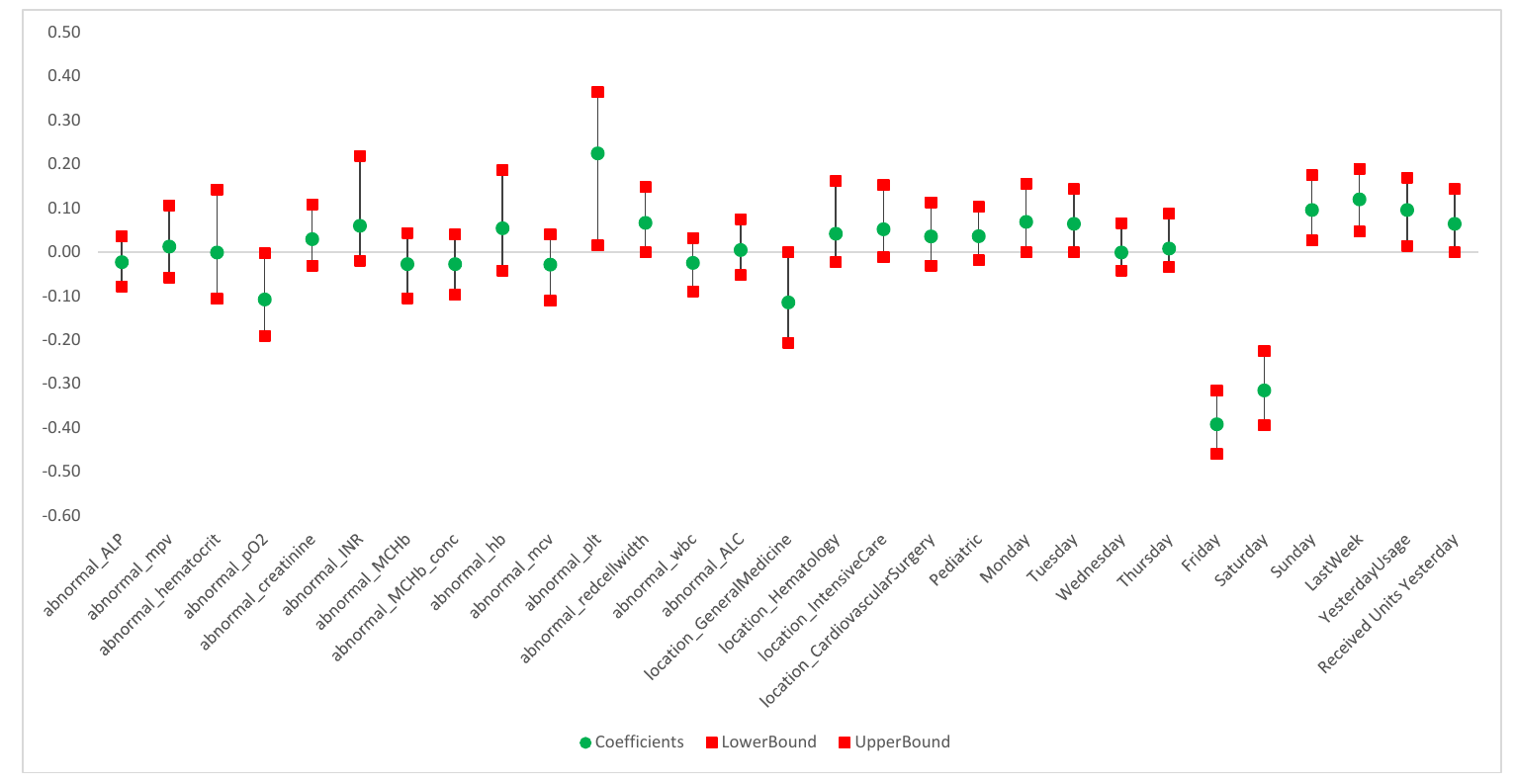}
  \captionsetup{justification=centering}
  \caption{Confidence interval for predictors' coefficients - Lasso regression}
  \label{fig:Lasso_Wght}
\end{figure}

We also calculate the confidence intervals for these clinical predictors (also referred to as the model predictors). There are multiple methods for calculating a confidence interval for the predictors; one of the most popular is the bootstrap method \citep{efron1994introduction}. The bootstrap method is used in the experiments for calculating the confidence intervals for the predictors used in the multivariate models. As shown in Figure \ref{fig:Lasso_Wght}, the predictors' coefficients have a wide range, so we see high values (abnormal\_plt = 0.23) as well as low values (Friday = -0.39) for the lab tests and day of the week. Overall, we can see that the range of the predictors' coefficients for the 95\% confidence interval is narrow. Detailed information about the predictors and their corresponding coefficients are given in Table \ref{tab:CovLasso} of Appendix A.

\subsubsection{Comparisons of Multivariate Models forecasts}
\label{sec:MultiComp}
\begin{figure}[H]
  \centering
  \includegraphics[width=0.9\textwidth, height=0.4\hsize]{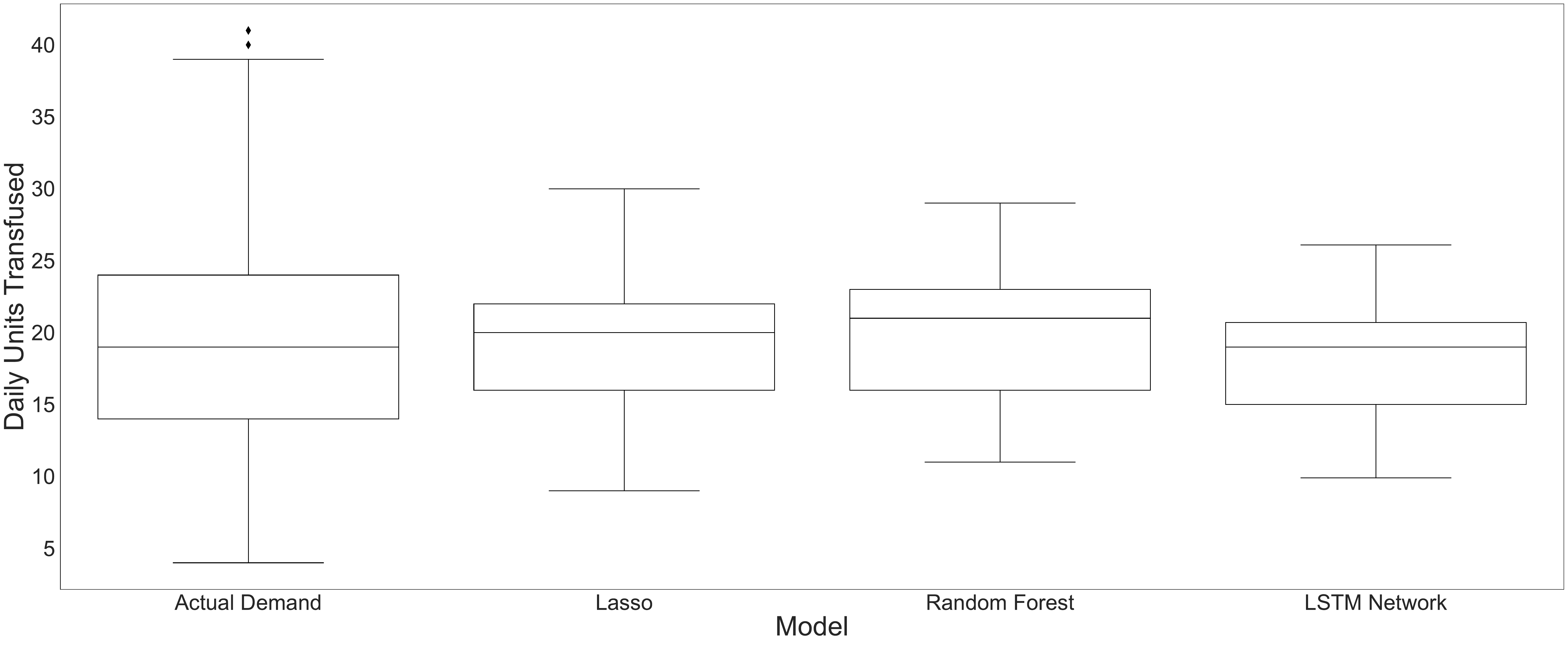}
  \captionsetup{justification=centering}
  \caption{Comparison of the actual demand and the predicted demand from multivariate models}
  \label{fig:MultiModels}
\end{figure}

Figure \ref{fig:MultiModels} shows the actual daily units transfused and the forecasts generated by the multivariate models, lasso regression, random forest and LSTM network, with a training window of two years and by retraining every day. The forecast means of lasso regression (mean [sd]: 19.12 [3.62]) and random forest (mean [sd]: 19.72 [4.28]) are very close to the actual mean demand, but forecast standard deviations are much lower than the actual demand standard variation. LSTM network forecasts have a slightly lower mean (mean [sd]: 18.01 [3.55]) but significantly lower standard deviation than the actual demand.
Next, a repeated measures ANOVA test is performed for comparing the multivariate models' residuals with each other. The results of the test show a statistically significant difference between the lasso regression, random forest, and LSTM Network residuals ($F$ = 35.86, $P$ value \textless 0.001). To show which models' residuals are significantly different, we perform pairwise comparisons by using a pairwise t-test. Table \ref{tab:t-test} gives the results of the pairwise t-test for the models' residuals, showing that they are significantly different from each other. The $P$ values are adjusted using the Bonferroni multiple testing correction method.

\begin{table}[H]
\captionsetup{justification=centering}
\caption{Comparison of multivariate models residuals using a pairwise t-test}
\scalebox{0.9}{
\begin{tabular}{|c|cc|ccc|}
\hline
\multirow{2}{*}{\textbf{Model}} & \multicolumn{2}{c|}{\textbf{Descriptive statistics}}             & \multicolumn{3}{c|}{\textbf{T-test}}                                                                                                                                                   \\ \cline{2-6} 
                                & \multicolumn{1}{c|}{\textbf{Mean}} & \textbf{Standard Deviation} & \multicolumn{1}{c|}{\textbf{Model}} & \multicolumn{1}{c|}{\textbf{\begin{tabular}[c]{@{}c@{}}95\% confidence interval \\ for the difference in means\end{tabular}}} & \textbf{$P$ value} \\ \hline
\textbf{Lasso Regression}                  & \multicolumn{1}{c|}{0.16}          & 6.39                        & \multicolumn{1}{c|}{Random Forest}  & \multicolumn{1}{c|}{(0.09, 1.12)}                                                                                            & 0.020            \\ \hline
\textbf{Random Forest}          & \multicolumn{1}{c|}{-0.44}         & 8.77                        & \multicolumn{1}{c|}{LSTM Network}   & \multicolumn{1}{c|}{(1.60, 1.83)}                                                                                         & \textless 0.001  \\ \hline
\textbf{LSTM Network}           & \multicolumn{1}{c|}{1.27}          & 8.34                        & \multicolumn{1}{c|}{Lasso Regression}          & \multicolumn{1}{c|}{(-1.57, -0.65)}                                                                                           & \textless 0.001  \\ \hline
\end{tabular}
}
\label{tab:t-test}
\end{table}

\subsection{Performance Comparisons}
\label{Sec:PerfMeasure}
\begin{figure}[H]
\begin{subfigure}{0.5\linewidth}
\centerline{\includegraphics[width=\textwidth, height=0.55\hsize]{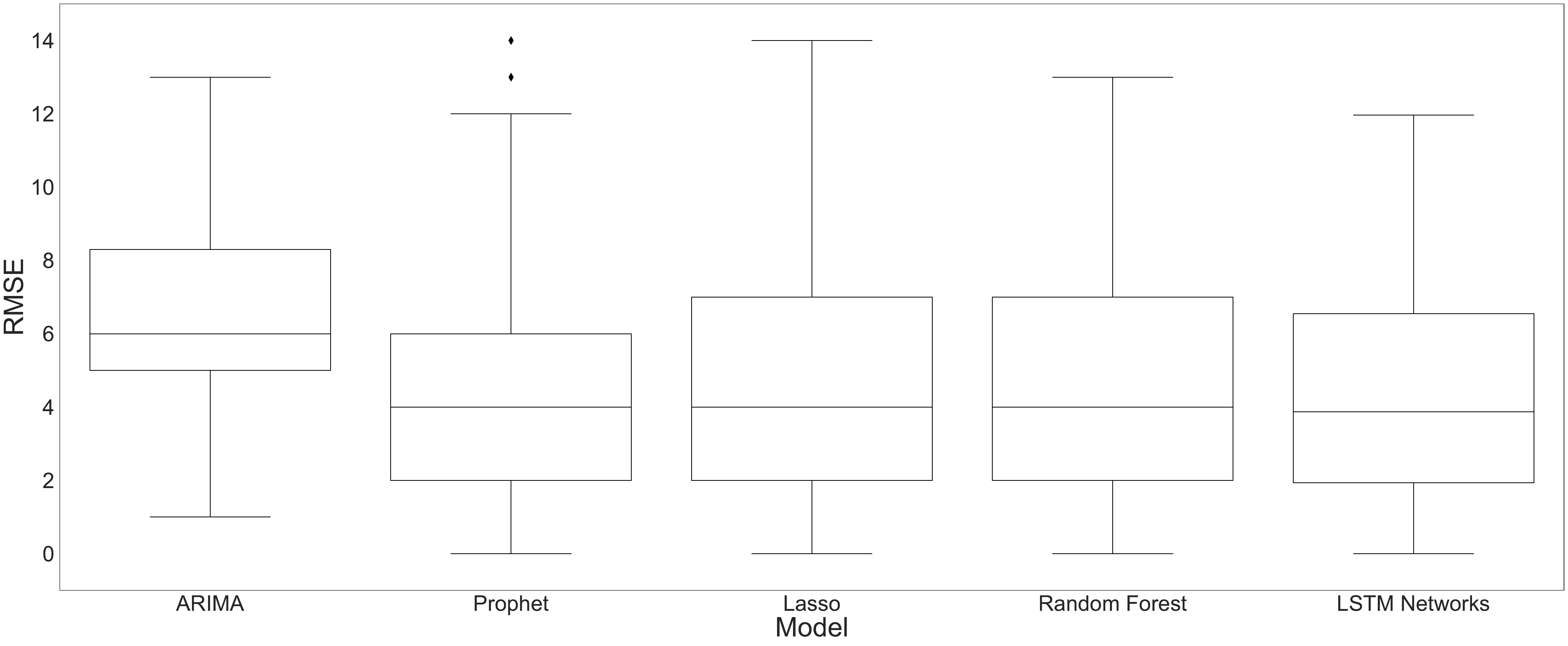}}
\captionsetup{justification=centering}
\caption{2 years rolling window, retraining every day}
\label{fig:RMSE(a)}
\end{subfigure}%
\begin{subfigure}{0.5\linewidth}
\centerline{\includegraphics[width=\textwidth, height=0.55\hsize]{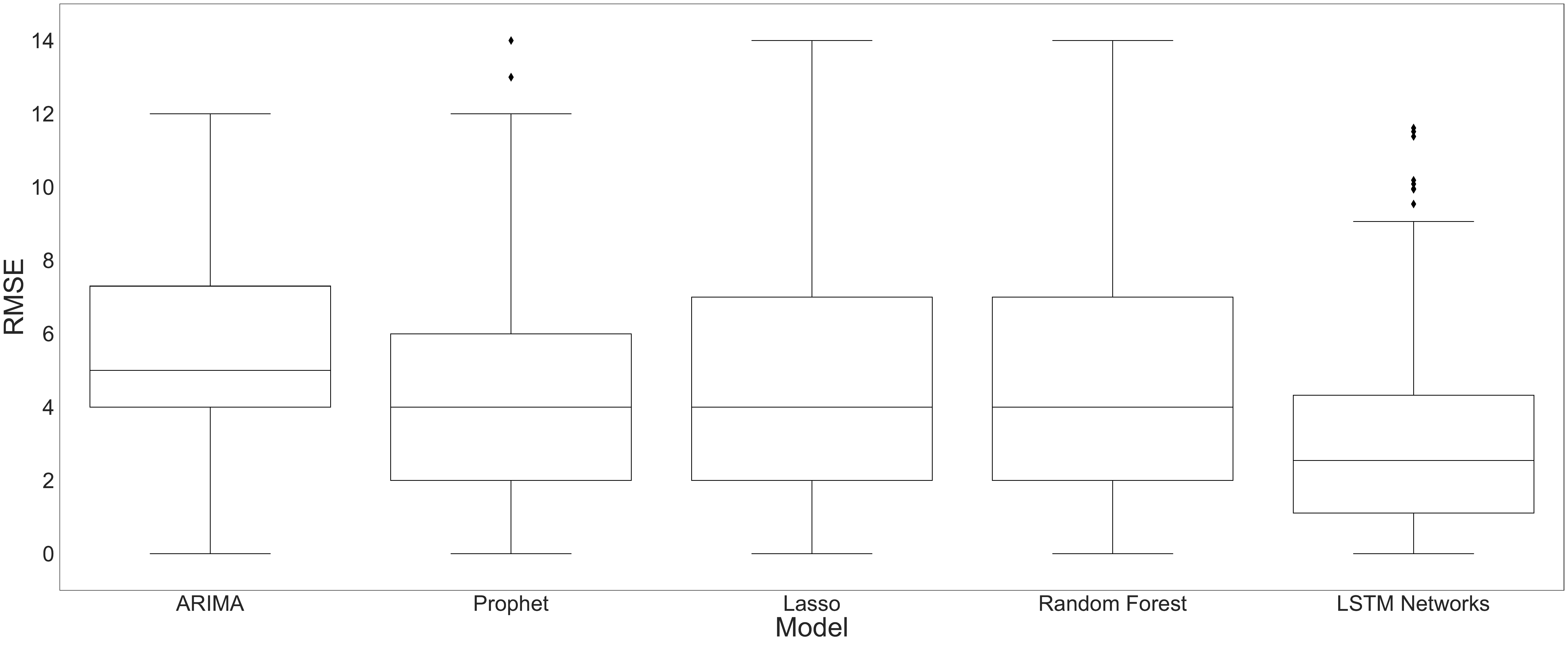}}
\captionsetup{justification=centering}
\caption{8 years rolling window, retraining every day}
\end{subfigure}
\begin{subfigure}{0.5\linewidth}
\centerline{\includegraphics[width=\textwidth, height=0.55\hsize]{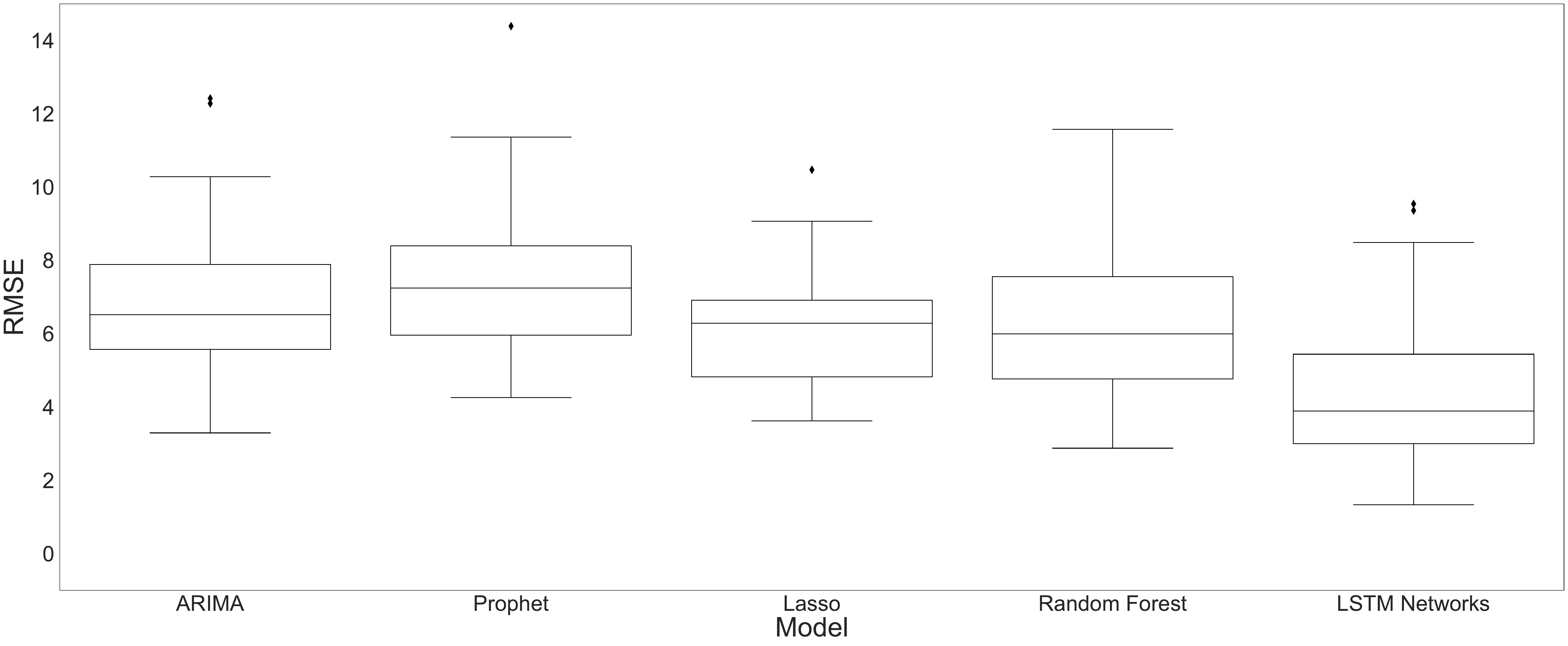}}
\caption{2 years rolling window, retraining every 7 days}
\end{subfigure}
\begin{subfigure}{0.5\linewidth}
\centerline{\includegraphics[width=\textwidth, height=0.55\hsize]{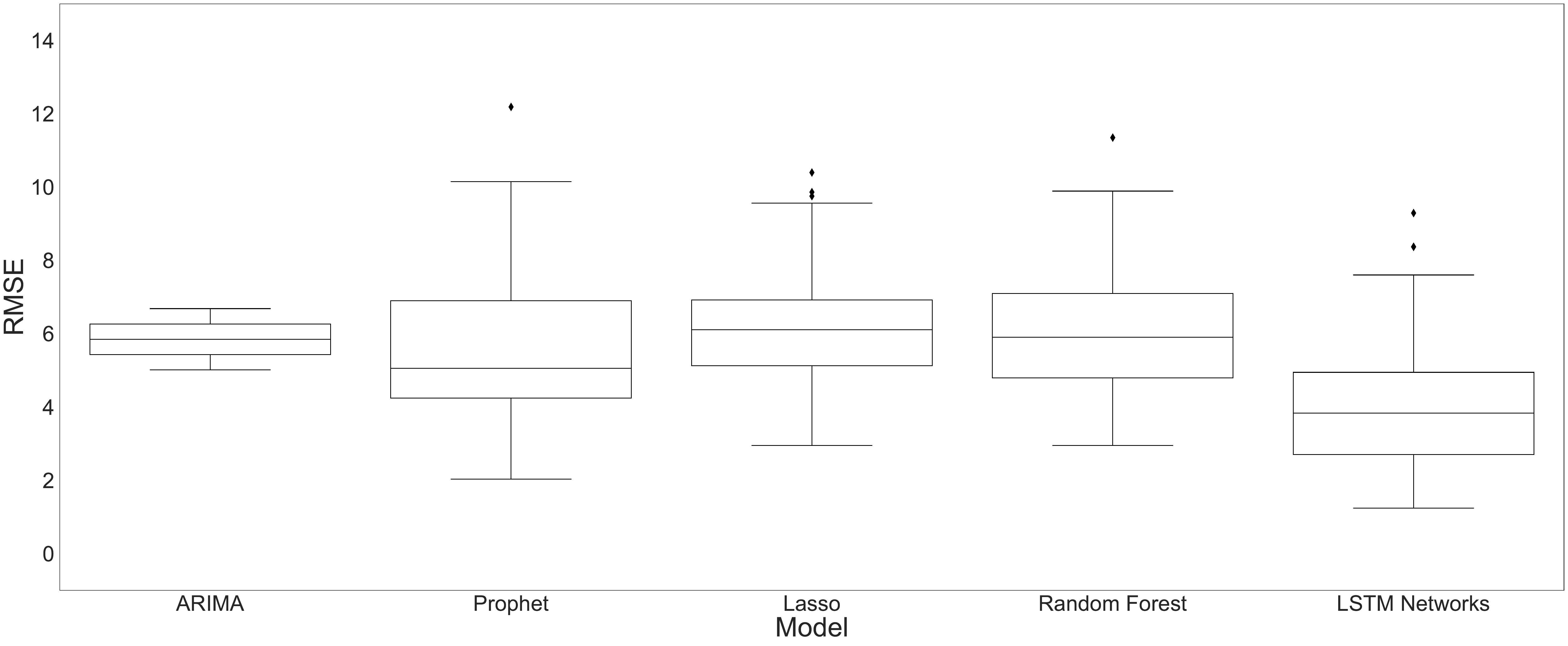}}
\captionsetup{justification=centering}
\caption{8 years rolling window, retraining every 7 days}
\label{f124}
\end{subfigure}
\begin{subfigure}{0.5\linewidth}
\centerline{\includegraphics[width=\textwidth, height=0.55\hsize]{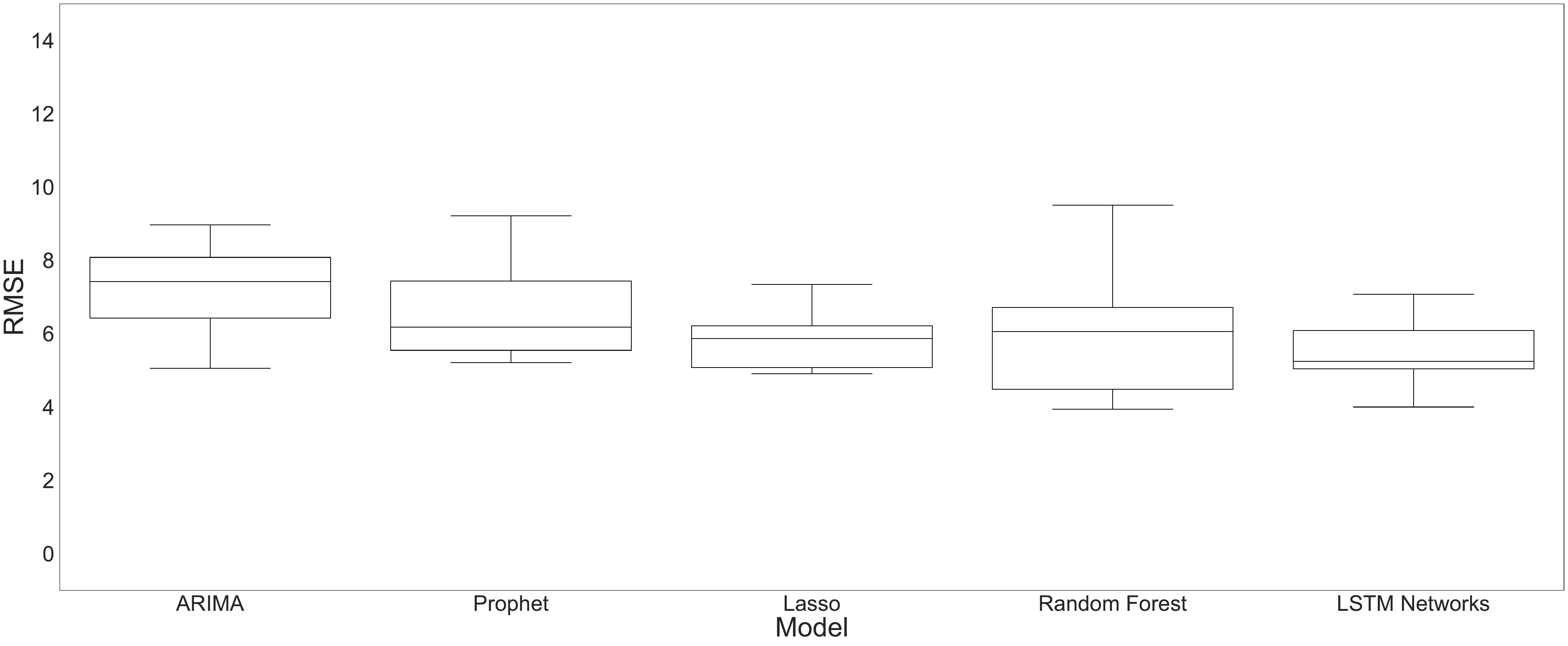}}
\captionsetup{justification=centering}
\caption{2 years rolling window, retraining every 30 days}
\end{subfigure}
\begin{subfigure}{0.5\linewidth}
\centerline{\includegraphics[width=\textwidth, height=0.55\hsize]{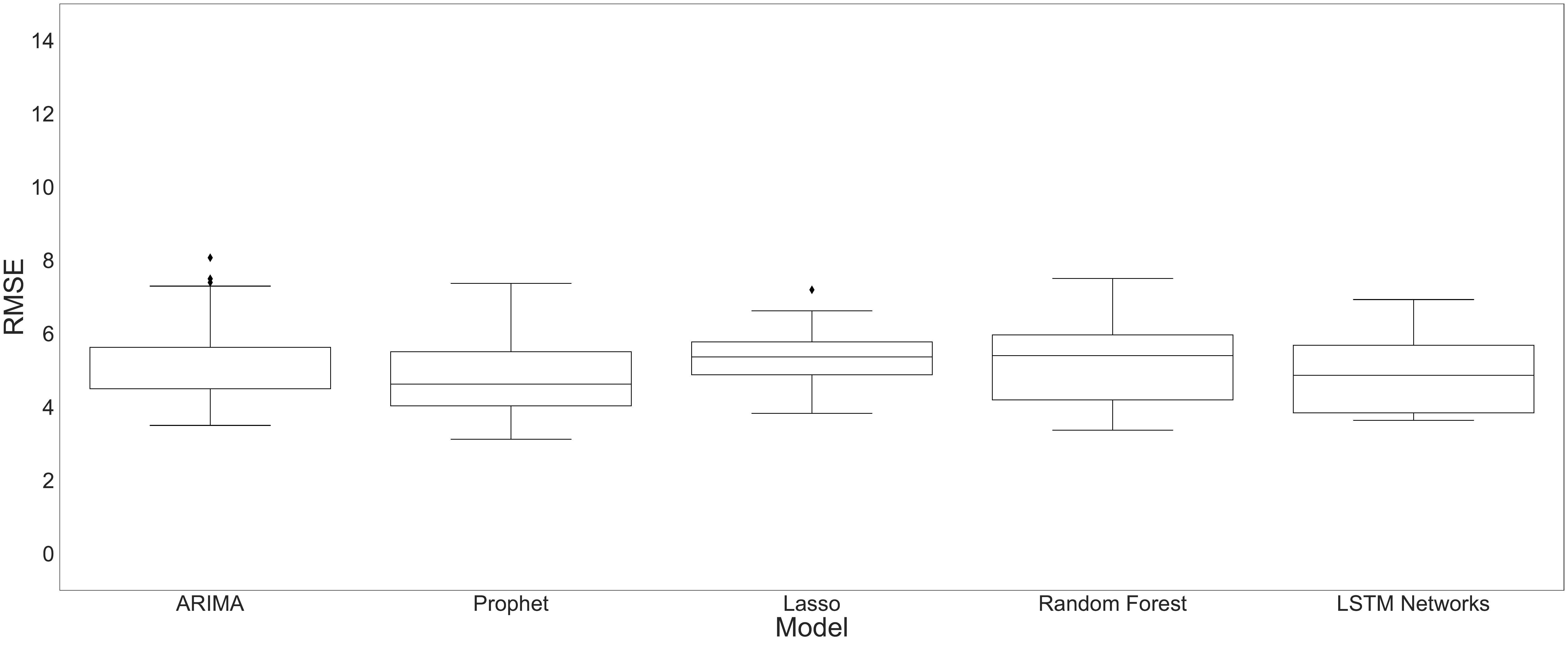}}
\captionsetup{justification=centering}
\caption{8 years rolling window, retraining every 30 days}
\end{subfigure}
\begin{subfigure}{0.5\linewidth}
\centerline{\includegraphics[width=\textwidth, height=0.55\hsize]{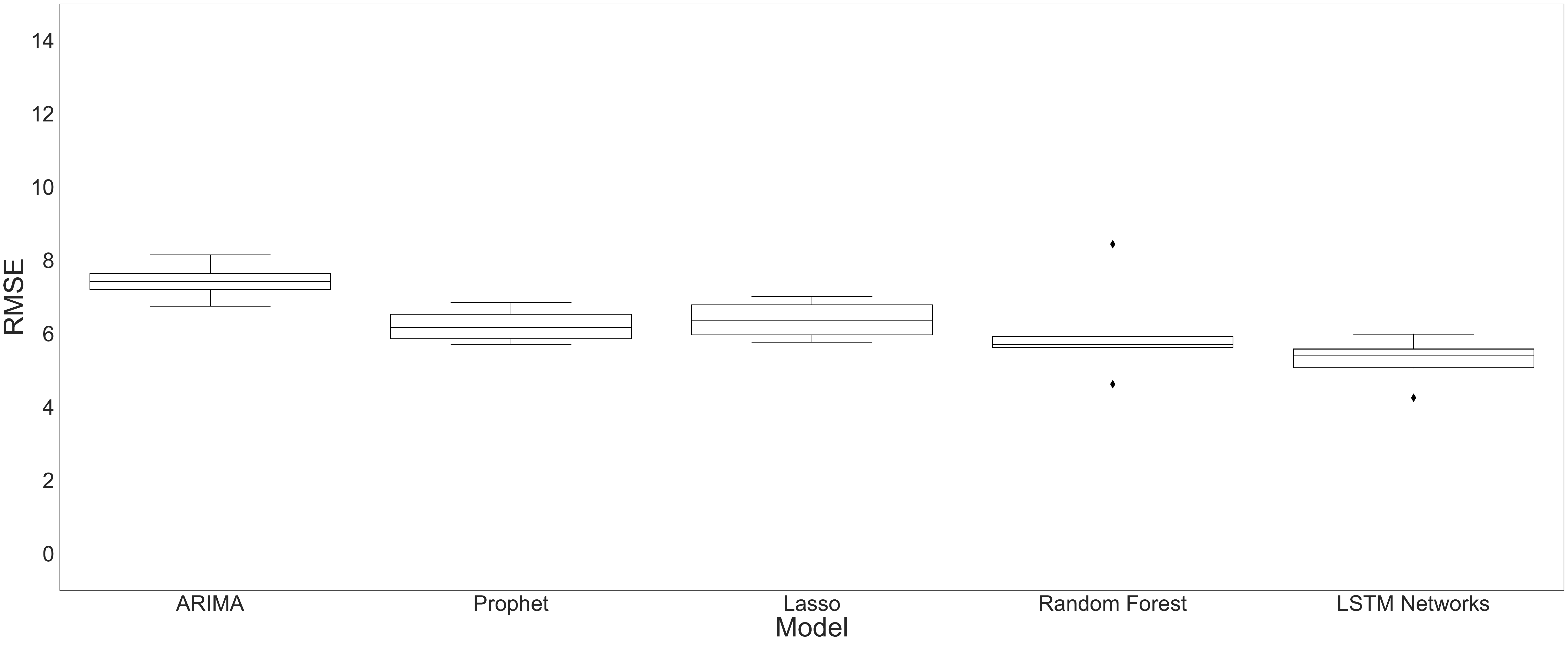}}
\captionsetup{justification=centering}
\caption{2 years rolling window, retraining every 90 days}
\label{fig:RMSE(g)}
\end{subfigure}
\begin{subfigure}{0.5\linewidth}
\centerline{\includegraphics[width=\textwidth, height=0.55\hsize]{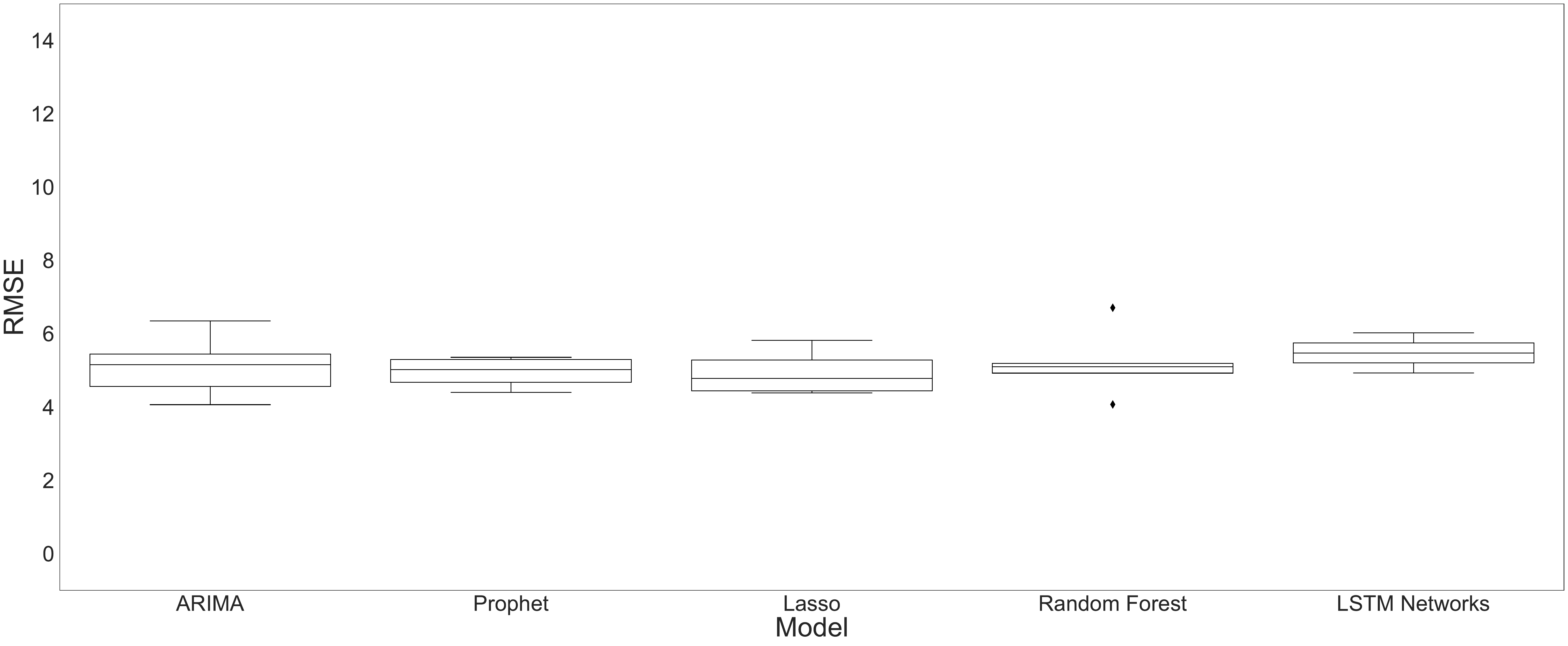}}
\captionsetup{justification=centering}
\caption{8 years rolling window, retraining every 90 days}
\end{subfigure}
\captionsetup{justification=centering}
\caption{RMSE with different training window sizes and retraining periods}
\label{fig:RMSE}
\end{figure}

\begin{figure}[H]
\begin{subfigure}{0.5\linewidth}
\centerline{\includegraphics[width=\textwidth, height=0.55\hsize]{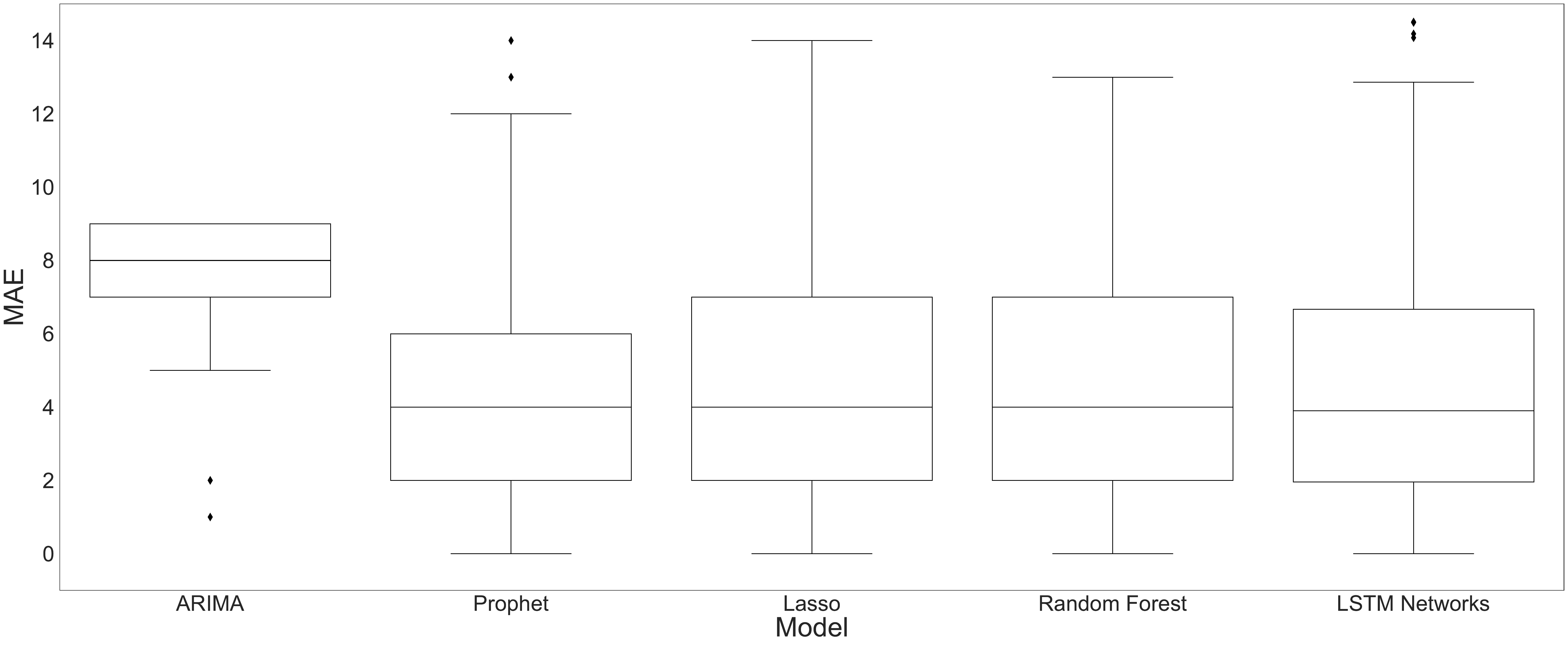}}
\captionsetup{justification=centering}
\caption{2 years rolling window, retraining every day}
\label{fig:MAE(a)}
\end{subfigure}%
\begin{subfigure}{0.5\linewidth}
\centerline{\includegraphics[width=\textwidth, height=0.55\hsize]{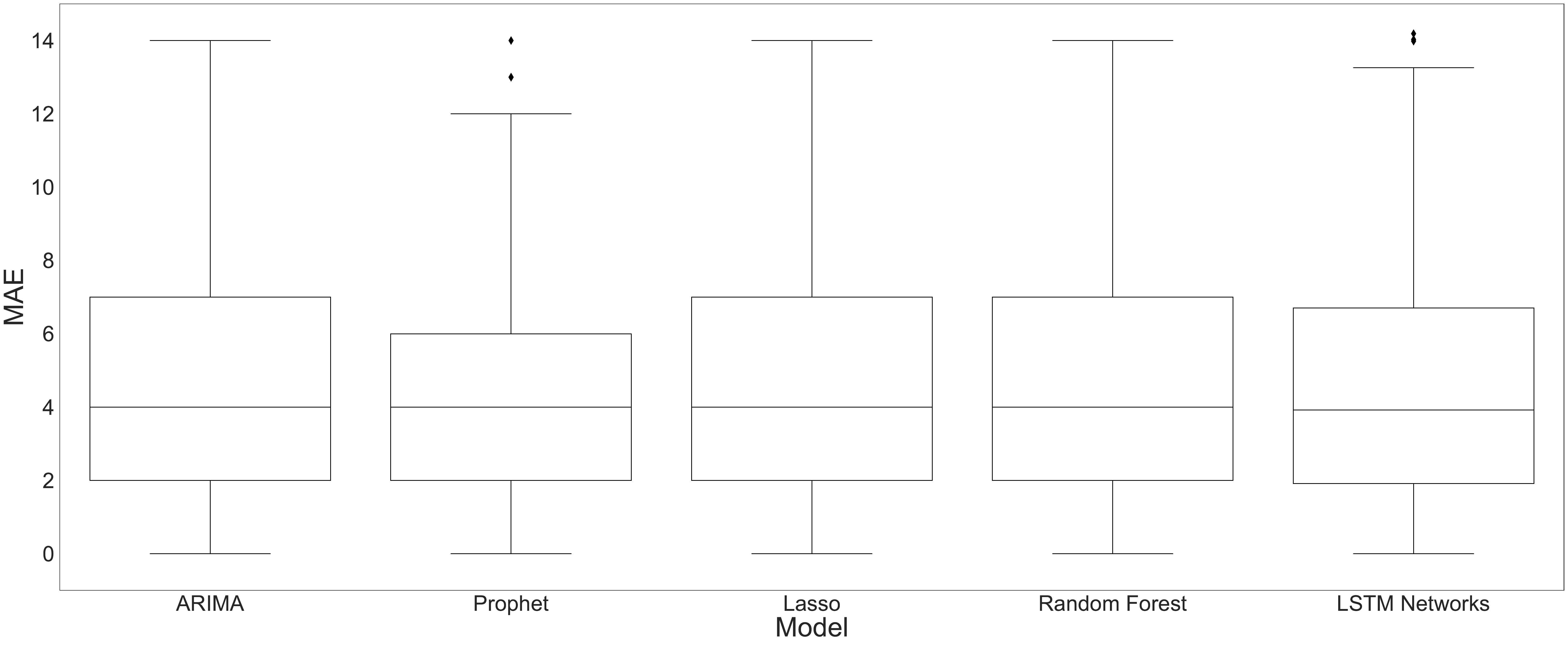}}
\captionsetup{justification=centering}
\caption{8 years rolling window, retraining every day}
\end{subfigure}
\begin{subfigure}{0.5\linewidth}
\centerline{\includegraphics[width=\textwidth, height=0.55\hsize]{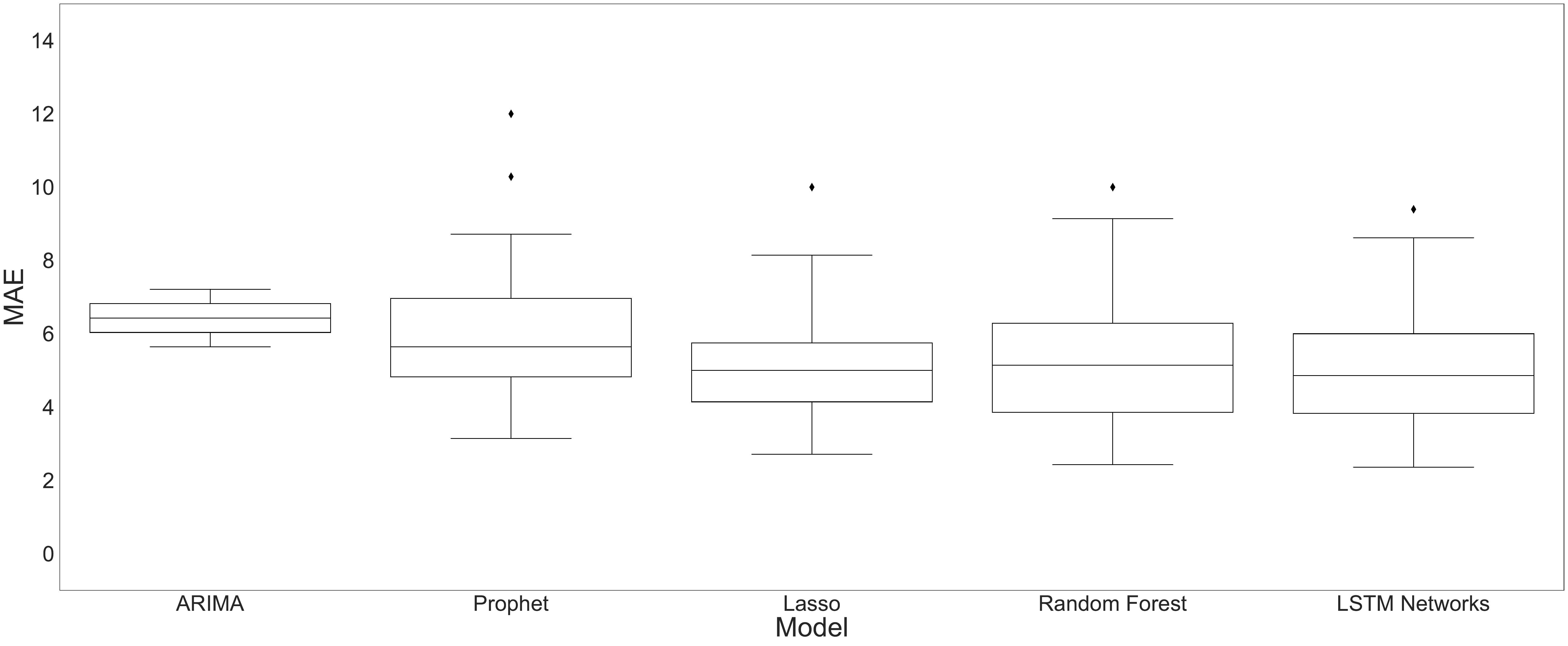}}
\caption{2 years rolling window, retraining every 7 days}
\end{subfigure}
\begin{subfigure}{0.5\linewidth}
\centerline{\includegraphics[width=\textwidth, height=0.55\hsize]{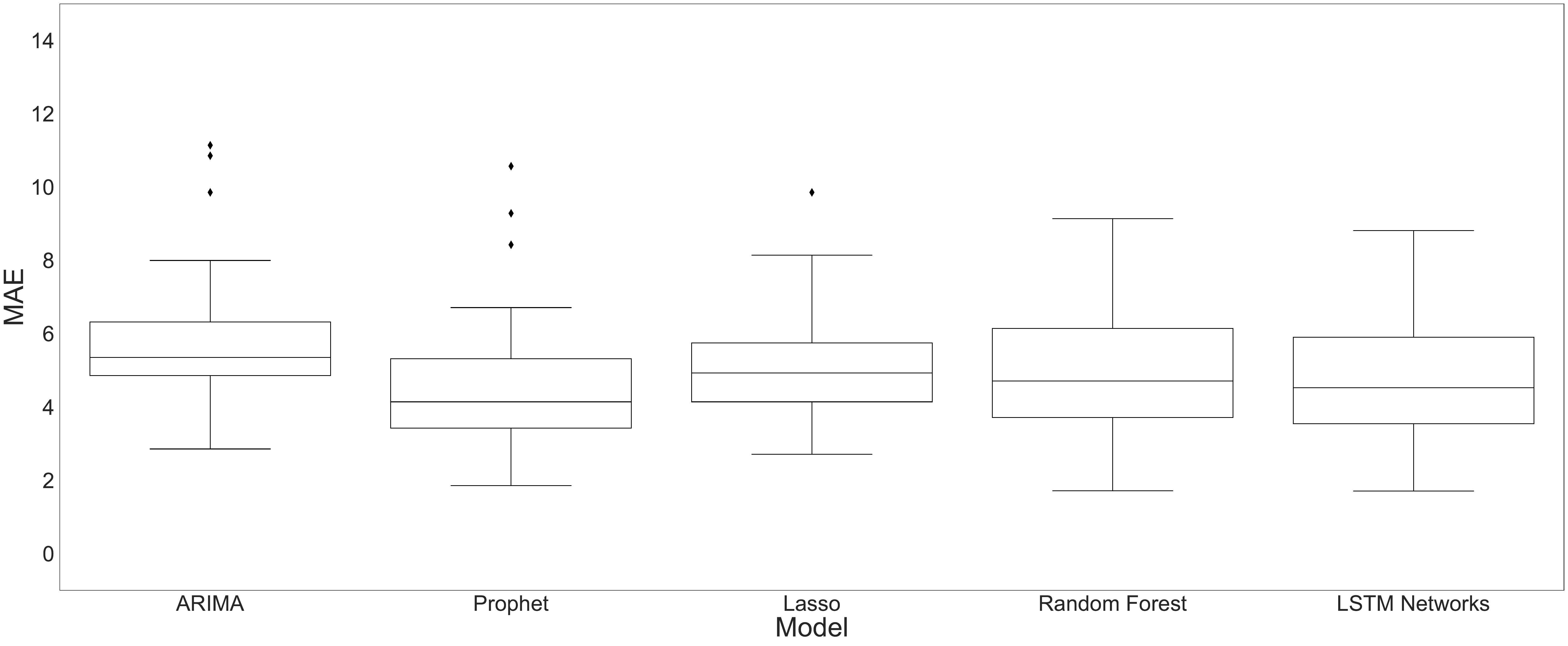}}
\captionsetup{justification=centering}
\caption{8 years rolling window, retraining every 7 days}
\label{f124}
\end{subfigure}
\begin{subfigure}{0.5\linewidth}
\centerline{\includegraphics[width=\textwidth, height=0.55\hsize]{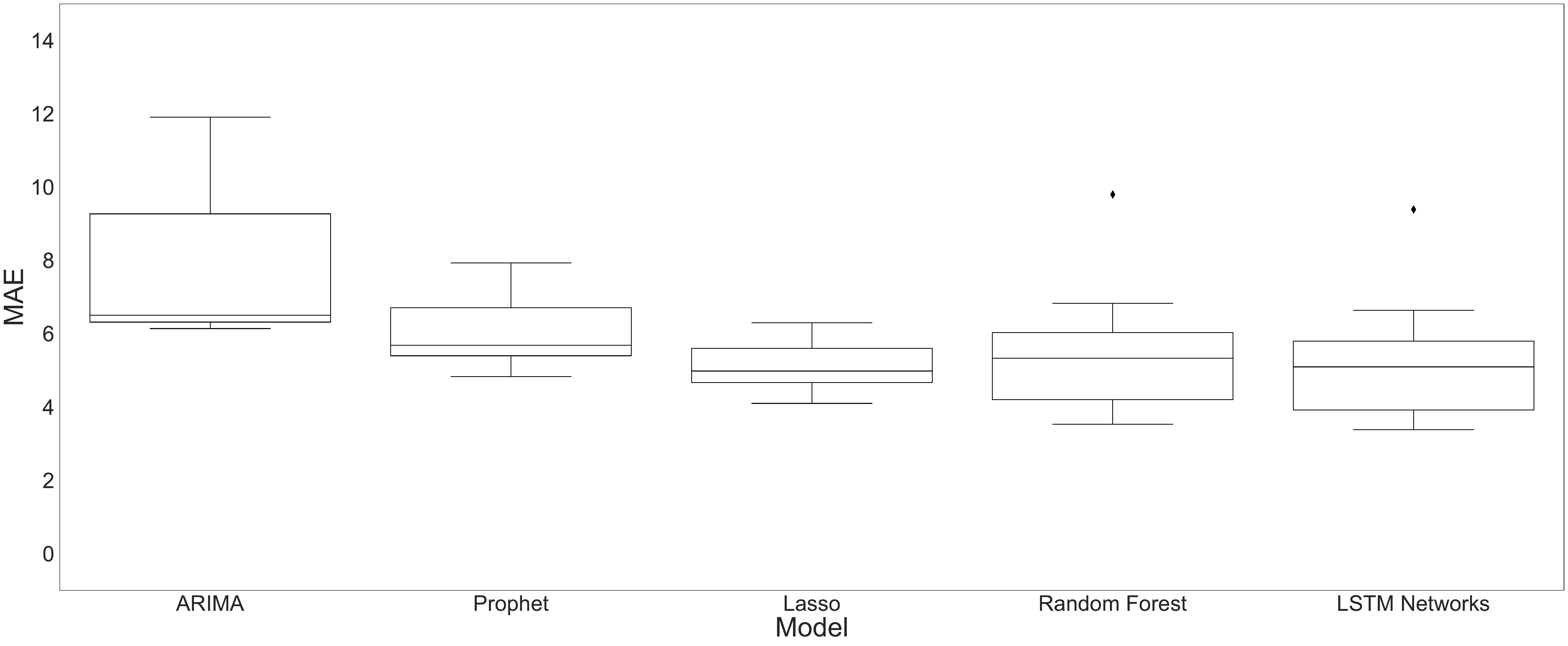}}
\captionsetup{justification=centering}
\caption{2 years rolling window, retraining every 30 days}
\end{subfigure}
\begin{subfigure}{0.5\linewidth}
\centerline{\includegraphics[width=\textwidth, height=0.55\hsize]{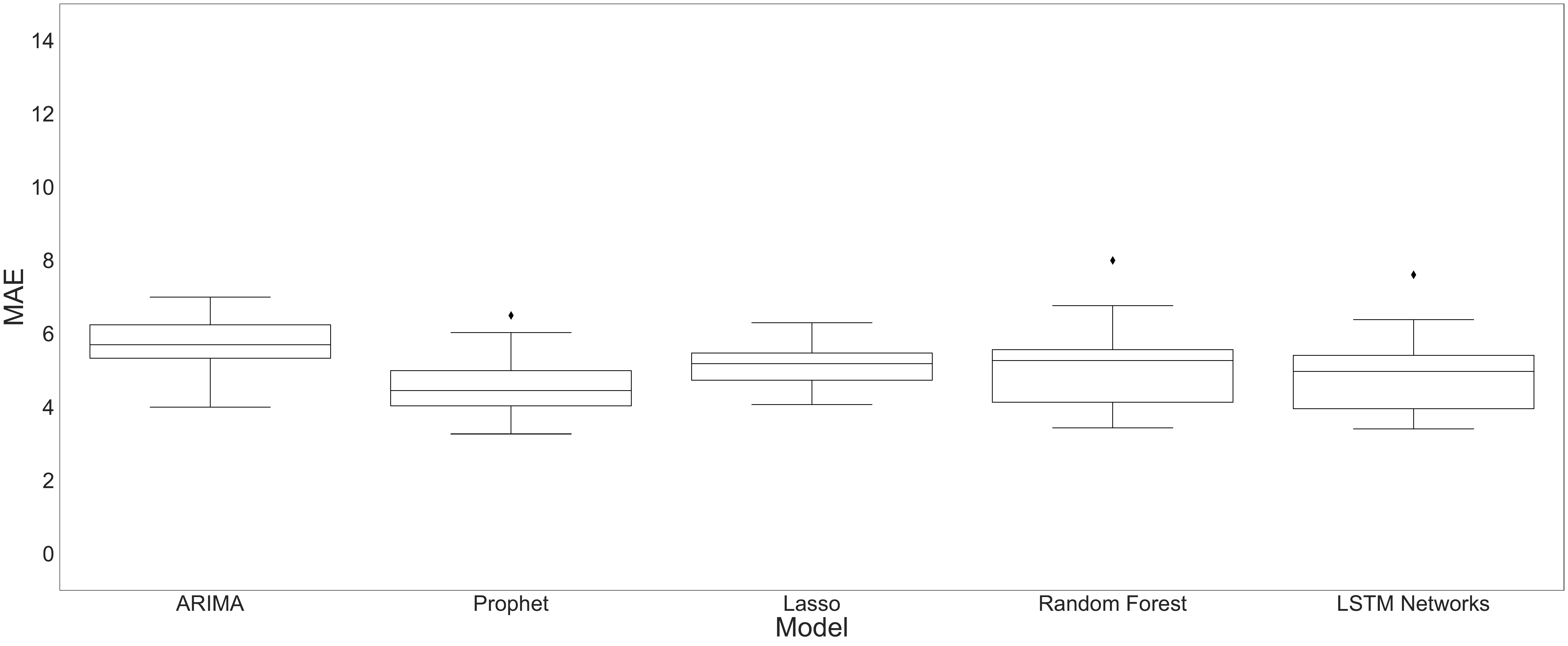}}
\captionsetup{justification=centering}
\caption{8 years rolling window, retraining every 30 days}
\end{subfigure}
\begin{subfigure}{0.5\linewidth}
\centerline{\includegraphics[width=\textwidth, height=0.55\hsize]{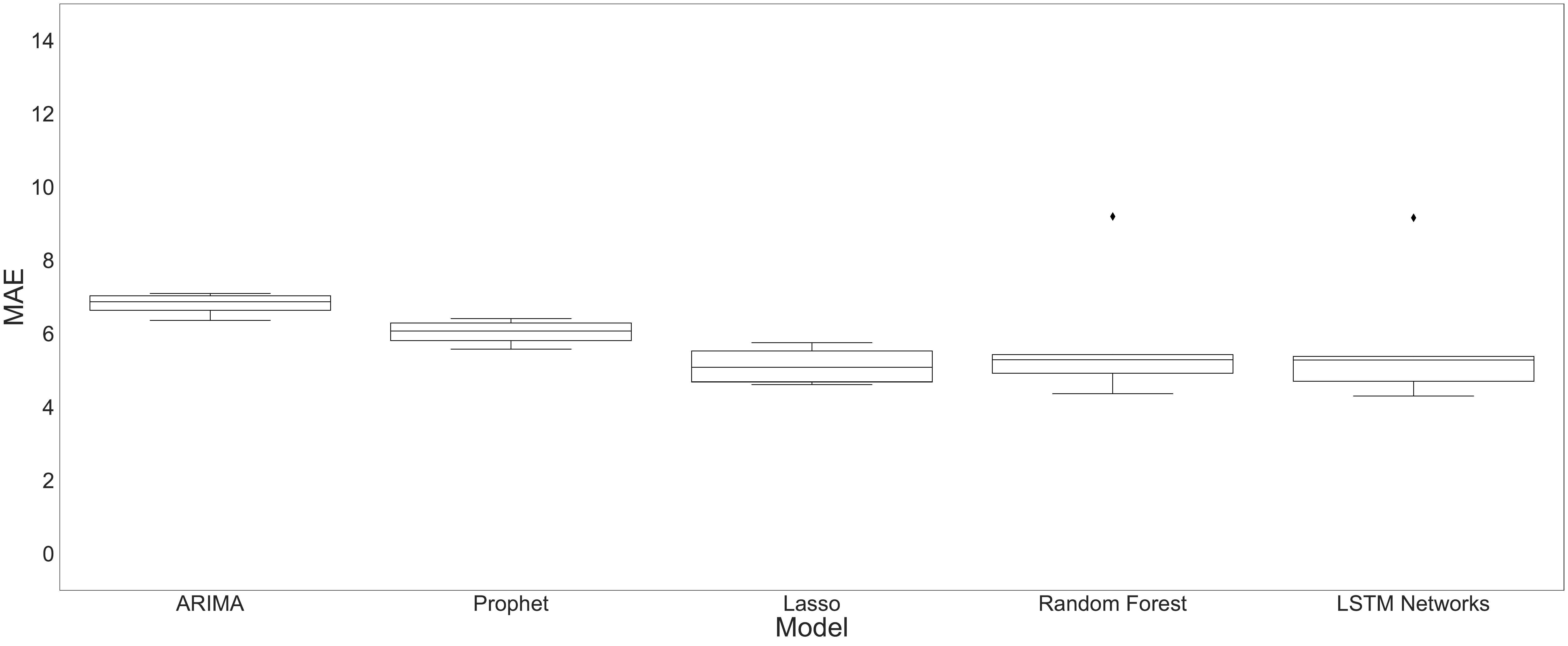}}
\captionsetup{justification=centering}
\caption{2 years rolling window, retraining every 90 days}
\label{fig:MAE(g)}
\end{subfigure}
\begin{subfigure}{0.5\linewidth}
\centerline{\includegraphics[width=\textwidth, height=0.55\hsize]{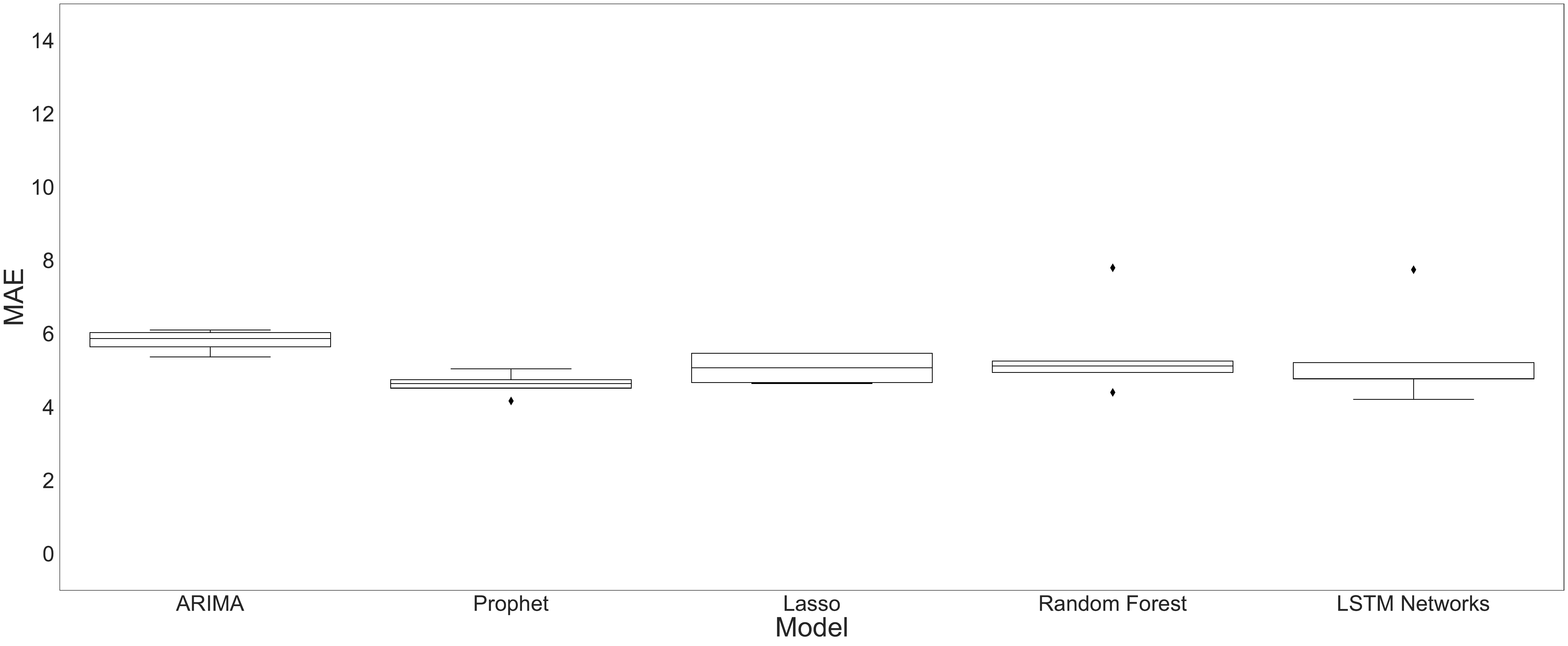}}
\captionsetup{justification=centering}
\caption{8 years rolling window, retraining every 90 days}
\end{subfigure}
\captionsetup{justification=centering}
\caption{MAE with different training window sizes and retraining periods}
\label{fig:MAE}
\end{figure}

\begin{figure}[H]
\begin{subfigure}{0.5\linewidth}
\centerline{\includegraphics[width=\textwidth, height=0.55\hsize]{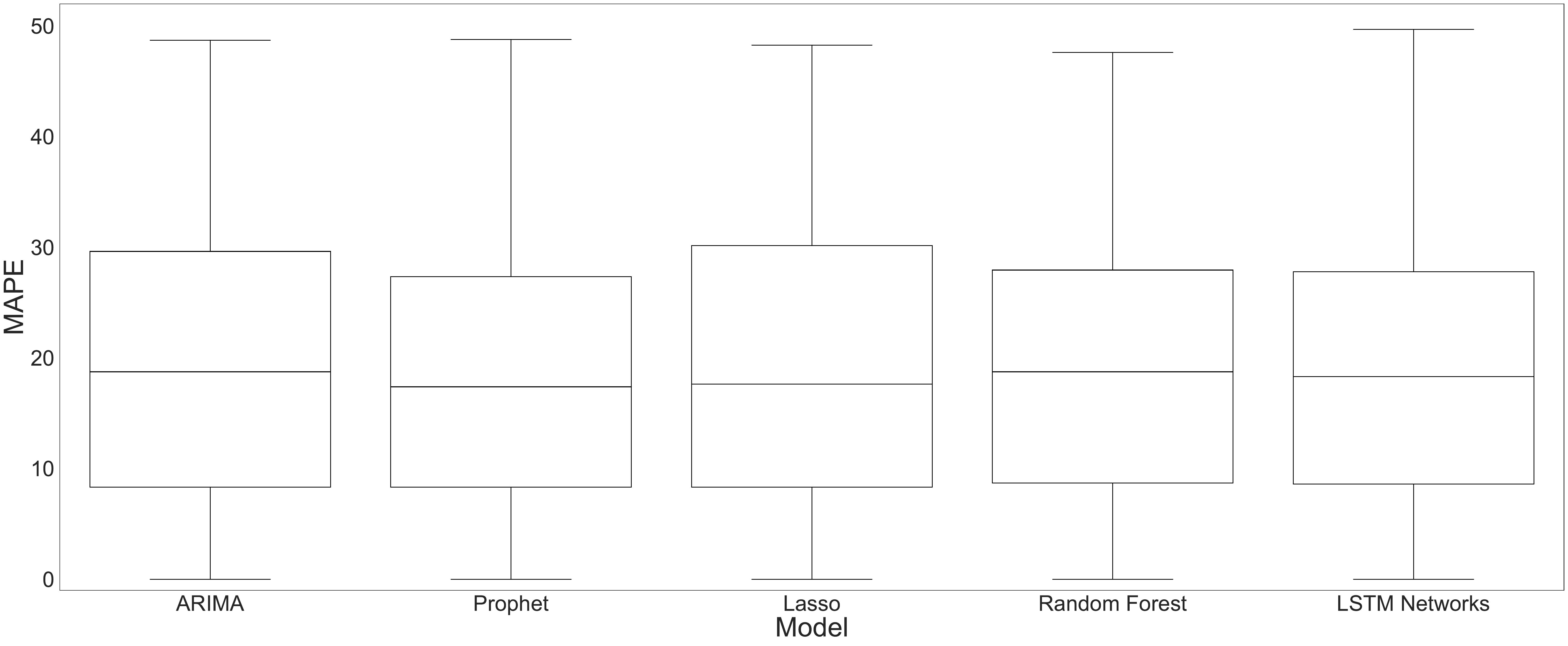}}
\captionsetup{justification=centering}
\caption{2 years rolling window, retraining every day}
\end{subfigure}%
\begin{subfigure}{0.5\linewidth}
\centerline{\includegraphics[width=\textwidth, height=0.55\hsize]{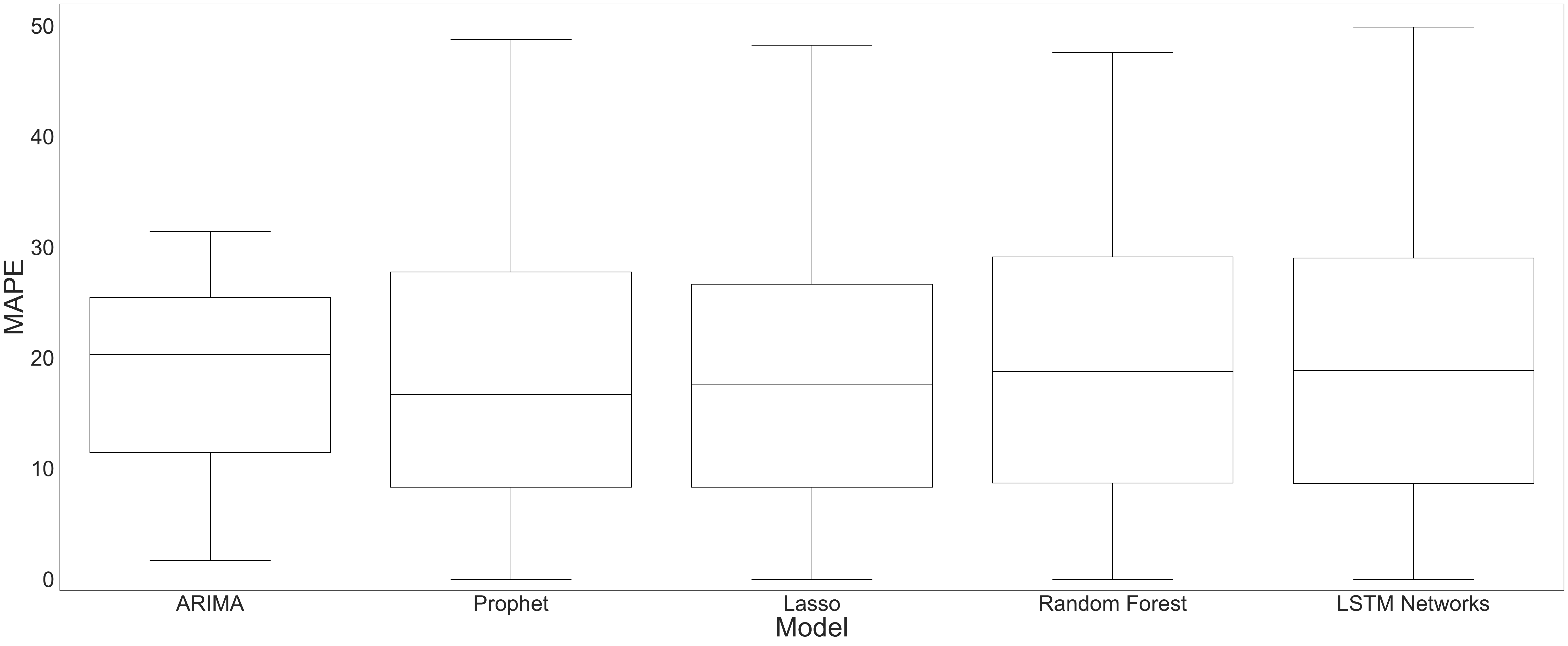}}
\captionsetup{justification=centering}
\caption{8 years rolling window, retraining every day}
\end{subfigure}
\begin{subfigure}{0.5\linewidth}
\centerline{\includegraphics[width=\textwidth, height=0.55\hsize]{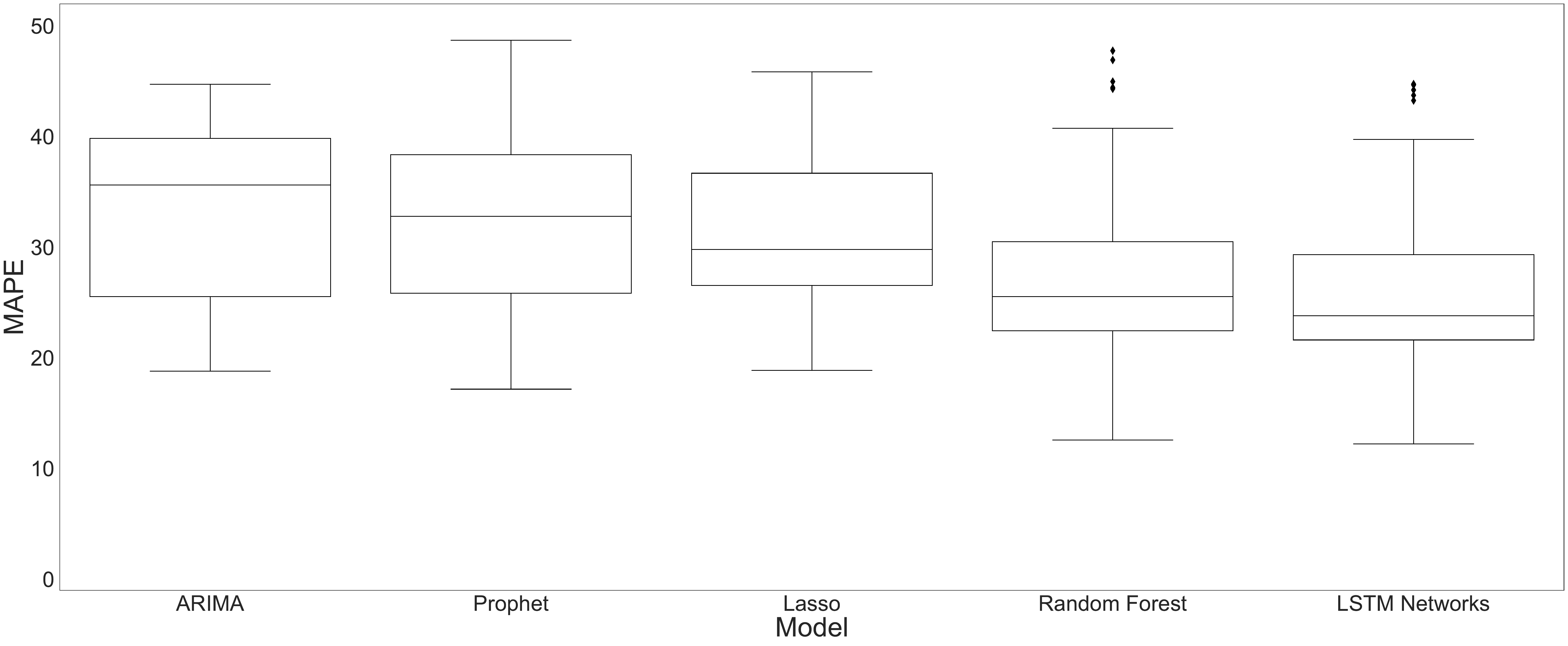}}
\caption{2 years rolling window, retraining every 7 days}
\end{subfigure}
\begin{subfigure}{0.5\linewidth}
\centerline{\includegraphics[width=\textwidth, height=0.55\hsize]{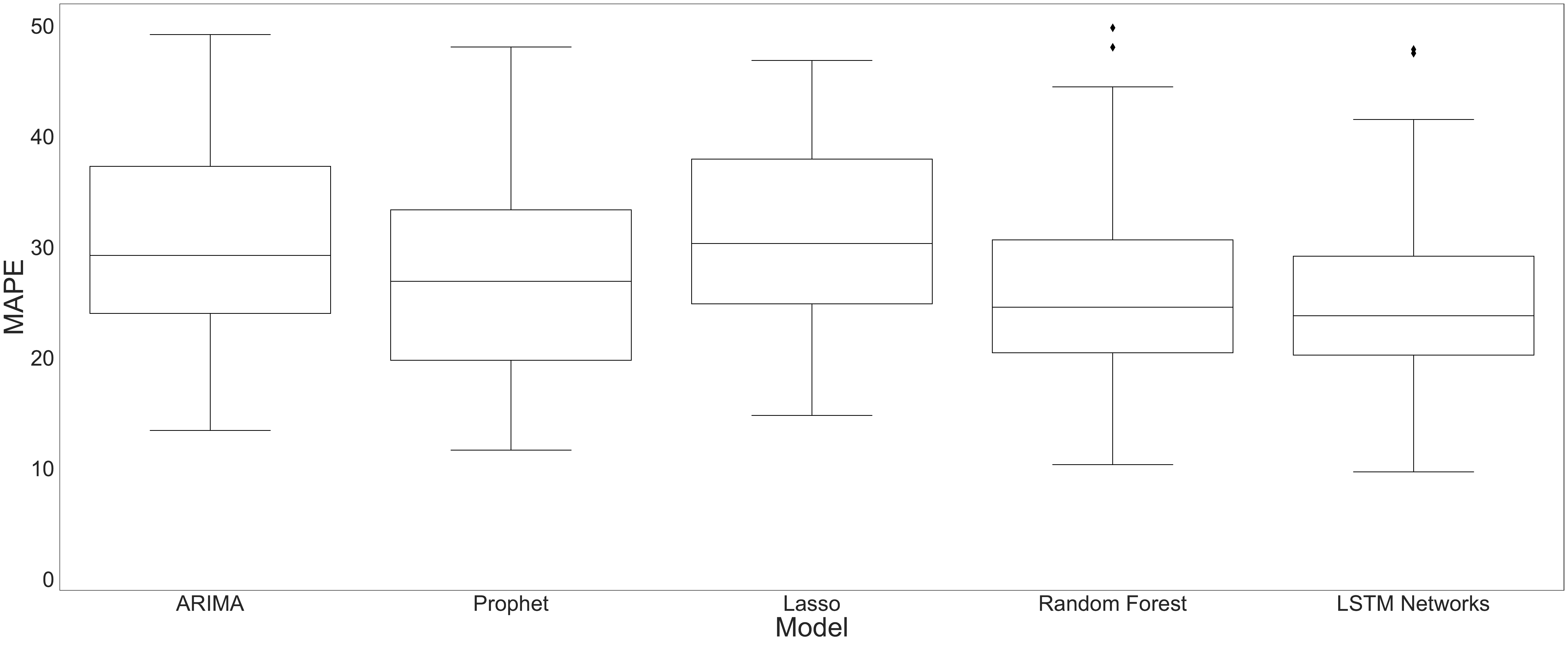}}
\captionsetup{justification=centering}
\caption{8 years rolling window, retraining every 7 days}
\label{f124}
\end{subfigure}
\begin{subfigure}{0.5\linewidth}
\centerline{\includegraphics[width=\textwidth, height=0.55\hsize]{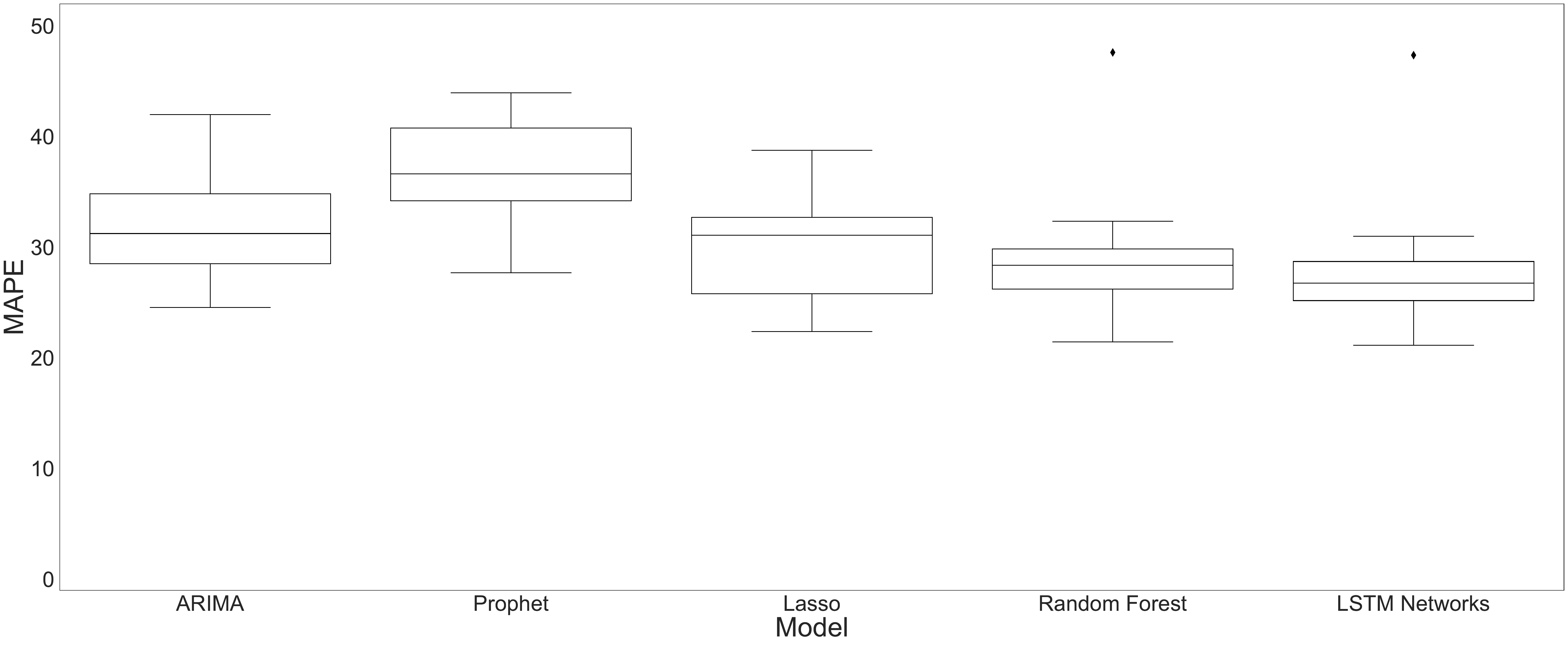}}
\captionsetup{justification=centering}
\caption{2 years rolling window, retraining every 30 days}
\end{subfigure}
\begin{subfigure}{0.5\linewidth}
\centerline{\includegraphics[width=\textwidth, height=0.55\hsize]{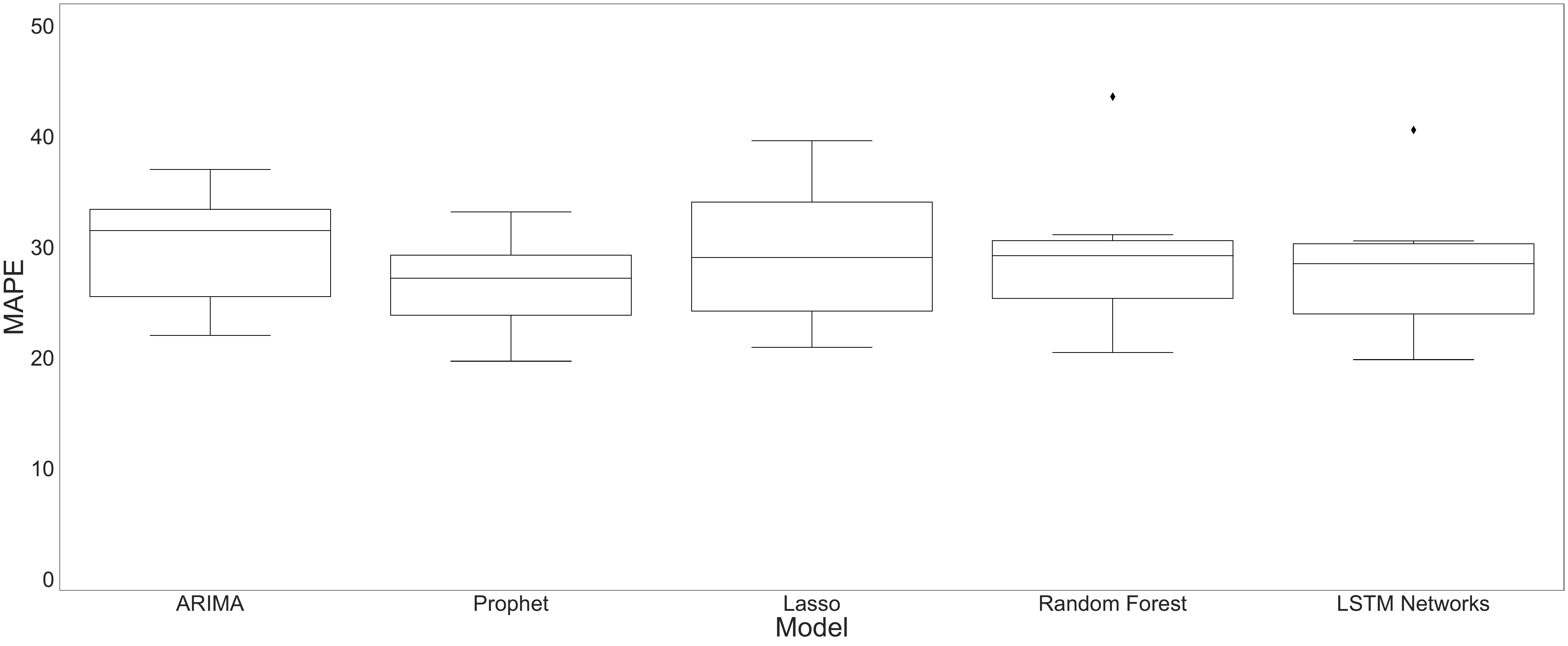}}
\captionsetup{justification=centering}
\caption{8 years rolling window, retraining every 30 days}
\end{subfigure}
\begin{subfigure}{0.5\linewidth}
\centerline{\includegraphics[width=\textwidth, height=0.55\hsize]{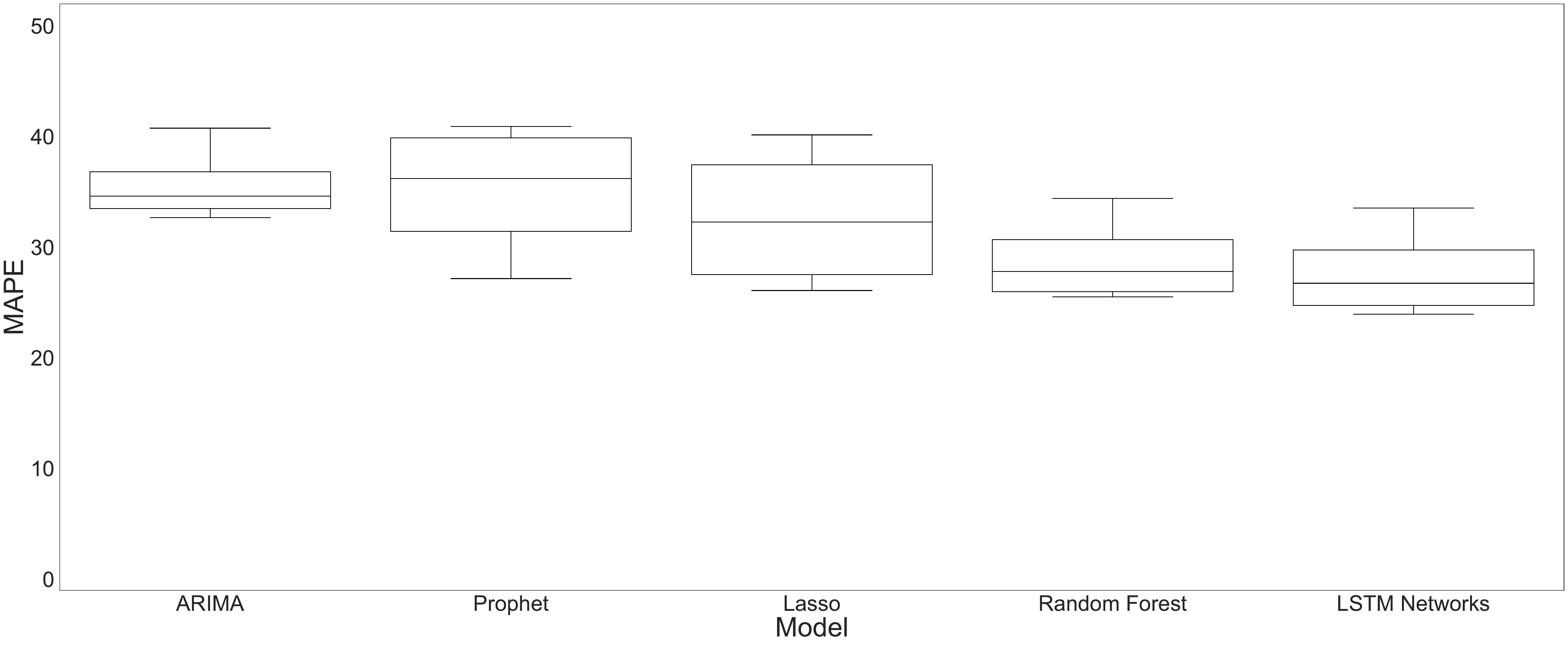}}
\captionsetup{justification=centering}
\caption{2 years rolling window, retraining every 90 days}
\end{subfigure}
\begin{subfigure}{0.5\linewidth}
\centerline{\includegraphics[width=\textwidth, height=0.55\hsize]{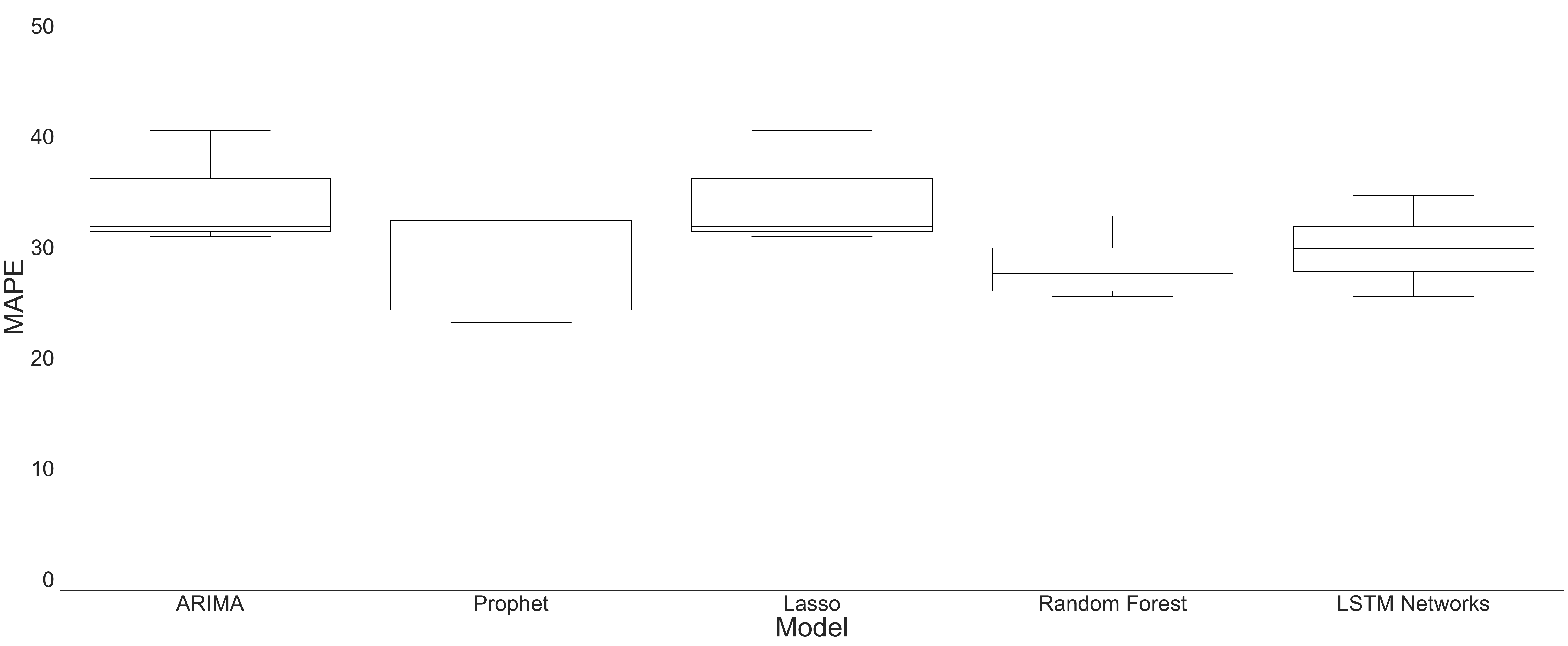}}
\captionsetup{justification=centering}
\caption{8 years rolling window, retraining every 90 days}
\end{subfigure}
\captionsetup{justification=centering}
\caption{MAPE with different training window sizes and retraining periods}
\label{fig:MAPE}
\end{figure}

\begin{figure}[H]
\begin{subfigure}{0.5\linewidth}
\centerline{\includegraphics[width=\textwidth, height=0.55\hsize]{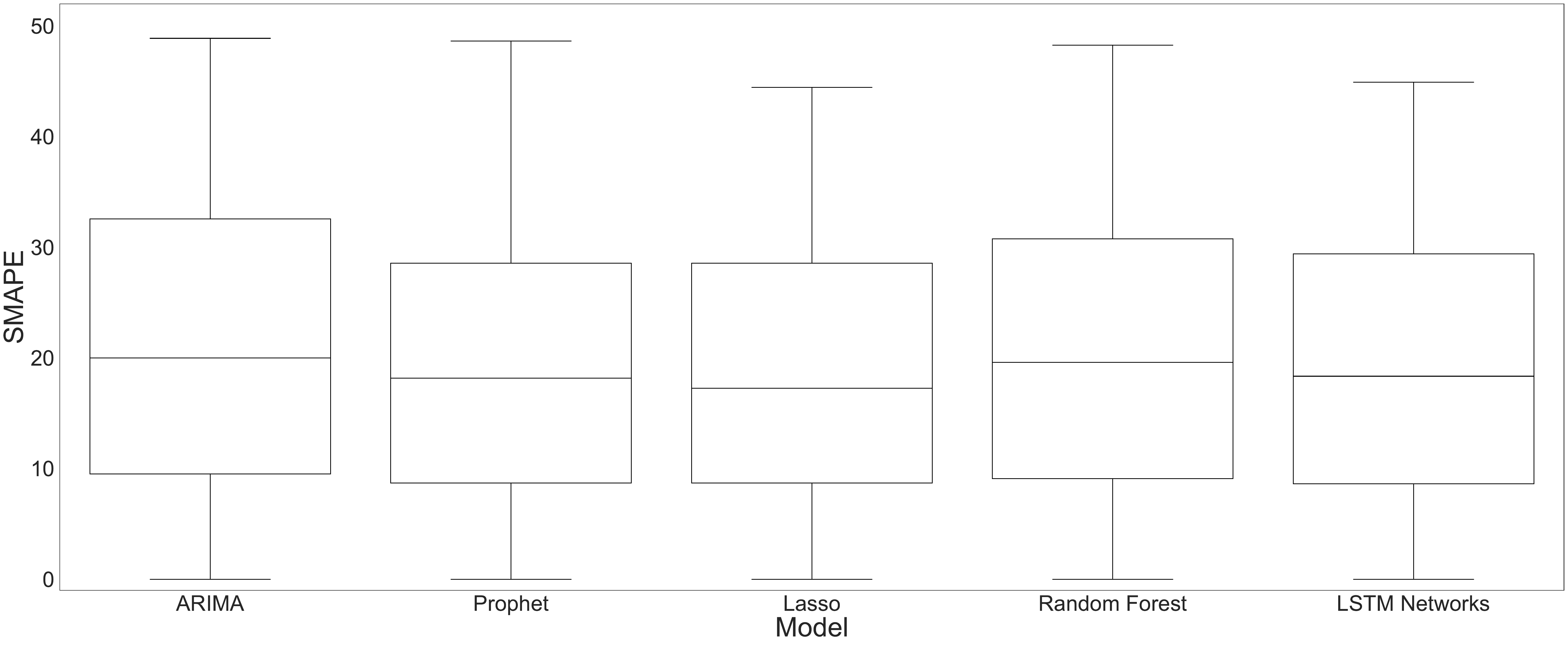}}
\captionsetup{justification=centering}
\caption{2 years rolling window, retraining every day}
\end{subfigure}%
\begin{subfigure}{0.5\linewidth}
\centerline{\includegraphics[width=\textwidth, height=0.55\hsize]{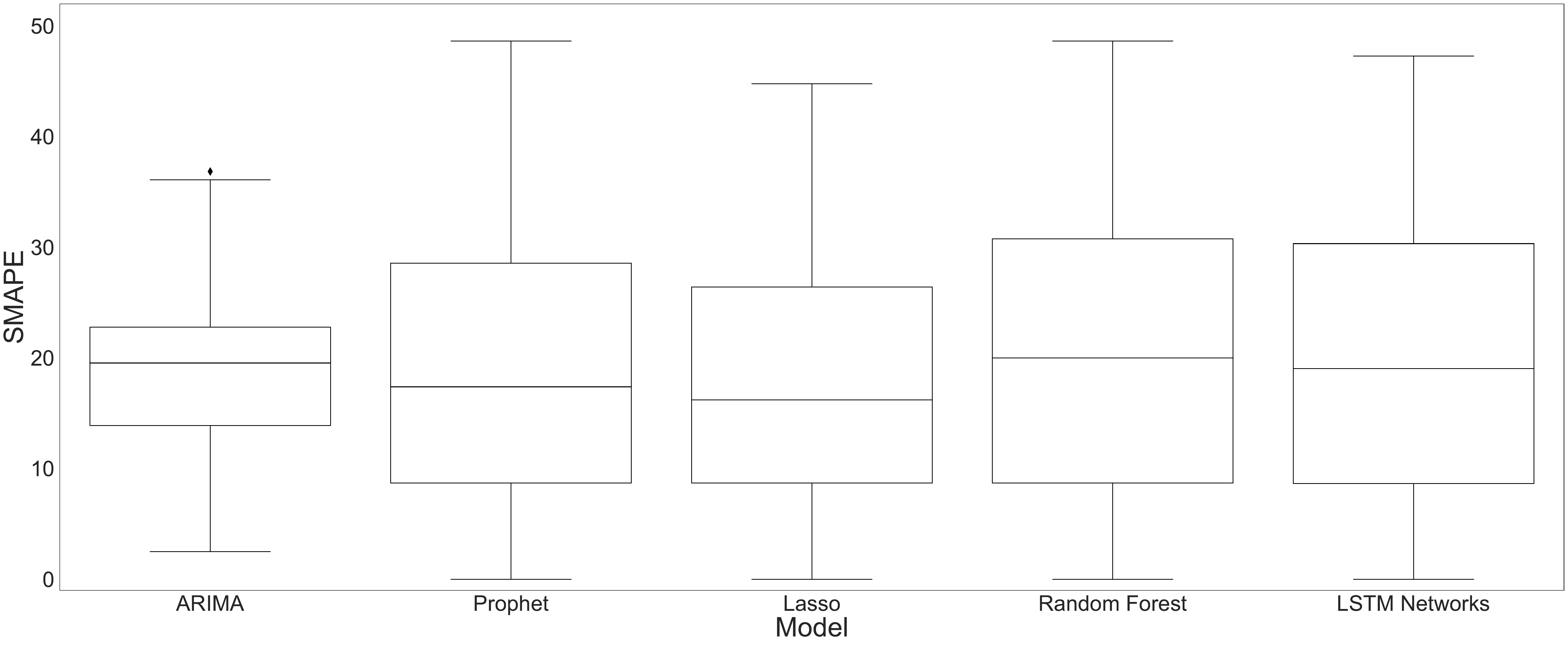}}
\captionsetup{justification=centering}
\caption{8 years rolling window, retraining every day}
\end{subfigure}
\begin{subfigure}{0.5\linewidth}
\centerline{\includegraphics[width=\textwidth, height=0.55\hsize]{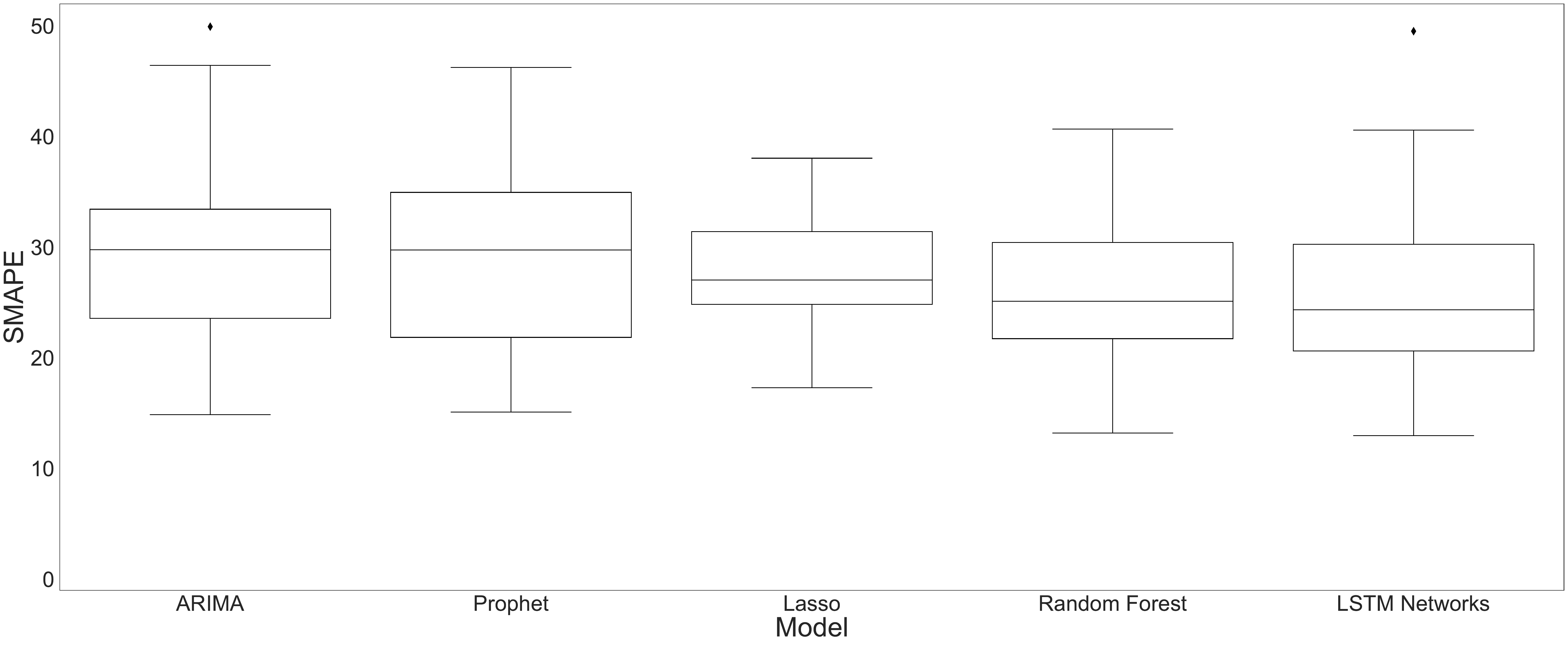}}
\caption{2 years rolling window, retraining every 7 days}
\end{subfigure}
\begin{subfigure}{0.5\linewidth}
\centerline{\includegraphics[width=\textwidth, height=0.55\hsize]{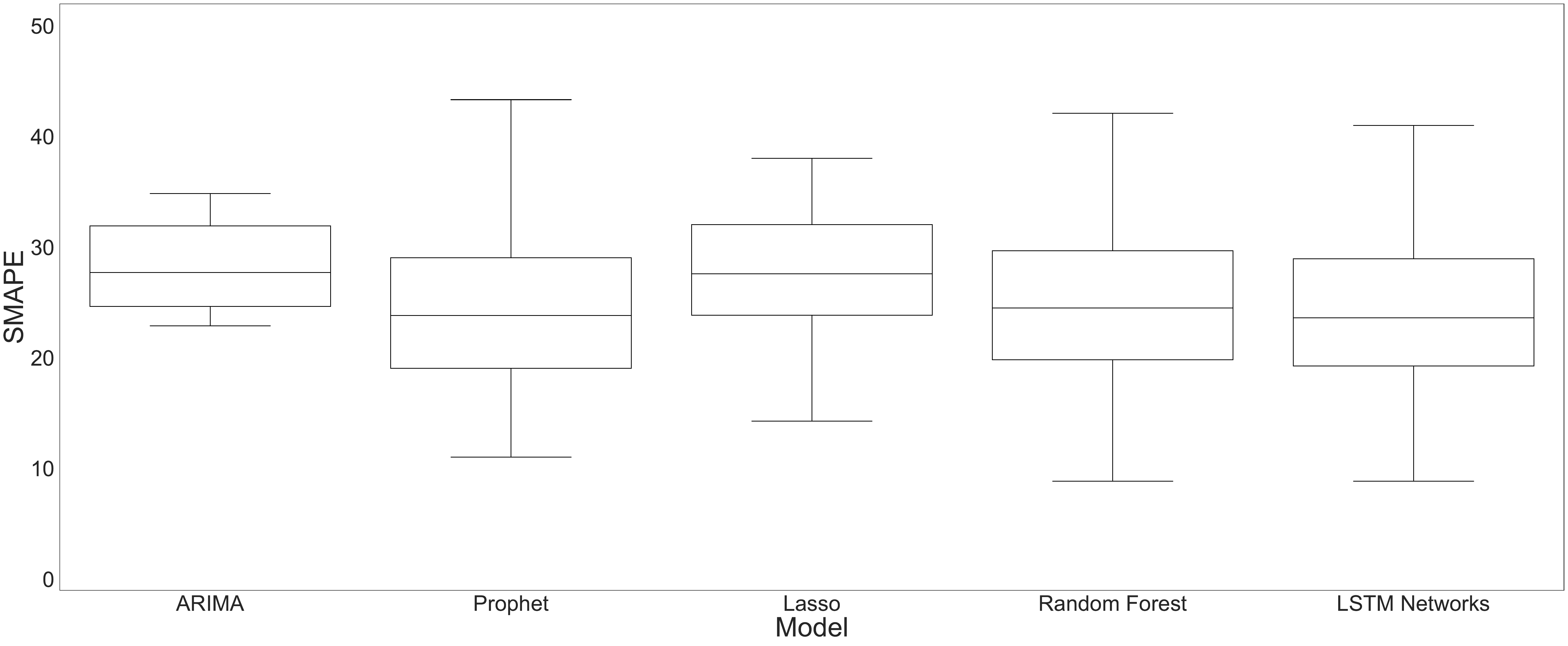}}
\captionsetup{justification=centering}
\caption{8 years rolling window, retraining every 7 days}
\label{f124}
\end{subfigure}
\begin{subfigure}{0.5\linewidth}
\centerline{\includegraphics[width=\textwidth, height=0.55\hsize]{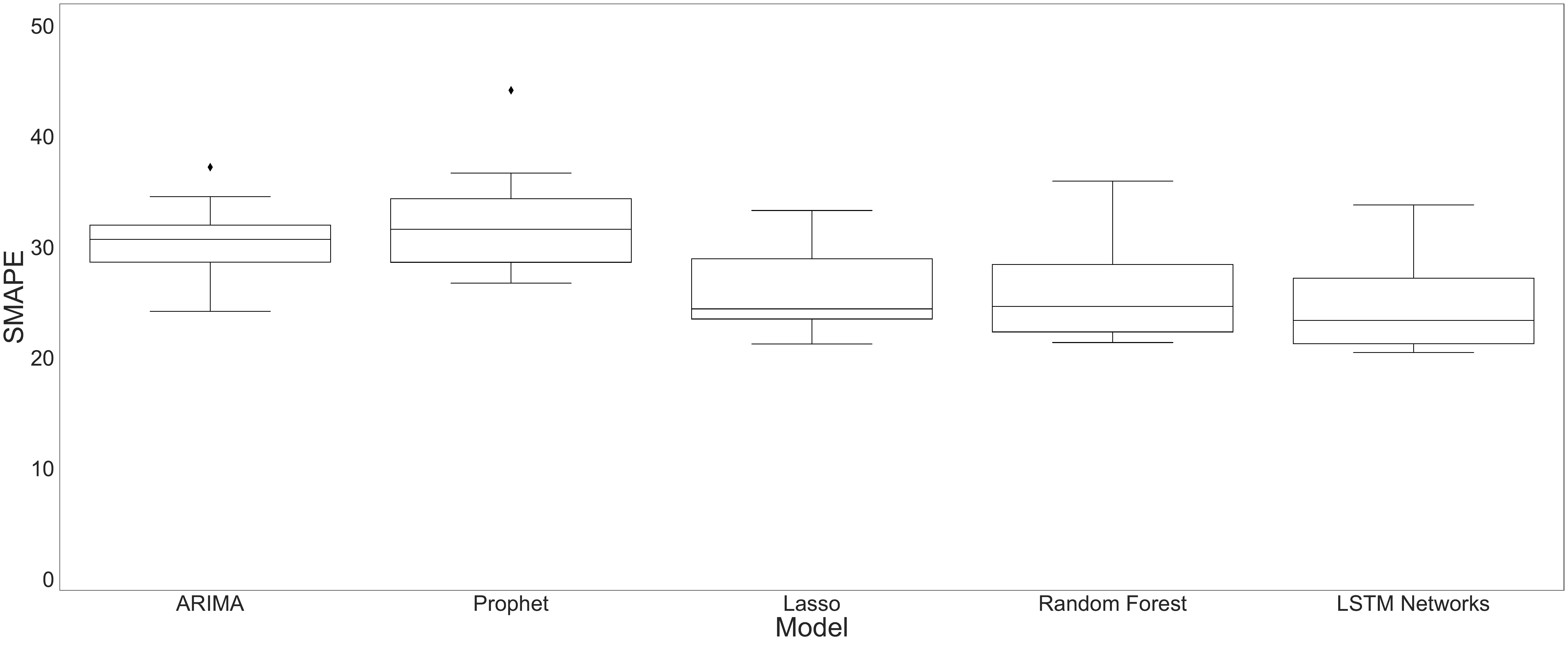}}
\captionsetup{justification=centering}
\caption{2 years rolling window, retraining every 30 days}
\end{subfigure}
\begin{subfigure}{0.5\linewidth}
\centerline{\includegraphics[width=\textwidth, height=0.55\hsize]{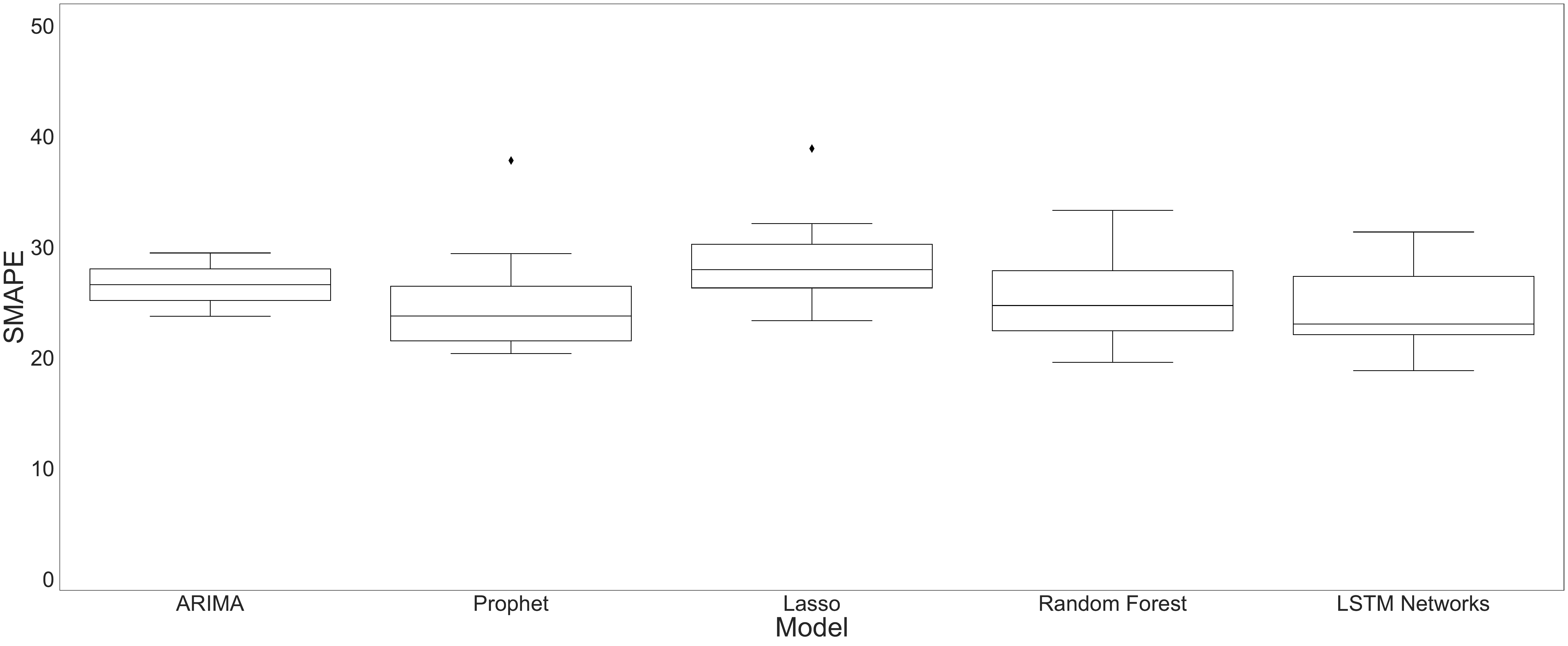}}
\captionsetup{justification=centering}
\caption{8 years rolling window, retraining every 30 days}
\end{subfigure}
\begin{subfigure}{0.5\linewidth}
\centerline{\includegraphics[width=\textwidth, height=0.55\hsize]{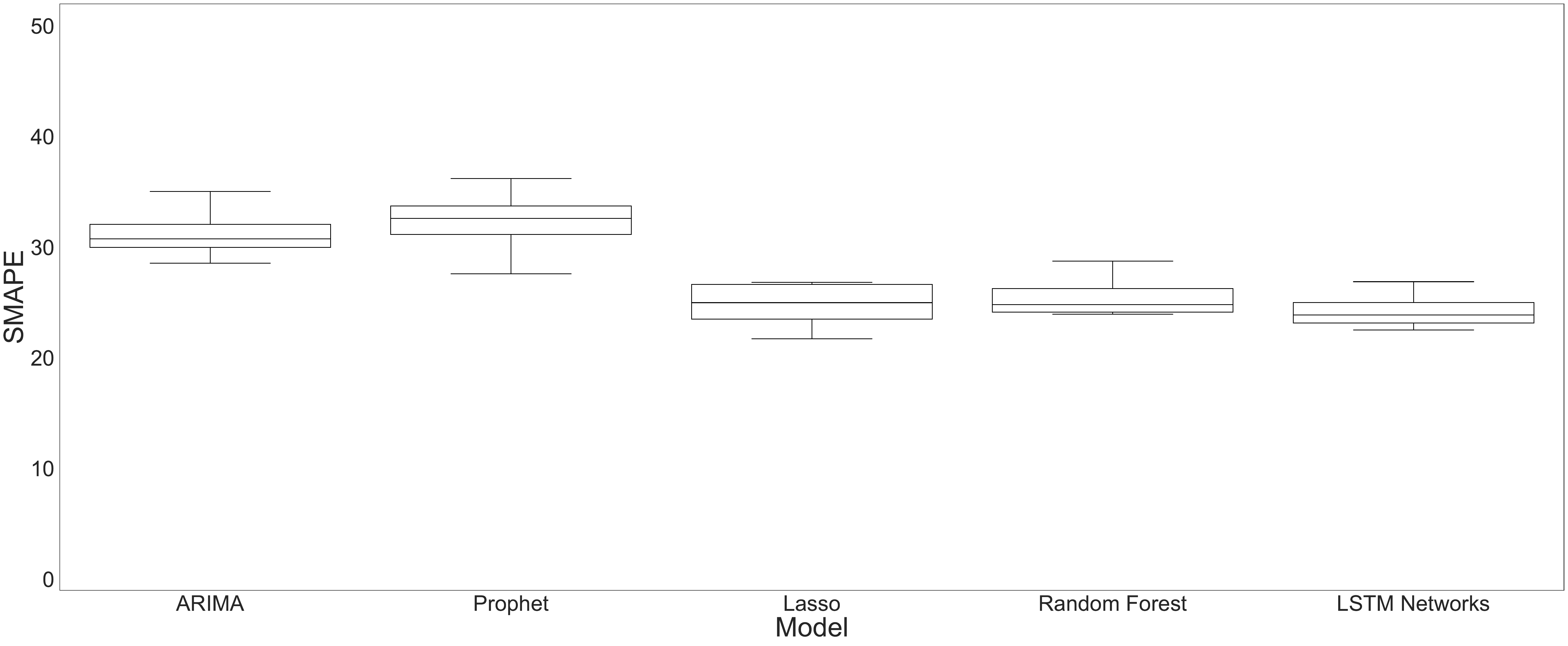}}
\captionsetup{justification=centering}
\caption{2 years rolling window, retraining every 90 days}
\end{subfigure}
\begin{subfigure}{0.5\linewidth}
\centerline{\includegraphics[width=\textwidth, height=0.55\hsize]{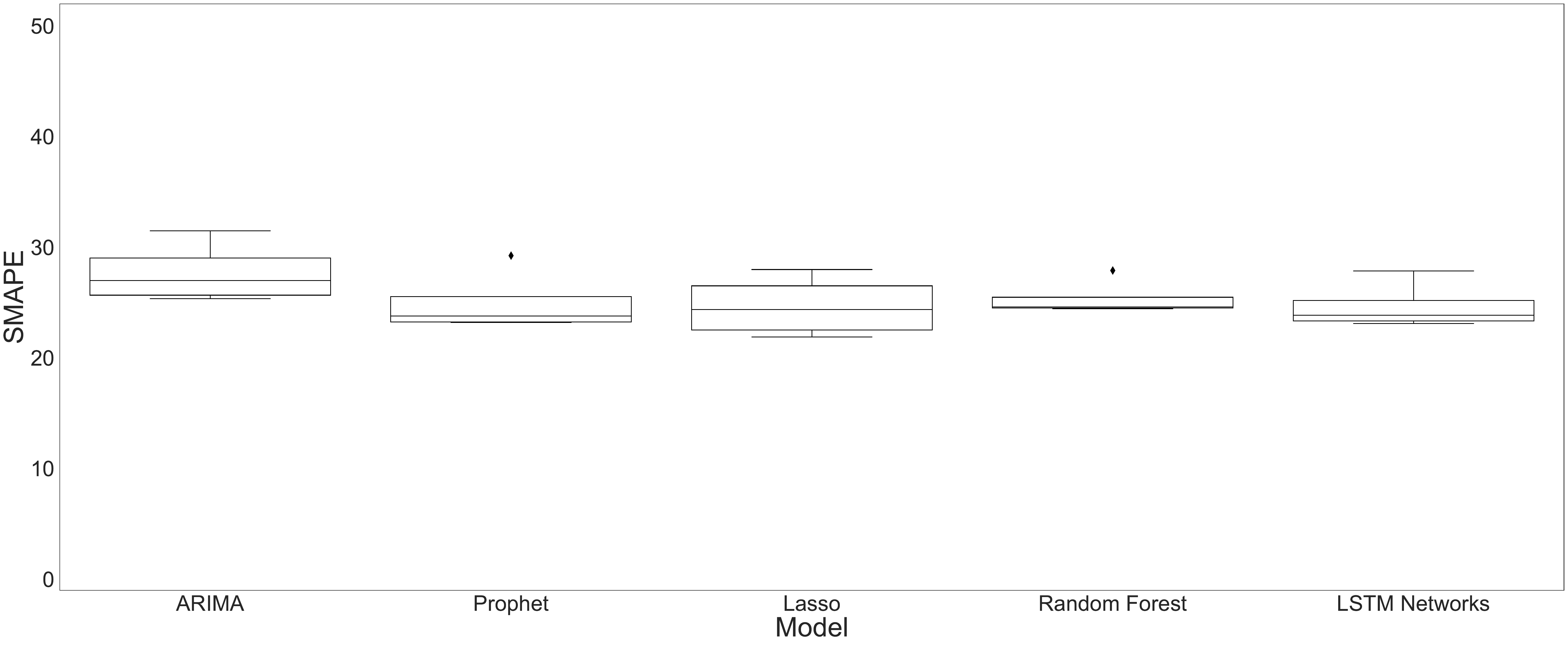}}
\captionsetup{justification=centering}
\caption{8 years rolling window, retraining every 90 days}
\end{subfigure}
\captionsetup{justification=centering}
\caption{SMAPE with different training window sizes and retraining periods}
\label{fig:SMAPE}
\end{figure}

The performance of the forecasting models is computed based on the rolling-origin evaluation and by four error measures, RMSE, MAE, MAPE, SMAPE. The first two error measures, RMSE and MAE, are absolute measures while the remaining ones, MAPE and SMAPE, are relative measures. The errors are measured for each rolling origin for the test data and reported in Figures \ref{fig:RMSE}-\ref{fig:SMAPE} and Table \ref{tab:errors}. Table \ref{tab:errors} gives the mean and standard deviation of the errors for different training window sizes and retraining periods.

Figures \ref{fig:RMSE} and \ref{fig:MAE} compare the RMSE and MAE of the models trained with different training window sizes and retraining periods. As we can see in these figures and in Table \ref{tab:errors}, increasing the size of the training window mostly affects the univariate models, ARIMA and Prophet. ARIMA's performance improves when moving from two years to eight years of data. Since ARIMA's forecasts are only based on the previous demands, and the seasonality in data has not changed significantly during the eight years, the model parameters, $p$ and $q$, are more robust for longer time series data (including 5 lagged values and a moving average of 2), resulting in more accurate forecasts. In general, when a limited amount of data are available, the ARIMA model has a high forecast error not only because its forecasts are solely based on the previous demands, but also due to the fact that it cannot capture the seasonality in data. Prophet's accuracy is also improved as the amount of data increases. However, unlike ARIMA, forecast errors are similar for different retraining periods. The results for the lasso regression and LSTM network indicate that there is not much difference for these methods when there is a large amount of data for training, or when different retraining periods are considered. Random forest does see a slight improvement with eight years of data, and it is the only multivariate model to see this improvement. Its forecast errors are very close for different retraining periods.

In terms of the retraining periods, retraining the models less frequently reduces the variability of the error. If we compare Figure \ref{fig:RMSE}(\subref{fig:RMSE(a)}) with Figure \ref{fig:RMSE}(\subref{fig:RMSE(g)}), we see that the RMSE error is less variable in Figure \ref{fig:RMSE}(\subref{fig:RMSE(g)}) for all the models, similarly for MAE in Figure \ref{fig:MAE}. This can also be verified from the results in Table \ref{tab:errors}, where we see lower standard deviations as we move down to retraining every 90 days.

Figures \ref{fig:MAPE} and \ref{fig:SMAPE} compare the MAPE and SMAPE of the models trained with different training window sizes and retraining periods. As we can see from these figures and from Table \ref{tab:errors}, increasing the training window size does not necessarily decrease the errors. There is similar behavior for the retraining periods, especially for the multivariate models, but we see that retraining less frequently results in less variable errors.

Overall, the results indicate that while univariate models can benefit from a larger training window size and frequent retraining, the performance of the multivariate models is not affected by a larger training window, meaning that these models have robust performance with different data volumes.

\begin{landscape}
\begin{table}[H]
\captionsetup{justification=centering}
\caption{Model performance with different training window sizes and retraining periods}
\begin{tabular}{|c|c|cccc|cccc|}
\hline
                           & \textbf{}      & \multicolumn{4}{c|}{\textbf{Two years}}                                                                                      & \multicolumn{4}{c|}{\textbf{Eight years}}                                                                                    \\ \hline
\begin{tabular}[x]{@{}c@{}}\textbf{Retraining}\\\textbf{Period}\end{tabular}  & \textbf{Model} & \multicolumn{1}{c|}{\textbf{RMSE}} & \multicolumn{1}{c|}{\textbf{MAE}} & \multicolumn{1}{c|}{\textbf{MAPE}} & \textbf{SMAPE} & \multicolumn{1}{c|}{\textbf{RMSE}} & \multicolumn{1}{c|}{\textbf{MAE}} & \multicolumn{1}{c|}{\textbf{MAPE}} & \textbf{SMAPE} \\ \hline
\multirow{5}{*}{1 day}     & ARIMA          & \multicolumn{1}{c|}{6.7$\pm$2.76}     & \multicolumn{1}{c|}{7.37$\pm$1.75}    & \multicolumn{1}{c|}{19.46$\pm$12.55}   & 21.37$\pm$14.01    & \multicolumn{1}{c|}{5.72$\pm$2.76}     & \multicolumn{1}{c|}{4.84$\pm$3.43}    & \multicolumn{1}{c|}{18.77$\pm$8.09}    & 18.79$\pm$6.68     \\ \cline{2-10} 
                           & Prophet        & \multicolumn{1}{c|}{4.20$\pm$3.19}     & \multicolumn{1}{c|}{4.20$\pm$3.19}    & \multicolumn{1}{c|}{18.52$\pm$12.85}   & 19.05$\pm$13.12    & \multicolumn{1}{c|}{4.28$\pm$3.12}     & \multicolumn{1}{c|}{4.28$\pm$3.12}    & \multicolumn{1}{c|}{18.99$\pm$12.43}   & 19.72$\pm$13.04    \\ \cline{2-10} 
                           & Lasso Regression         & \multicolumn{1}{c|}{4.88$\pm$3.40}     & \multicolumn{1}{c|}{4.88$\pm$3.40}    & \multicolumn{1}{c|}{19.66$\pm$12.86}   & 18.92$\pm$12.25    & \multicolumn{1}{c|}{4.80$\pm$3.39}     & \multicolumn{1}{c|}{4.80$\pm$3.39}    & \multicolumn{1}{c|}{18.96$\pm$12.43}   & 18.08$\pm$11.63    \\ \cline{2-10} 
                           & Random Forest  & \multicolumn{1}{c|}{4.59$\pm$3.24}     & \multicolumn{1}{c|}{4.59$\pm$3.24}    & \multicolumn{1}{c|}{19.49$\pm$12.58}   & 20.23$\pm$13.05    & \multicolumn{1}{c|}{4.59$\pm$3.27}     & \multicolumn{1}{c|}{4.59$\pm$3.27}    & \multicolumn{1}{c|}{19.57$\pm$12.95}   & 20.38$\pm$13.32    \\ \cline{2-10} 
                           & LSTM Networks  & \multicolumn{1}{c|}{4.31$\pm$2.99}     & \multicolumn{1}{c|}{4.54$\pm$3.28}    & \multicolumn{1}{c|}{19.4$\pm$12.69}    & 19.23$\pm$12.36    & \multicolumn{1}{c|}{3.09$\pm$2.49}     & \multicolumn{1}{c|}{4.49$\pm$3.24}    & \multicolumn{1}{c|}{19.84$\pm$13.37}   & 19.77$\pm$12.94    \\ \hline
\multirow{5}{*}{7 days}    & ARIMA          & \multicolumn{1}{c|}{6.81$\pm$2.09}     & \multicolumn{1}{c|}{6.43$\pm$0.79}    & \multicolumn{1}{c|}{33.19$\pm$8.33}    & 30.0$\pm$7.74      & \multicolumn{1}{c|}{5.85$\pm$0.84}     & \multicolumn{1}{c|}{5.56$\pm$1.75}    & \multicolumn{1}{c|}{31.14$\pm$8.75}    & 28.26$\pm$3.86     \\ \cline{2-10} 
                           & Prophet        & \multicolumn{1}{c|}{7.34$\pm$2.01}     & \multicolumn{1}{c|}{5.95$\pm$1.74}    & \multicolumn{1}{c|}{32.32$\pm$8.36}    & 28.65$\pm$8.25     & \multicolumn{1}{c|}{5.6$\pm$1.94}      & \multicolumn{1}{c|}{4.61$\pm$1.69}    & \multicolumn{1}{c|}{27.22$\pm$9.09}    & 24.32$\pm$6.76     \\ \cline{2-10} 
                           & Lasso  Regression        & \multicolumn{1}{c|}{6.22$\pm$1.59}     & \multicolumn{1}{c|}{5.15$\pm$1.42}    & \multicolumn{1}{c|}{31.7$\pm$7.50}     & 27.7$\pm$4.94      & \multicolumn{1}{c|}{6.31$\pm$1.80}     & \multicolumn{1}{c|}{5.16$\pm$1.49}    & \multicolumn{1}{c|}{31.64$\pm$8.37}    & 27.28$\pm$5.55     \\ \cline{2-10} 
                           & Random Forest  & \multicolumn{1}{c|}{6.3$\pm$2.00}      & \multicolumn{1}{c|}{5.17$\pm$1.68}    & \multicolumn{1}{c|}{27.48$\pm$8.70}    & 25.64$\pm$6.62     & \multicolumn{1}{c|}{6.12$\pm$1.94}     & \multicolumn{1}{c|}{5.01$\pm$1.63}    & \multicolumn{1}{c|}{26.8$\pm$8.78}     & 25.43$\pm$7.33     \\ \cline{2-10} 
                           & LSTM Networks  & \multicolumn{1}{c|}{4.32$\pm$2.01}     & \multicolumn{1}{c|}{4.97$\pm$1.60}    & \multicolumn{1}{c|}{26.45$\pm$8.39}    & 25.83$\pm$7.25      & \multicolumn{1}{c|}{4.07$\pm$1.82}     & \multicolumn{1}{c|}{4.87$\pm$1.61}    & \multicolumn{1}{c|}{25.83$\pm$8.43}    & 24.53$\pm$7.08     \\ \hline
\multirow{5}{*}{30 days}   & ARIMA          & \multicolumn{1}{c|}{7.18$\pm$1.22}     & \multicolumn{1}{c|}{7.96$\pm$2.34}    & \multicolumn{1}{c|}{32.49$\pm$5.34}    & 30.46$\pm$3.43     & \multicolumn{1}{c|}{4.86$\pm$0.94}     & \multicolumn{1}{c|}{5.68 $\pm$ 0.88}  & \multicolumn{1}{c|}{30.06$\pm$4.90}    & 26.63$\pm$2.86     \\ \cline{2-10} 
                           & Prophet        & \multicolumn{1}{c|}{6.62$\pm$1.23}     & \multicolumn{1}{c|}{6.01$\pm$0.87}    & \multicolumn{1}{c|}{36.92$\pm$5.06}    & 32.38$\pm$4.80     & \multicolumn{1}{c|}{4.85$\pm$1.20}     & \multicolumn{1}{c|}{4.62$\pm$0.95}    & \multicolumn{1}{c|}{26.42$\pm$4.05}    & 25.01$\pm$4.78     \\ \cline{2-10} 
                           & Lasso Regression        & \multicolumn{1}{c|}{5.86$\pm$0.74}     & \multicolumn{1}{c|}{5.12$\pm$0.62}    & \multicolumn{1}{c|}{30.2$\pm$5.14}     & 25.79$\pm$3.48     & \multicolumn{1}{c|}{5.44$\pm$0.89}     & \multicolumn{1}{c|}{5.11$\pm$0.62}    & \multicolumn{1}{c|}{29.4$\pm$6.28}     & 28.73$\pm$3.66     \\ \cline{2-10} 
                           & Random Forest  & \multicolumn{1}{c|}{6.13$\pm$1.61}     & \multicolumn{1}{c|}{5.42$\pm$1.60}    & \multicolumn{1}{c|}{29.09$\pm$6.41}    & 25.93$\pm$4.33     & \multicolumn{1}{c|}{5.41$\pm$1.36}     & \multicolumn{1}{c|}{5.19$\pm$1.22}    & \multicolumn{1}{c|}{28.66$\pm$5.70}    & 25.52$\pm$3.87     \\ \cline{2-10} 
                           & LSTM Networks  & \multicolumn{1}{c|}{5.41$\pm$0.81}     & \multicolumn{1}{c|}{5.2$\pm$1.54}     & \multicolumn{1}{c|}{28.06$\pm$6.51}    & 24.68$\pm$3.98     & \multicolumn{1}{c|}{4.99$\pm$1.18}     & \multicolumn{1}{c|}{4.96$\pm$1.14}    & \multicolumn{1}{c|}{27.66$\pm$5.29}    & 24.37$\pm$3.58     \\ \hline
\multirow{5}{*}{90 days}   & ARIMA          & \multicolumn{1}{c|}{7.44$\pm$0.50}     & \multicolumn{1}{c|}{6.8$\pm$0.29}     & \multicolumn{1}{c|}{35.68$\pm$3.10}    & 31.29$\pm$2.36     & \multicolumn{1}{c|}{5.08$\pm$0.72}     & \multicolumn{1}{c|}{5.8$\pm$0.29}     & \multicolumn{1}{c|}{34.47$\pm$4.33}    & 27.71$\pm$2.44     \\ \cline{2-10} 
                           & Prophet        & \multicolumn{1}{c|}{6.23$\pm$0.45}     & \multicolumn{1}{c|}{6.03$\pm$0.32}    & \multicolumn{1}{c|}{35.13$\pm$5.51}    & 32.27$\pm$3.07     & \multicolumn{1}{c|}{4.95$\pm$0.39}     & \multicolumn{1}{c|}{4.62$\pm$0.31}    & \multicolumn{1}{c|}{28.87$\pm$5.32}    & 25.01$\pm$2.49     \\ \cline{2-10} 
                           & Lasso Regression         & \multicolumn{1}{c|}{6.38$\pm$0.50}     & \multicolumn{1}{c|}{5.13$\pm$0.49}    & \multicolumn{1}{c|}{32.71$\pm$5.83}    & 24.82$\pm$1.81     & \multicolumn{1}{c|}{4.94$\pm$0.58}     & \multicolumn{1}{c|}{5.06$\pm$0.40}    & \multicolumn{1}{c|}{34.47$\pm$4.33}    & 24.66$\pm$2.46     \\ \cline{2-10} 
                           & Random Forest  & \multicolumn{1}{c|}{6.06$\pm$1.27}     & \multicolumn{1}{c|}{5.84$\pm$1.72}    & \multicolumn{1}{c|}{28.89$\pm$3.53}    & 25.59$\pm$1.92     & \multicolumn{1}{c|}{5.2$\pm$0.85}      & \multicolumn{1}{c|}{5.5$\pm$1.18}     & \multicolumn{1}{c|}{28.4$\pm$2.86}     & 25.39$\pm$1.45     \\ \cline{2-10} 
                           & LSTM Networks  & \multicolumn{1}{c|}{5.26$\pm$0.63}     & \multicolumn{1}{c|}{5.77$\pm$1.74}    & \multicolumn{1}{c|}{27.76$\pm$3.74}    & 24.29$\pm$1.64     & \multicolumn{1}{c|}{5.48$\pm$0.55}     & \multicolumn{1}{c|}{5.34$\pm$1.24}    & \multicolumn{1}{c|}{29.95$\pm$3.03}    & 24.68$\pm$1.89     \\ \hline
\end{tabular}
\label{tab:errors}
\end{table}
\end{landscape}

\section{Comparison and Discussion}
\label{sec:Comparison}
In this section, we compare the models and provide recommendations for using these models in various scenarios. In Section \ref{sec:UnivartiateVsMultivariate} we compare the models based on a training window of two years, in Section \ref{sec:TwoVsEight} we discuss the impact of an increased amount of data on the forecasting models, and in Section \ref{sec:Retraining} we discuss the effect of different retraining periods on the models. Finally, in Section \ref{sec:Theoretical} we provide the overall methodological implications of the study and in Section \ref{sec:Managerial} discuss managerial implications.
\subsection{Univariate versus Multivariate Models}
\label{sec:UnivartiateVsMultivariate}
We have presented five different models for platelet demand forecasting that can be divided into two groups: univariate and multivariate. Univariate models, ARIMA and Prophet, forecast future demand based only on the demand history. Although the ARIMA model only considers a limited number of previous values for forecasting the demand, retraining it every day, week or month leads to a slight performance improvement. The Prophet model incorporates the historical data, seasonality and holiday effects into the demand forecasting model which results in an improvement in the forecasting accuracy by approximately 10\% compared to ARIMA. This highlights the impact of weekday/weekend and holiday effects in the platelet demand variation. As we discussed in Section \ref{sec:Trends}, there is a weekday/weekend effect for platelet demand, which is not (directly) captured in the ARIMA model.

Multivariate models, on the other hand, incorporate clinical predictors as well as historical demand data for demand forecasting. We use lasso regression to select the dominant clinical predictors that affect the demand. Lasso regression examines the linear relationship among the clinical predictors and their influence on the demand. However, as presented in Figure \ref{fig:Corr}, there are correlations among the clinical predictors. There may also be nonlinear relationships among these clinical predictors that cannot be captured by a linear regression model. These issues motivated us to use two machine learning approaches, random forest and LSTM network. Random forest is capable of capturing nonlinearities among variables and its forecasting method of averaging past values provides some contrast to LSTM network's modelling approach. An LSTM network can also account for nonlinearities among variables. Moreover, an LSTM network is capable of retaining past information while forgetting some parts of the historical data. As we can see from Table \ref{tab:errors}, in general, random forest, LSTM network and lasso regression have low forecast errors for different training window sizes, owing to the inclusion of the clinical predictors.

\subsection{Two Years versus Eight Years of Data}
\label{sec:TwoVsEight}
As discussed in Section \ref{sec:DemandForecasting}, we train our models for two training window sizes, with two years and eight years of data, respectively. Since there is no trend in the data from 2016 onwards (see Figure \ref{fig:TSD}), in the first scenario the models are trained for two years (training window size of two years, starting from 2016). With this amount of data and by retraining every year, forecasts are not accurate for univariate time series approaches, and one needs to include the clinical predictors in the forecasting model. However, by considering a training window of eight years, the ARIMA model's performance improves by approximately 20\%, compared to the case of a two year training window. The Prophet model's performance also improves when more data are available, specifically when it is trained less frequently (30 and 90 days retraining windows).

In general the multivariate models result in small forecasting errors for two years of data for training, and do not perform significantly better as the amount of data increases, which shows that there is not much sensitivity to the training window size. This highlights the importance of including the clinical predictors in the forecasting process.
\subsection{Different Retraining Periods}
\label{sec:Retraining}
We also compare different retraining periods and provide insight on how to choose the appropriate retraining period for this data (and in general). Our results show that considering different retraining periods does not affect the models in the same manner. While in general all the models benefit from retraining more frequently, univariate models benefit more. For the univariate models, the greatest performance increase is for the ARIMA model when retrained every day, resulting in a decrease of 50\% in MAPE and SMAPE. For the multivariate models, lasso regression has an impressive performance increase when retrained every day, while random forest and LSTM networks show less sensitivity to the retraining period. So, by considering the overhead of retraining these models more frequently, one may decide to choose to use a larger retraining window for random forest and LSTM networks.

Generally, if the retraining period is small, meaning that the models are retrained more frequently, the mean forecast accuracy representing the long-term overall performance is improved.
\subsection{Methodological implications of the study}
\label{sec:Theoretical}
In general, when there is access only to previous demand values, using a univariate model and retraining it frequently is effective. In the case that several data variables are available, lasso regression, random forest models and LSTM networks can forecast the demand with higher accuracy even when a small amount of data is available and without frequent retraining. Forecasting problems can have linear or nonlinear relationships among the model variables. Due to the fact that LSTM networks can work on both linear and nonlinear time series, and are able to capture nonlinear dependencies, they can outperform linear regression models when long term correlations exist in the time series. Based on the LSTM results, we conclude that long term correlations and nonlinearity are not major issues for our data since the LSTM model does not significantly outperform lasso regression.

While LSTM networks perform well even with a limited amount of data and they can capture nonlinear relationships, they lack interpretability. Interpretability is an important feature of any prediction model used in a safety critical setting like blood product distribution. Considering the time and memory complexity, and interpretability of these models, lasso regression has lower time and memory complexity while it is also very interpretable. Random forest models maintain interpretability while also having the ability to capture nonlinear relationships. Random forests do well when their training data has good coverage of the different feature combinations the model is forecasting. This is because random forest models make forecasts for a set of features by averaging together similar data points from the training data. This allows random forest models to extract nonlinear relationships but also means they cannot extract trends effectively and may need a large amount of data in order to work well. This can be seen in our model (see Table \ref{tab:errors}), a training window of eight years, with more training data points to reference, has a small improvement in the error measures over a training window of two years for different retraining periods.

Training random forest models and LSTM networks requires expertise in the machine learning area since poor training will cause low-precision results. It is also worth mentioning that the LSTM network is a robust learning model and is capable of learning linear and nonlinear relationships among the model variables even in very short time series data \citep{boulmaizimpact, lipton2015learning}. However, as the number of inputs increases, both the data variables that make data wide and the data rows that make data tall, LSTM performance tends to decrease because it is highly dependent on the input size. Moreover, wide data results in model overfitting \citep{lai2018modeling}. Having wide data, one can apply a feature selection method such as lasso regression to reduce the number of variables and regularize the input.

\begin{figure}[H]
  \centering
  \includegraphics[width=0.9\textwidth]{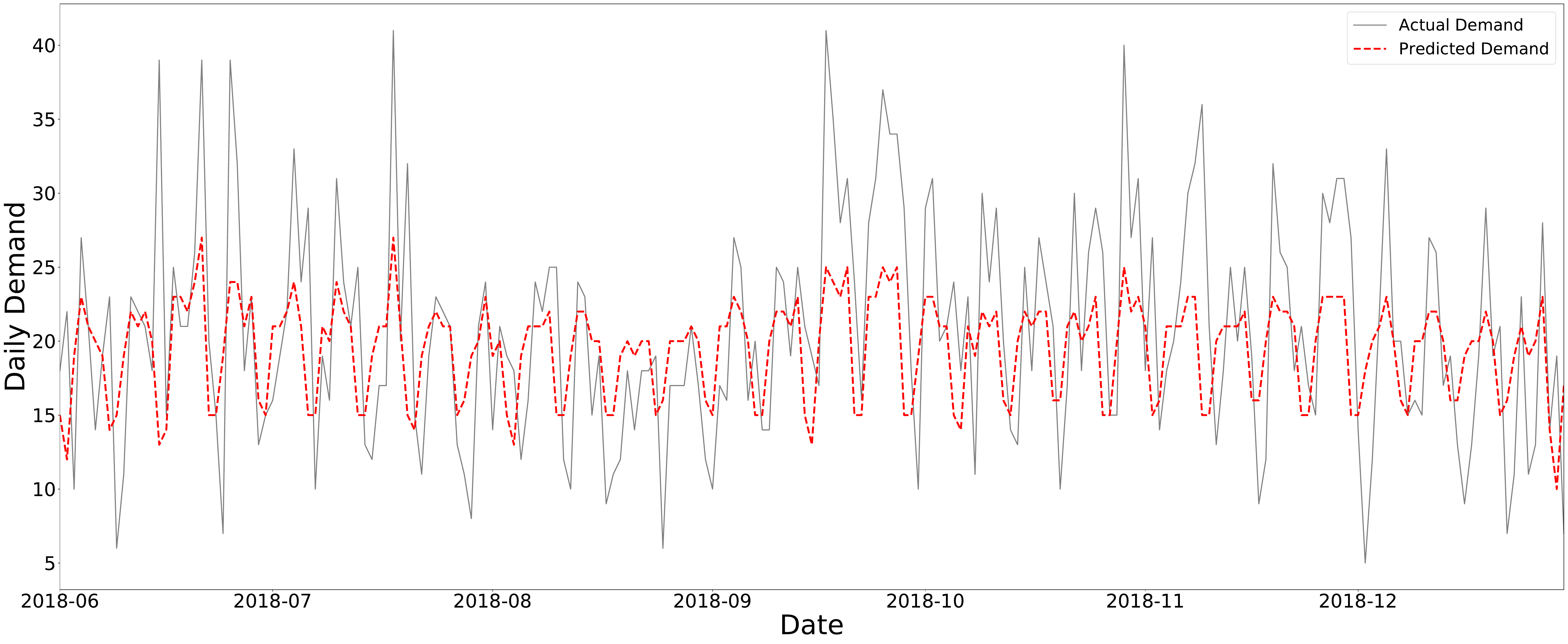}
  \captionsetup{justification=centering}
  \caption{Demand forecasting with lasso regression with a training window of two years and a retraining period of seven days}
  \label{fig:Lasso_Pre}
\end{figure}

One limitation of forecasting models is that they cannot capture sharp peaks in demand. Figure \ref{fig:Lasso_Pre} depicts the actual and predicted demands for the second half of 2018 with a training window of two years and a retraining period of 7 days (retraining weekly) using lasso regression. It appears that the model does some degree of smoothing and thus cannot detect the sharp peaks. One possible explanation is that regression models are regressed on the expectation of the outcome, and are not good at capturing the extreme deviations from this expectation. However, as shown in Figure \ref{fig:Lasso_Pre}, smoothing mostly occurs for the maxima rather than the minima. In other words, the model potentially has large errors when there is excess demand, for example in emergency situations. The results presented in Sections \ref{sec:ForecastsUni} and \ref{sec:ForecastsMulti} support that all the models struggle with capturing the peaks in demand.
To sum up, when a sufficient amount of data is available, using a univariate model results in a low forecast error, particularly in the case that it is retrained every day. Specifically, when there is only access to the previous demand (as is currently the case for CBS) and adequate historical data are available, one can benefit from a simple univariate model like ARIMA or Prophet, since univariate models are simpler than the multivariate models. Multivariate models are useful when there is access to a limited amount of data. Also, they do not necessarily require frequent retraining, which may be an important implementation concern.
\subsection{Managerial implications of the study}
\label{sec:Managerial}
The short shelf life of platelets results in wastages which not only incur large costs but also affect the environment since they cannot be reused, recycled, or recovered \citep{jemai2020environmental}. Moreover, since platelet demand is highly variable, urgent same-day deliveries are placed frequently. Apart from the high cost of urgent orders, platelet shortage can increase the risk of putting patients' lives in danger. Currently, blood suppliers are not aware of the demand at the hospitals since hospitals hold excess inventory to manage the highly variable platelet demand. Indeed, this has its roots in the bullwhip effect, that is hospitals tend to order more than their actual demand. We see an opportunity to better coordinate supply (number of units received) with demand (number of units transfused) through the development of a daily demand predictor. Forecasting the demand improves the transparency between blood suppliers and hospitals, and helps blood suppliers to make better-informed decisions.

From the clinical perspective, accurate demand forecasting is important for clinical and supply chain management purposes. Demand forecasting can be used for placing optimal platelet orders and for decision making in many parts of the supply chain such as donation planning, and resource and staff management. As we can see in Section \ref{sec:Res}, there is some fundamental limit to how accurate the demand forecasts can be, so one important challenge would be how to use the demand forecasts to inform an ordering policy in an effective manner. Clearly, forecasts themselves do not reflect an optimal ordering decision but they can be used as additional information in building effective ordering/inventory management policies (incorporating such forecasts is one of our current research directions).

Moreover, this research provides a holistic analysis of the predictors that affect the platelet demand, including the clinical predictors, hospital locations, day of the week and demand history. This can help blood suppliers with adapting clinically relevant factors into the decision making process, like decisions regarding the assignment of transfusion related staff/resources (beds or equipment).

Overall, there is a significant caveat with all of these approaches in that there are still forecasting errors, in particular they all struggle with capturing peaks. These underestimations may cause significant concerns for using such forecasts directly as there is the danger of severe underestimation. Therefore, some adjustments may be required for using these forecasts according to specific objectives. For instance, one may need an optimization model for inventory control if the demand forecasts are used for inventory management.

\section{Conclusion}
\label{sec:Conclusion}
In this study, we utilized two types of methods for platelet demand forecasting, univariate and multivariate methods. Univariate methods, ARIMA and Prophet, forecast platelet demand only by considering the historical demand information, while multivariate methods, lasso regression, random forest and LSTM networks, also consider clinical predictors. The error levels for the univariate models, particularly in the case that a small amount of data is available, motivates us to utilize clinical predictors to investigate their ability to improve the accuracy of forecasts. Results show that lasso regression, random forest and LSTM networks outperform the univariate methods when a limited amount of data are available. Moreover, since they include clinical predictors in the forecasting process, their results can aid in building a robust decision making and blood utilization system. However, their application is not limited to platelet products. We believe that they can be used in various areas, including healthcare in general, finance and climate studies, when data features are available. On the other hand, when there is access to a sufficient amount of data, the marginal improvement for a simple univariate model such as ARIMA is higher than for multivariate models. In such scenarios, univariate models can be applied to historical data for demand forecasting, regardless of the product, which makes these models generalizable and widely applicable.

Future extensions of this work will include: (i) proposing an optimal ordering policy based on the predicted demand over a planning horizon with ordering cost, wastage cost and shortage (same-day order) cost; (ii) further exploring the lasso regression approach to enhance variable selection, with a particular focus on interpretability (this will not only affect the lasso regression itself, but also may improve LSTM forecasting accuracy since LSTM inputs are selected using lasso regression); (iii) more extensive empirical evaluation of the proposed models; (iv) exploring the generality of the results (outside of Hamilton).
%
\bibliography{Ref}
\pagebreak
\appendix
\setcounter{table}{0}
\renewcommand{\thetable}{A\arabic{table}}

{\Large \textbf{\chapter{Appendix A}}}
\label{App:A}

Table \ref{tab:CovLasso} gives the selected predictors using lasso regression. Considering the coefficients for the predictors and their corresponding confidence intervals in Table \ref{tab:CovLasso}, and based on \citep{ranstam2012p}, variables that have a coefficient of zero and confidence intervals that are symmetric around zero are candidates to be eliminated. As we can see from Table \ref{tab:CovLasso}, abnormal\underline{{ }}plt has the highest coefficient. The predictors abnormal\_hb and abnormal\_redcellwidth can be considered as two other important lab tests for forecasting the demand. Day of the week, last week's platelet usage and yesterday's platelet usage also have notable impact on the platelet demand. As we can see in Table \ref{tab:CovLasso}, unexpectedly, some of the predictors have a negative coefficient in the demand forecasting model. The reason is that, as we can see from Figure \ref{fig:Corr}, there are high correlations among the predictors that result in interactions among the model predictors, which may cause multicollinearity issues. Specifically, the predictors abnormal\_hb, abnormal\_INR, abnormal\_hematocrit, and abnormal\_MPV are correlated with abnormal\_plt. The predictors abnormal\_hematocrit and abnormal\_hb also have high correlations with most of the other abnormal laboratory test results.
\begin{table}[htbp]
    \centering
    \caption{predictors and their corresponding coefficients for lasso regression}
    \scalebox{0.8}{
    \begin{tabular}{|l|l|l|}
    \hline
    \textbf{predictors} & \textbf{~~~~~~Coefficients~~~~~~}  & \textbf{~~~~~~$95\%$ Confidence Interval~~~~~~} \\ \hline
    abnormal\_ALP & -0.02 & (-0.08 , 0.04) \\ \hline
    abnormal\_MPV	& 0.01	& (-0.06 , 0.11) \\ \hline
    abnormal\_hematocrit	& 0.00 & (-0.11 , 0.14) \\ \hline
    abnormal\_PO2 & -0.11 & (-0.19 , 0.00) \\ \hline
    abnormal\_creatinine	& 0.03 & (-0.03 , 0.11) \\ \hline
    abnormal\_INR &  0.06 & (-0.02 , 0.22) \\ \hline
    abnormal\_MCHb & 	-0.03 & (-0.10 , 0.04) \\ \hline
    abnormal\_MCHb\_conc & -0.03 & (-0.10 , 0.04) \\ \hline
    abnormal\_hb	& 0.05	& (-0.04 , 0.19) \\ \hline
    abnormal\_mcv	& -0.03	& (-0.11 , 0.04) \\ \hline
    abnormal\_plt	& 0.23	& (0.02 , 0.36) \\ \hline
    abnormal\_redcellwidth & 0.07	& (0.00 , 0.15) \\ \hline
    abnormal\_wbc	& -0.02	& (-0.09 , 0.03) \\ \hline
    abnormal\_ALC	& 0.01	& (-0.05 , 0.08) \\ \hline
    location\_GeneralMedicine & -0.11 & (-0.21 , 0.00) \\ \hline
    location\_Hematology & 0.04 & (-0.02 , 0.16) \\ \hline
    location\_IntensiveCare & 0.05 & (-0.01 , 0.15) \\ \hline
    location\_CardiovascularSurgery & 0.04 & (-0.03 , 0.11) \\ \hline
    location\_Pediatric	& 0.04	& (-0.02 , 0.10) \\ \hline
    Monday & 0.07 & (0.00 , 0.16) \\ \hline
    Tuesday & 0.07 & (0.00 , 0.14) \\ \hline
    Wednesday & 0.00 & (-0.04 , 0.07) \\ \hline
    Thursday & 0.01 & (-0.03 , 0.09) \\ \hline
    Friday & -0.39 & (-0.46 , -0.31) \\ \hline
    Saturday & -0.31 & (-0.39 , -0.23) \\ \hline
    Sunday & 0.10 & (0.03 , 0.18) \\ \hline
    lastWeek\_Usage & 0.12 & (0.05 , 0.19) \\ \hline
    yesterday\_Usage & 0.10 & (0.02 , 0.17) \\ \hline
    yesterday\_ReceivedUnits & 0.06 & (0.00 , 0.14) \\ \hline
    \end{tabular}
    }
    \captionsetup{justification=centering}
    \label{tab:CovLasso}
\end{table}
The coefficients for lab tests are high. This is consistent with the observation that the lab test results are significant indicators for platelet transfusion. The predictors abnormal\_plt, abnormal\_hb, abnormal\_ALC and abnormal\_wbc have higher coefficients and consequently higher impact on platelet demand. For day of the week, Friday and Saturday have negative coefficients due to the fact that they cover the weekend (Friday: -0.39 and Saturday: \mbox{-0.31}). For hospital census data, except for location\_GeneralMedicine, all the coefficients are in a similar range to the lab tests.

\end{document}